\documentclass[letterpaper]{article}

\usepackage{iftex}
\RequireXeTeX
\usepackage[dvipsnames,table]{xcolor}
\usepackage{iclr2027_conference,times}

\usepackage[T1]{fontenc}
\usepackage{algorithm}
\usepackage{hyperref}
\usepackage{url}
\usepackage{booktabs}
\usepackage{capt-of}
\usepackage{ragged2e}
\usepackage{amsmath}
\usepackage{amssymb}
\usepackage{amsfonts}
\usepackage{amsthm}
\usepackage{graphicx}
\usepackage{flafter}
\usepackage{microtype}
\usepackage{nicefrac}
\usepackage{algpseudocode}
\usepackage{longtable}
\usepackage{tabularx}
\usepackage{multirow}
\usepackage{array}
\usepackage{adjustbox}
\usepackage[skins,breakable]{tcolorbox}

\newtcolorbox{promptbox}[1]{
  enhanced,breakable,
  colback=black!4,colframe=black!22,
  colbacktitle=black!9,coltitle=black,
  title={#1},fonttitle=\small\sffamily\bfseries,
  fontupper=\small\ttfamily,
  boxrule=0.4pt,arc=1mm,
  left=8pt,right=8pt,top=6pt,bottom=6pt,
  before skip=8pt,after skip=8pt,
  before upper={%
    \justifying\frenchspacing\setlength{\parindent}{0pt}\setlength{\parskip}{5pt}%
    \spaceskip=\fontdimen2\font plus 0.2em minus 0.2em\relax
  }
}

\newcommand{\method}{OmniOPSD}

\definecolor{TableGroup}{gray}{0.93}
\newcommand{\standardtablesetup}{%
  \small
  \setlength{\tabcolsep}{5pt}%
  \renewcommand{\arraystretch}{1.08}%
}
\newcommand{\widetablesetup}{%
  \footnotesize
  \setlength{\tabcolsep}{3.5pt}%
  \renewcommand{\arraystretch}{1.08}%
}
\newcommand{\tablegroup}[2]{%
  \rowcolor{TableGroup}\multicolumn{#1}{c}{\itshape #2} \\
}

\title{OmniOPSD: Rationale-Privileged\\
On-Policy Self-Distillation for Affective Computing}

\iclrfinalcopy
\author{%
Zebang Cheng\textsuperscript{*,1,2}\quad
Shuimu Chen\textsuperscript{*,3}\quad
Boxue Yang\textsuperscript{*,4}\quad
Yuanshen Guan\textsuperscript{5}\quad
Jingyi Chen\textsuperscript{1,6}\\[1.5pt]
Zheng Lian\textsuperscript{7}\quad
Xiaojiang Peng\textsuperscript{6}\quad
Fei Ma\textsuperscript{\textdagger,2}\quad
LaiZhong Cui\textsuperscript{\textdagger,1}\quad
Qi Tian\textsuperscript{2,8}\\[4pt]
{\fontsize{8}{9.5}\selectfont
\textsuperscript{1} Shenzhen University\quad
\textsuperscript{2} Guangdong Laboratory of Artificial Intelligence and Digital Economy (SZ)\\
\textsuperscript{3} Tsinghua University\quad
\textsuperscript{4} Shanghai Jiao Tong University\quad
\textsuperscript{5} University of Science and Technology of China\\
\textsuperscript{6} Shenzhen Technology University\quad
\textsuperscript{7} Tongji University\quad
\textsuperscript{8} Huawei\\[3pt]
\textsuperscript{*} Equal contribution.\quad
\textsuperscript{\textdagger} Corresponding authors.\quad
\href{https://omniopsd.github.io/}{omniopsd.github.io}}%
}

\makeatletter
\renewcommand{\@maketitle}{%
  {\centering
    {\normalfont\fontsize{16}{19}\selectfont\bfseries\@title\par}%
    \vskip8pt
    {\normalfont\fontsize{10}{12}\selectfont\@author\par}%
  }%
  \vskip10pt
}
\makeatother
\renewenvironment{abstract}{%
  \par\begingroup
  \leftskip=14pt\rightskip=14pt
  \fontsize{10}{11.5}\selectfont
  \noindent\textbf{Abstract.}\enspace\ignorespaces
}{\par\endgroup\vspace{8pt}}
\hypersetup{
  hidelinks,
  pdftitle={OmniOPSD: Rationale-Privileged On-Policy Self-Distillation for Affective Computing},
  pdfauthor={Zebang Cheng, Shuimu Chen, Boxue Yang, Yuanshen Guan, Jingyi Chen, Zheng Lian, Xiaojiang Peng, Fei Ma, LaiZhong Cui, Qi Tian}
}

\begin{document}

\maketitle
\lhead{}
\renewcommand{\headrulewidth}{0pt}

\begin{abstract}
Multimodal large language models (MLLMs) have advanced affective computing
by enabling nuanced reasoning over visual, acoustic, and linguistic evidence.
A key approach to improving their generalization is outcome-based
reinforcement learning, which converts existing label-level annotations
into rewards for final-answer correctness.
However, its sparse sequence-level supervision provides limited
fine-grained guidance for interpreting multimodal affective cues and may
reward correct labels even when those cues are misinterpreted.
To address this challenge, we introduce \textbf{\method{}}, the first
on-policy self-distillation framework for omni-modal affective reasoning.
At its core, \method{} transforms label-level supervision into
evidence-rich, teacher-only rationales by leveraging reference affective
judgments as semantic priors for offline multimodal evidence elicitation
from frontier MLLMs.
Through rationale-privileged self-distillation, it couples on-policy
student rollouts with dense, evidence-informed token-level supervision
from a local self-teacher, allowing the student to learn along its own
trajectories rather than imitate fixed reasoning traces.
This decoupling of offline evidence acquisition from local policy
learning enables training without additional human annotation,
frontier-model logits, or online frontier-model calls, while preserving
student-only inference from the original task input.
At both 3B and 7B scales, \method{} outperforms supervised fine-tuning
(SFT) and group relative policy optimization (GRPO) in mean macro-F1
on both in-domain and out-of-domain evaluations.
Among the compared models, it ranks first by MER-UniBench mean across
all three input-modality settings, reaching $84.13$ with
audio--video--text input, and achieves the highest MME-Emotion
recognition score (Rec-S) of $42.4$.
\end{abstract}

\section{Introduction}

Affective computing advances human-centered artificial intelligence by
modeling human emotion, sentiment, and intent, with applications spanning
education, mental healthcare, and human--computer interaction
\citep{poria2017review}.
Inferring these states requires interpreting facial expressions, vocal
prosody, and language whose affective meaning is often ambiguous and
context dependent \citep{barrett2019emotional}.
Multimodal large language models (MLLMs) extend affective computing
beyond fixed-category prediction toward richer reasoning
\citep{lian2025affectgpt,lian2025mer}.
This shift calls for models that not only predict affective states
accurately but also explain how multimodal evidence supports their
judgments, promoting transparent and interpretable inference
\citep{zhang2026mmeemotion}.

\begin{figure*}[!t]
    \centering
    \includegraphics[width=\textwidth]{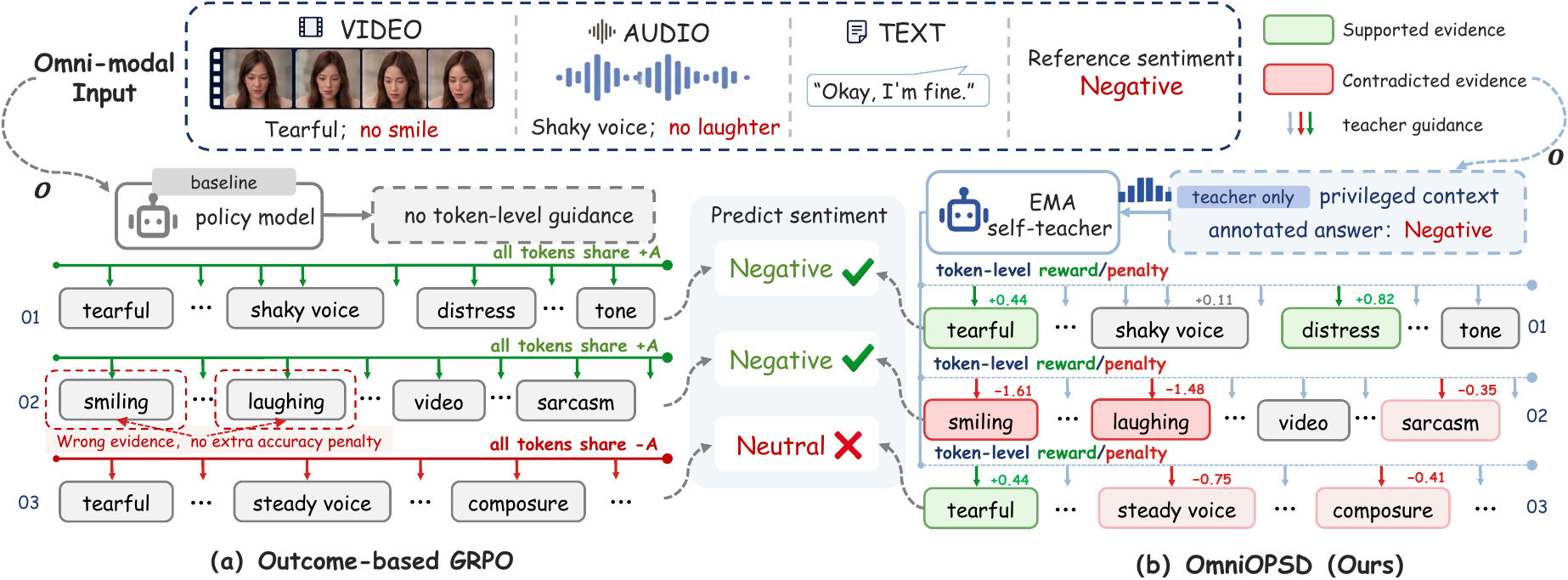}
    \caption{Comparison of (a) outcome-based GRPO and (b) \method{}.
    GRPO assigns the same sequence-level advantage to all tokens, including
    those in erroneous reasoning. \method{} provides dense, differentiated
    token-level supervision from a rationale-privileged self-teacher.}
    \label{fig:comparison}
\end{figure*}

Yet multimodal affective data resources primarily provide labels,
with limited human-authored evidence annotations
\citep{busso2008iemocap,zadeh2018multimodal,poria2019meld,zhang2024mintrec}.
Acquiring such annotations requires labor-intensive interpretation
of heterogeneous, context-dependent cues, making high-quality
reasoning supervision difficult to scale
\citep{cheng2024emotion,lin20243,lian2025affectgpt,zhang2026xemogpt}.
The practical question is therefore how to turn existing label-level
annotations into reasoning-level supervision for affective MLLMs.

A natural response is to use reference judgments to construct outcome
rewards for reinforcement learning with verifiable rewards (RLVR),
often optimized with group relative policy optimization (GRPO)
\citep{shao2024deepseekmath,lian2025affectgptr1,wu2026facial,ICLR2026_f96c211f}.
This enables on-policy learning from student-generated responses
without prescribing reference rationales.
However, final-answer and format rewards provide only sequence-level
feedback: responses are rewarded for correct final answers even when
their explanations misinterpret or hallucinate supporting cues
\citep{han2026omni}.
Explicit reasoning does not necessarily improve affective prediction
\citep{lian2025affectgptr1}, highlighting the need for evidence-informed
guidance along student-generated trajectories.

On-policy distillation (OPD) provides dense token-level supervision
along student-generated trajectories
\citep{agarwal2024policy,lu2025onpolicydistillation}.
Distillation from frontier teachers faces limited logit
access and tokenizer mismatches \citep{boizard2024towards}, while
logit-free alternatives require online teacher queries \citep{zhou2026omniopd}.
On-policy self-distillation (OPSD) uses a local self-teacher whose
informational advantage comes from feedback or reference solutions
\citep{hubotter2026reinforcement,zhao2026self}, or asymmetric access
to original and masked or degraded images
\citep{li2026selfsupervised}.
Extending these designs to omni-modal affective reasoning, however,
is nontrivial: labels specify outcomes rather than their evidential
basis, while richer perceptual access alone leaves implicit how
ambiguous, potentially conflicting cross-modal cues jointly support
affective judgments.

To bridge this gap, we introduce \textbf{\method{}}, a
\textbf{Rationale-Privileged On-Policy Self-Distillation} framework
for omni-modal affective reasoning (Figure~\ref{fig:comparison}).
Our central insight is that reference affective judgments can do
more than evaluate outcomes: they can anchor the acquisition of
multimodal evidence that informs learning.
\method{} leverages this insight to construct teacher-only rationales
from offline frontier-elicited evidence and reference judgments,
recasting rationales as a basis for constructing supervision rather
than fixed targets for student imitation.
A local self-teacher translates this privileged context into dense
token-level guidance tailored to the student's own evolving
trajectories, turning offline evidence into trajectory-dependent
supervision.
By decoupling external evidence acquisition from local policy
learning, \method{} bridges label-level annotation and
evidence-informed on-policy learning without additional human
annotation, frontier-model logits, or online frontier-model calls,
while retaining student-only inference from the original task input.
Our contributions are threefold:
\begin{enumerate}
\item We introduce \method{}, which leverages reference
judgments as semantic priors to elicit multimodal evidence
descriptions from frontier MLLMs offline, combining both into
teacher-only privileged context, not imitation targets.
To our knowledge, \method{} is the first on-policy
self-distillation framework for omni-modal affective reasoning.

\item We formulate rationale-privileged on-policy self-distillation,
coupling on-policy student rollouts with dense token-level supervision
from an EMA self-teacher conditioned on privileged context.
This requires neither frontier-model logits nor additional human
annotation.

\item At 3B and 7B scales, \method{} outperforms supervised
fine-tuning (SFT) and GRPO in mean in-domain and out-of-domain macro-F1.
Among the compared models, it achieves the highest MER-UniBench
mean score ($84.13$), ranks first by mean score in all three
input-modality settings, and attains the highest MME-Emotion
Rec-S ($42.4$).
\end{enumerate}

\section{Related Work}

\subsection{Multimodal Affective Reasoning}

Masked pretraining and transfer learning strengthen facial, acoustic,
and textual affect representations
\citep{sun2023maedfer,chen2024static,chen2026staticfordynamic,
ma2024emotion2vec,tian2020skep}.
Building on these representations, multimodal emotion recognition
fuses visual, acoustic, and linguistic features to capture
complementary affective cues and contextual dependencies
\citep{cheng2023semi,wang2025big,fang2025emoe}.
Large language models extend affective modeling toward cause analysis
and descriptive understanding \citep{cheng2024mips,lian2025mer},
supported by emotion-oriented instruction data and supervised fine-tuning
\citep{xie2024emovit,cheng2024emotion,lian2025affectgpt}.
Recent GRPO-based post-training further explores task rewards,
perceptual grounding, and tool-assisted or slow--fast reasoning
\citep{zhao2025r1,lian2025affectgptr1,han2026omni,
liu2026actfer,han2026mer}.
\method{} complements these advances by using offline evidence
descriptions as privileged teacher context for dense supervision
on student-generated responses.

\subsection{On-Policy Distillation and Self-Distillation}

On-policy distillation applies teacher supervision to student-generated
trajectories, typically through token-level distribution matching
\citep{agarwal2024policy,lu2025onpolicydistillation}.
When teacher logits are unavailable for direct distribution matching,
logit-free alternatives instead derive feedback from sampled teacher
continuations \citep{zhou2026omniopd}.
Context-conditioned on-policy self-distillation uses reference solutions
or textual feedback to guide a self-teacher's predictions on
student-generated prefixes
\citep{hubotter2026reinforcement,zhao2026self,cao2026x}.
Further work examines generalization risks from rich teacher context
\citep{kim2026does} and develops reflection-guided, localized correction
\citep{zhao2026rosd}.
Multimodal extensions ground teacher supervision in image evidence
through image crops or textual descriptions of question-relevant
visual cues \citep{yuan2026vision,tian2026vicur}.
Alongside these visual grounding strategies, \citet{wang2026seeing}
separate image-based perceptual supervision from reference-conditioned
reasoning guidance to reduce answer-driven shortcuts.
\method{} studies rationale-privileged on-policy self-distillation
for omni-modal affective reasoning, using offline multimodal evidence
descriptions as teacher context.
\section{Method}
\label{sec:method}

\begin{figure}[!t]
  \centering
  \includegraphics[width=\textwidth]{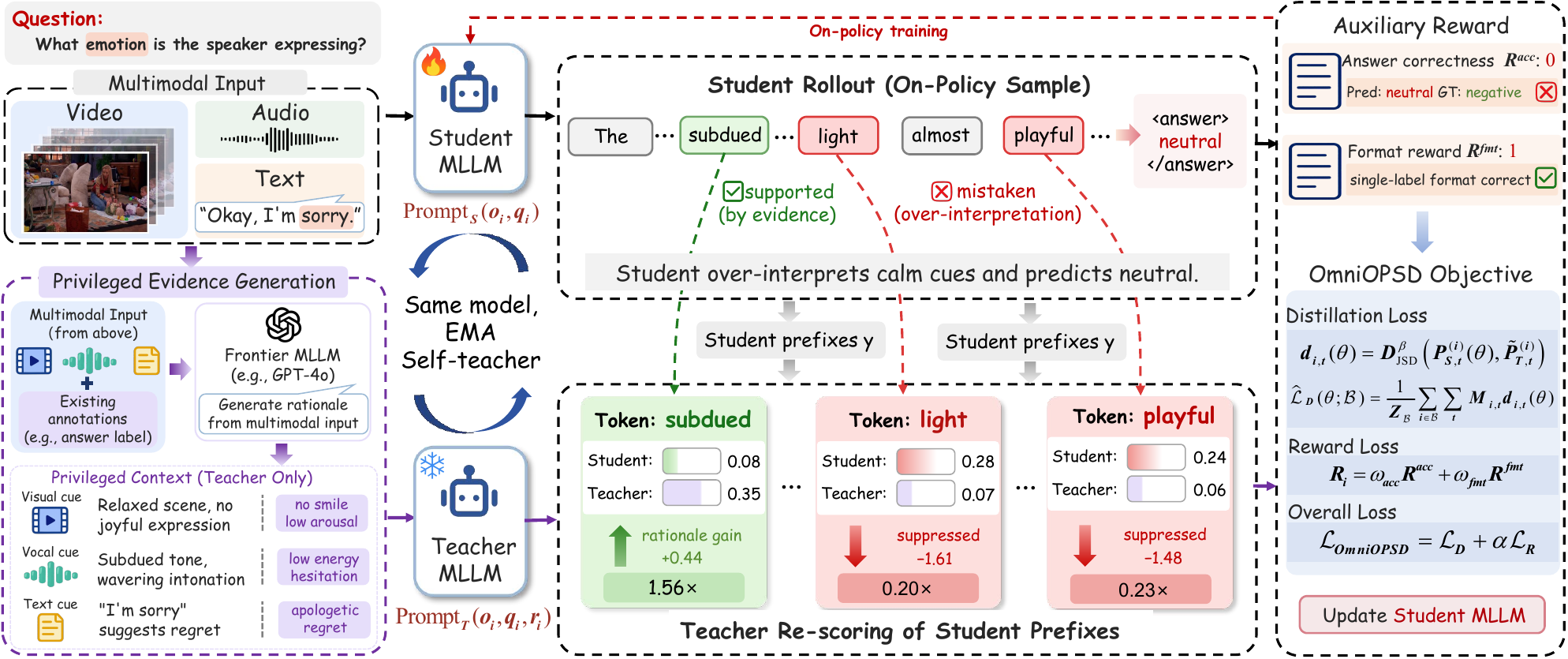}
  \caption{\textbf{Overview of \method{}.}
  The framework supports video, audio, and text inputs and trains the
  student through distillation from a teacher with privileged information.
  The example illustrates guidance on affective reasoning despite an
  incorrect final prediction.}
  \label{fig:omniopsd-overview}
\end{figure}

This section presents \method{}, a rationale-privileged on-policy
self-distillation framework for omni-modal affective reasoning
(Figure~\ref{fig:omniopsd-overview}). We define the task and privileged
context, describe student rollouts and teacher re-scoring, and introduce
the distillation objective and teacher update.

\subsection{Task and Privileged Context}

We consider multimodal affective tasks such as emotion, sentiment, and intent
recognition, where predictions depend on interpreting the available visual,
acoustic, and linguistic signals in context. Let
$\mathcal{D}=\{(o_i,q_i,a_i,r_i)\}_{i=1}^{N}$ contain $N$ training examples indexed by $i$,
where $o_i$ contains the available omni-modal input, $q_i$ is the task
instruction, and $a_i$ is the annotated or consensus answer.
Each reference \emph{rationale}
$r_i=\operatorname{Serialize}(e_i,a_i)$ pairs $a_i$ with
frontier-generated reasoning $e_i$. When available, this reasoning describes
potentially relevant facial behavior, vocal delivery, linguistic content,
and temporal context, including how verbal and nonverbal cues inform the
affective interpretation. These descriptions give the teacher an auxiliary
interpretation of the input
(Appendix~\ref{app:rationale-prompts}).

The student and teacher receive the same omni-modal input and task
instruction, with the reference rationale included only in the teacher
context:
\begin{align}
 c_i^S &= \operatorname{Prompt}_S(o_i,q_i), \label{eq:student-context}\\
 c_i^T &= \operatorname{Prompt}_T(o_i,q_i,r_i). \label{eq:teacher-context}
\end{align}

During these updates, $r_i$ serves exclusively as teacher context and is not
a supervised target sequence. The student constructs its response from the
omni-modal input and task instruction in $c_i^S$.

\subsection{Student Rollout and Teacher Re-Scoring}

The student is prompted to explain its interpretation of the available
multimodal cues and then provide the task-specific answer.
At optimizer step $k$, the student with parameters $\theta_k$ samples
a completion $y_i=(y_{i,1},\ldots,y_{i,L_i})$ from $c_i^S$,
where $L_i$ is the completion length, $y_{i,t}$ its $t$-th token,
and $y_{i,<t}$ the prefix before position $t$.
Sampling follows decoding configuration $G$, with settings for the
main benchmark run given in Appendix~\ref{app:backbones-cold-start}.
The teacher obtains its target distributions by re-scoring these
student-generated prefixes under $c_i^T$, without decoding a separate
completion. Reference rationales thus inform token-level continuation targets
along the student's own interpretations of affective cues, regardless of
final-answer correctness.

The student and local teacher share an architecture, tokenizer, and vocabulary
$\mathcal{V}$, with $u_{\theta}(c,y_{<t})$ denoting the next-token logits
and $\tau>0$ the distillation temperature.  For a fixed rollout, the
differentiable student distribution and teacher target are
\begin{align}
 P_{S,t}^{(i)}(\theta)
 &=\operatorname{softmax}\!\left(
 u_{\theta}(c_i^S,y_{i,<t})/\tau\right),
 \label{eq:student-distribution}\\
 \widetilde P_{T,t}^{(i)}
 &=\operatorname{stopgrad}\!\left[
 \operatorname{softmax}\!\left(
 u_{\widetilde\theta_k}(c_i^T,y_{i,<t})/\tau\right)\right],
 \label{eq:teacher-distribution}
\end{align}
where $\bar\theta_k$ denotes the EMA teacher parameters, with
$\widetilde\theta_k=\theta_k$ before their initialization and
$\widetilde\theta_k=\bar\theta_k$ afterward. Student gradients are evaluated at
$\theta=\theta_k$, with teacher targets fixed during backpropagation
(Appendix~\ref{app:algorithm-implementation}).

\subsection{Token-Normalized Self-Distillation}

At each valid completion position, \method{} minimizes the generalized
Jensen--Shannon divergence (JSD) between the student and the fixed teacher target.
For mixing coefficient $\beta\in(0,1)$,
\begin{equation}
 d_{i,t}(\theta)
 =D_{\mathrm{JSD}}^{\beta}
 \!\left(P_{S,t}^{(i)}(\theta),\widetilde P_{T,t}^{(i)}\right),
\end{equation}
with the mixture and the decomposition into Kullback--Leibler (KL)
divergences given in Appendix~\ref{app:theory-jsd}.
For a local batch $\mathcal B$, the token-normalized loss is
\begin{equation}
 \widehat{\mathcal L}_{D}(\theta;\mathcal B)
 =
 \frac{1}{Z_{\mathcal B}}
 \sum_{i\in\mathcal B}\sum_t M_{i,t}\,d_{i,t}(\theta),
 \label{eq:distill-loss}
\end{equation}
where $M_{i,t}\in\{0,1\}$ selects valid completion tokens and
$Z_{\mathcal B}=\sum_{i\in\mathcal B}\sum_t M_{i,t}>0$ counts them.

Distillation covers all valid completion tokens, including those used for the
explanation and final answer, so rationale-conditioned teacher preferences can
shape the full generated response.

With sampled tokens and teacher targets fixed, differentiating
Eq.~\eqref{eq:distill-loss} yields an on-policy semi-gradient over visited
prefixes (see Appendix~\ref{app:theory} for JSD properties and derivations).

\subsection{Optimization and Teacher Update}

For emotion, sentiment, and intent tasks, we optionally complement
distillation with rewards $R^{\mathrm{acc}}$ for agreement with the annotated
answer and $R^{\mathrm{fmt}}$ for output-format compliance. Their weighted
sum, with fixed hyperparameters $\omega_{\mathrm{acc}}$ and
$\omega_{\mathrm{fmt}}$, defines the sequence reward
(Appendix~\ref{app:backbones-cold-start}):
\begin{equation}
 R_i=\omega_{\mathrm{acc}}R^{\mathrm{acc}}(y_i,a_i)+\omega_{\mathrm{fmt}}R^{\mathrm{fmt}}(y_i).
 \label{eq:reward}
\end{equation}
A sequence-level advantage derived from $R_i$ weights the per-completion,
length-normalized surrogate $\widehat{\mathcal L}_{R}(\theta;\mathcal B)$
(Appendix~\ref{app:algorithm-reward}). On the same local batch, the combined
objective uses $\alpha\ge0$ to weight this surrogate relative to distillation:
\begin{equation}
 \widehat{\mathcal L}_{\mathrm{\method{}}}(\theta;\mathcal B)
 =
 \widehat{\mathcal L}_{D}(\theta;\mathcal B)
 +\alpha\,\widehat{\mathcal L}_{R}(\theta;\mathcal B),
 \qquad \alpha\geq0.
 \label{eq:final-loss}
\end{equation}

An exponential-moving-average (EMA) teacher smooths the rationale-conditioned
scoring parameters. Scoring initially uses the current student parameters
without gradients. The updated student initializes the teacher state after
the first optimizer step, and subsequent steps use EMA updates with
decay $\mu\in[0,1)$ (Appendix~\ref{app:algorithm}):
\begin{equation}
 \bar\theta_{k+1}
 =\mu\bar\theta_k+(1-\mu)\theta_{k+1},
 \qquad \mu\in[0,1).
 \label{eq:ema-teacher}
\end{equation}

With reference rationales cached offline, this optimization stage requires
no frontier-model logits or online frontier-model calls. At inference, the
student performs the affective task using only the available omni-modal
input and task instruction.

\section{Experiments}
\label{sec:experiments}

We evaluate \method{} on MER-UniBench and MME-Emotion for affective
recognition and reasoning, with controlled comparisons against SFT
and GRPO at 3B and 7B scales. Ablations assess the contributions of
on-policy sampling and rationale-privileged self-distillation.
Training dynamics, token-level diagnostics, and qualitative cases further
characterize learning behavior and teacher guidance along
student-generated reasoning trajectories.

\begin{table*}[t]
  \centering
  \widetablesetup
  \setlength{\tabcolsep}{3pt}
  \caption{\textbf{Main results on MER-UniBench (\%).}
  The best score in each column within each input-modality block is in
  bold, and \method{} is post-trained from Qwen2.5-Omni-7B.}
  \label{tab:mer-unibench-main}
  \begin{adjustbox}{max width=\textwidth,center}
  \begin{tabular}{@{}lcccccccccc@{}}
    \toprule
    \multirow[c]{2}{*}[-0.6ex]{\textbf{Model}}
    & \multicolumn{4}{c}{\textbf{Sentiment}}
    & \multicolumn{4}{c}{\textbf{Basic emotion}}
    & \textbf{Fine-grained}
    & \multirow{2}{*}{\textbf{Mean}} \\
    \cmidrule(lr){2-5}
    \cmidrule(lr){6-9}
    \cmidrule(lr){10-10}
    & MOSI & MOSEI & SIMS & SIMS v2
    & MER2023 & MER2024 & MELD & IEMOCAP
    & OV-MERD+ & \\
    \midrule
    \tablegroup{11}{Input modality: Audio + Text}
    OneLLM & 64.01 & 54.09 & 63.39 & 61.98 & 25.52 & 17.21 & 28.32 & 33.44 & 22.25 & 41.14 \\
    SECap & 55.76 & 54.18 & 59.51 & 57.41 & 40.95 & 52.46 & 25.56 & 36.92 & 36.97 & 46.64 \\
    PandaGPT & 66.06 & 61.33 & 62.93 & 58.88 & 33.57 & 39.04 & 31.91 & 36.55 & 31.33 & 46.84 \\
    Qwen-Audio & 70.09 & 46.90 & 70.73 & 65.26 & 41.85 & 31.61 & 49.09 & 35.47 & 32.36 & 49.26 \\
    SALMONN & 81.00 & 67.03 & 68.69 & 65.93 & 55.53 & 45.38 & 45.62 & 46.84 & 45.00 & 57.89 \\
    AffectGPT & 83.46 & 80.74 & 82.99 & 83.75 & 72.94 & 73.41 & 56.63 & 55.68 & 59.98 & 72.18 \\
    AffectGPT-R1 & 80.13 & 80.01 & 84.49 & 82.31 & \textbf{81.69} & \textbf{93.49} & 63.74 & 63.85 & \textbf{65.49} & 77.24 \\
    \textbf{\method{} (Ours)} & \textbf{88.90} & \textbf{87.37} & \textbf{87.14} & \textbf{86.52} & 77.27 & 86.47 & \textbf{71.06} & \textbf{82.75} & 56.09 & \textbf{80.40} \\
    \midrule
    \tablegroup{11}{Input modality: Video + Text}
    Otter & 52.89 & 50.44 & 57.56 & 53.12 & 16.41 & 14.65 & 22.57 & 29.08 & 16.63 & 34.82 \\
    Video-LLaVA & 56.37 & 61.64 & 53.28 & 57.45 & 36.93 & 30.25 & 30.73 & 38.95 & 34.00 & 44.40 \\
    PandaGPT & 58.50 & 64.25 & 62.07 & 65.25 & 39.13 & 47.16 & 38.33 & 47.21 & 35.07 & 50.77 \\
    Video-ChatGPT & 54.42 & 63.12 & 64.82 & 65.80 & 44.86 & 46.80 & 37.33 & 56.83 & 39.80 & 52.64 \\
    VideoChat2 & 66.84 & 54.32 & 69.49 & 70.66 & 33.67 & 54.50 & 36.64 & 48.70 & 39.21 & 52.67 \\
    LLaMA-VID & 61.78 & 63.89 & 69.35 & 67.48 & 50.72 & 57.60 & 42.75 & 46.02 & 45.01 & 56.07 \\
    VideoChat & 65.13 & 63.61 & 69.52 & 72.14 & 48.73 & 57.30 & 41.11 & 48.38 & 44.52 & 56.71 \\
    Chat-UniVi & 54.53 & 63.18 & 68.15 & 66.36 & 57.62 & 65.67 & 45.61 & 52.37 & 48.00 & 57.94 \\
    mPLUG-Owl & 72.40 & 72.91 & 72.13 & 75.00 & 56.86 & 59.89 & 49.11 & 55.54 & 48.18 & 62.45 \\
    AffectGPT & 82.39 & 81.57 & 87.20 & 86.29 & 74.58 & 75.29 & 57.63 & 62.19 & 61.65 & 74.31 \\
    AffectGPT-R1 & 78.78 & 79.07 & 85.91 & 85.85 & \textbf{77.72} & \textbf{85.29} & 61.09 & 67.42 & \textbf{62.42} & 75.95 \\
    \textbf{\method{} (Ours)} & \textbf{88.29} & \textbf{88.12} & \textbf{87.53} & \textbf{87.94} & 76.32 & 85.27 & \textbf{68.16} & \textbf{79.11} & 56.24 & \textbf{79.66} \\
    \midrule
    \tablegroup{11}{Input modality: Audio + Video + Text}
    PandaGPT & 61.92 & 67.61 & 68.38 & 67.23 & 40.21 & 51.89 & 37.88 & 44.04 & 37.12 & 52.92 \\
    R1-Omni & 58.02 & 56.48 & 71.82 & 68.58 & 64.17 & 67.43 & 43.20 & 51.58 & 55.24 & 59.61 \\
    Emotion-LLaMA & 66.13 & 67.66 & 78.32 & 77.23 & 59.38 & 73.62 & 46.76 & 55.47 & 52.97 & 64.17 \\
    Emotion-LLaMAv2 & 86.28 & 87.69 & 87.72 & 86.02 & 77.28 & 86.90 & 51.32 & 84.05 & 62.90 & 78.91 \\
    AffectGPT & 81.30 & 80.90 & 88.49 & 86.18 & 78.54 & 78.80 & 55.65 & 60.54 & 62.52 & 74.77 \\
    AffectGPT-R1 & 79.39 & 79.24 & 88.25 & 84.97 & 84.32 & \textbf{93.75} & 63.12 & 74.26 & \textbf{68.05} & 79.48 \\
    OPPO & 86.50 & 84.63 & 86.22 & 88.26 & \textbf{87.73} & 90.34 & 64.06 & 73.60 & 67.16 & 81.05 \\
    \textbf{\method{} (Ours)} & \textbf{90.06} & \textbf{89.03} & \textbf{90.76} & \textbf{89.14} & 85.79 & 92.43 & \textbf{71.89} & \textbf{88.59} & 59.52 & \textbf{84.13} \\
    \bottomrule
  \end{tabular}
  \end{adjustbox}
\end{table*}

\subsection{Experimental Setup}
\label{sec:experimental-setup}

\paragraph{Datasets and evaluation metrics.}
MER-UniBench~\citep{lian2025affectgpt} covers sentiment analysis, basic
emotion recognition, and fine-grained emotion recognition, measured by
weighted-average F1 (WAF), hit rate (HIT), and emotion-wheel F1 (EW-F1),
respectively, with an unweighted mean over its nine datasets.
MME-Emotion~\citep{zhang2026mmeemotion} assesses recognition and reasoning
through Rec-S and Rea-S, which evaluate final predictions and intermediate
reasoning steps, respectively, and are combined in CoT-S.
The controlled ablations jointly train on MELD~\citep{poria2019meld},
MIntRec~2.0~\citep{zhang2024mintrec}, and
IEMOCAP~\citep{busso2008iemocap} for sentiment analysis, intent recognition, and emotion
recognition, respectively.
Out-of-domain (OOD) evaluation further assesses generalization to
Mandarin conversations through intent and emotion recognition on
MC-EIU~\citep{liu2024emotion}, and to single-label and multi-label emotion
recognition on MAFW~\citep{liu2022mafw}.
All controlled ablations use macro-F1, and the overall in-domain (ID) and
OOD scores are unweighted means over three and four tasks, respectively.
Appendix~\ref{app:datasets-benchmarks} provides further details on dataset
characteristics, evaluation splits, task-specific label mappings, and scoring
protocols.

\paragraph{Implementation details.}
Tables~\ref{tab:mer-unibench-main} and~\ref{tab:mme-emotion-main} report
one Qwen2.5-Omni-7B model post-trained by an SFT cold start and one epoch
of \method{} updates and evaluated under all three input-modality settings
and on MME-Emotion (Appendix~\ref{app:backbones-cold-start}), while the ablations
post-train Qwen2.5-Omni-3B and Qwen2.5-Omni-7B on the joint MELD,
MIntRec~2.0, and IEMOCAP mixture.
\method{} uses a learning rate of $10^{-6}$, EMA decay $\mu=0.999$,
and full-vocabulary JSD with $\beta=0.5$ and temperature $\tau=1.0$.
The reward weight is $\alpha=0.1$ for the main benchmarks and $0.2$ for ablations.
During \method{} updates, only the teacher receives rationales combining
generated reasoning and ground-truth answers.
Details of rationale construction and experimental settings are provided in
Appendix~\ref{app:rationale-prompts} and Appendix~\ref{app:experimental-settings},
respectively.

\begin{table*}[t]
  \centering
  \widetablesetup
  \setlength{\tabcolsep}{3pt}
  \caption{\textbf{Main results on MME-Emotion (\%).}
  The best result among the closed-source models and among the open-source
  audio-video-text models in each score column is in bold, and Qwen2.5-Omni is
  the base model of \method{}.}
  \label{tab:mme-emotion-main}
  \begin{adjustbox}{max width=\textwidth,center}
  \begin{tabular}{@{}lccccccccccc@{}}
    \toprule
    \multirow[c]{2}{*}[-0.6ex]{\textbf{Model}}
    & \multicolumn{8}{c}{\textbf{Task-level CoT-S}}
    & \multicolumn{3}{c}{\textbf{Mean}} \\
    \cmidrule(lr){2-9}
    \cmidrule(l){10-12}
    & ER-Lab & ER-Wild & FG-ER & FG-SA & ML-ER & Noise-ER & IR & SA
    & Rec-S & Rea-S & CoT-S \\
    \midrule
    \tablegroup{12}{Closed-source MLLMs: Video + Text}
    GPT-4o & \textbf{50.6} & 50.8 & \textbf{50.2} & \textbf{43.6} & 56.2 & 53.0 & \textbf{59.0} & 60.2 & 27.8 & \textbf{79.8} & 53.8 \\
    GPT-4.1 & 33.9 & 42.1 & 47.8 & 39.1 & 55.6 & 47.0 & 52.2 & 54.0 & 28.8 & 65.2 & 47.0 \\
    Gemini-2.0-Flash & 39.8 & 49.0 & 44.7 & 27.5 & 55.0 & 60.0 & 46.3 & 51.6 & 36.3 & 60.0 & 48.1 \\
    Gemini-2.5-Flash & 33.9 & 45.3 & 29.2 & 31.3 & 34.4 & 63.8 & 45.1 & 50.1 & 34.7 & 52.7 & 43.7 \\
    Gemini-2.5-Pro & 42.7 & \textbf{52.0} & 49.8 & 40.5 & \textbf{65.5} & \textbf{69.4} & 56.4 & \textbf{63.4} & \textbf{39.3} & 72.7 & \textbf{56.0} \\
    \midrule
    \tablegroup{12}{Open-source MLLMs: Audio or Video + Text}
    Qwen2-Audio & 26.3 & 30.3 & 21.8 & 41.1 & 63.2 & 48.2 & 36.6 & 59.5 & 34.1 & 50.4 & 42.3 \\
    Qwen2-VL-7B & 9.1 & 31.9 & 15.4 & 22.6 & 27.3 & 47.8 & 43.6 & 47.3 & 29.2 & 38.1 & 33.7 \\
    Qwen2.5-VL-7B & 42.0 & 42.1 & 40.9 & 34.4 & 49.4 & 50.3 & 55.3 & 53.6 & 28.4 & 64.8 & 46.6 \\
    Video-LLaVA & 9.9 & 11.7 & 18.7 & 39.7 & 54.9 & 29.9 & 27.6 & 45.8 & 25.8 & 32.8 & 29.3 \\
    \midrule
    \tablegroup{12}{Open-source MLLMs: Audio + Video + Text}
    Video-LLaMA & 24.8 & 35.1 & 34.1 & 34.6 & 49.2 & 33.0 & 41.4 & 41.0 & 26.1 & 48.5 & 37.3 \\
    Video-LLaMA2 & 11.9 & 35.0 & 13.3 & 27.2 & 35.5 & 29.0 & 15.5 & 32.5 & 29.2 & 27.7 & 28.4 \\
    Qwen2.5-Omni & 35.2 & 36.7 & 26.3 & 25.7 & 38.0 & 42.1 & 43.6 & 46.3 & 17.4 & 59.3 & 38.4 \\
    Emotion-LLaMA & 10.5 & 16.1 & 7.5 & 2.9 & 3.1 & 29.2 & 12.0 & 13.1 & 25.1 & 0.4 & 12.8 \\
    HumanOmni & 12.4 & 26.4 & 8.6 & 3.9 & 18.5 & 34.2 & 8.0 & 16.0 & 36.0 & 0.3 & 18.1 \\
    R1-Omni & 39.3 & 43.6 & 27.3 & 29.8 & \textbf{49.6} & 59.7 & 37.7 & 44.2 & 26.3 & 58.6 & 42.4 \\
    AffectGPT & 41.2 & 25.1 & 22.9 & 32.1 & 38.3 & 38.1 & 22.9 & 35.0 & 11.9 & 50.6 & 31.2 \\
    OPPO & \textbf{54.6} & \textbf{45.5} & \textbf{42.0} & 40.7 & 39.6 & 66.2 & 34.8 & 60.6 & 31.0 & \textbf{68.1} & 49.5 \\
    \textbf{\method{} (Ours)} & 47.1 & 45.3 & 28.3 & \textbf{56.1} & 34.8 & \textbf{75.3} & \textbf{48.2} & \textbf{64.9} & \textbf{42.4} & 60.3 & \textbf{51.4} \\
    \bottomrule
  \end{tabular}
  \end{adjustbox}
\end{table*}

\subsection{Main Results on MER-UniBench}
\label{sec:main-results-mer-unibench}

Table~\ref{tab:mer-unibench-main} shows that \method{} achieves the highest
reported mean in all three input-modality settings: $84.13$ with
audio--video--text, compared with OPPO's reported $81.05$, and $80.40$ and
$79.66$ with audio--text and video--text, exceeding AffectGPT-R1 by $3.16\%$
and $3.71\%$, respectively. With all three modalities,
\method{} leads on six of nine datasets: all four sentiment datasets,
including $90.06$ WAF on CMU-MOSI and $90.76$ on CH-SIMS, as well as MELD
and IEMOCAP, with HIT scores of $71.89$ and $88.59$.
Its $59.52$ EW-F1 on OV-MERD+ still trails
AffectGPT-R1's $68.05$, identifying fine-grained open-vocabulary prediction
as a remaining limitation of the overall training approach.
These results highlight \method{}'s strong sentiment and basic emotion
recognition across modality combinations with a single post-trained checkpoint.

\subsection{Main Results on MME-Emotion}
\label{sec:main-results-mme-emotion}

Table~\ref{tab:mme-emotion-main} shows that \method{} achieves the highest
Rec-S ($42.4\%$) among all compared models and the highest CoT-S ($51.4\%$)
among the compared open-source audio--video--text models.
Relative to Qwen2.5-Omni, its gains primarily reflect more accurate final
judgments while preserving reasoning quality.
Compared with R1-Omni and AffectGPT, it achieves higher Rec-S and Rea-S,
with higher CoT-S on seven of eight tasks.
Its overall CoT-S also approaches that of GPT-4o.
Among all compared models, \method{} achieves the highest CoT-S on
fine-grained sentiment analysis ($56.1\%$), noise-robust emotion recognition
($75.3\%$), and sentiment analysis ($64.9\%$).
Together, these results highlight \method{}'s strengths in both fine-grained
sentiment understanding and robust emotion recognition, demonstrating its
effectiveness across diverse multimodal affective tasks.

\subsection{Ablation and Training Analysis}
\label{sec:ablation-study}

\paragraph{Post-training effectiveness and transfer.}
Table~\ref{tab:ablation-domain-ood} shows that \method{} outperforms SFT
and GRPO in mean ID and OOD macro-F1 at both model scales.
Its ID means reach $0.4021$ (3B) and $0.4146$ (7B),
improving over the base models by $4.80\%$ and $4.93\%$,
with the best scores in five of the six ID comparisons.
Among post-training methods, it leads on three of four OOD tasks
per scale, exceeding
SFT's mean by $1.80\%$ and $1.50\%$ and GRPO's by $1.30\%$ and $0.33\%$,
respectively. SFT remains competitive in-domain but yields the lowest OOD
mean among the post-training methods. Both on-policy methods transfer
better, with \method{} outperforming GRPO.
In this comparison, \method{} offers the strongest combination of ID
adaptation and OOD retention, supporting the value of dense teacher guidance
on student-generated trajectories.

\paragraph{Inference-time reasoning and on-policy sampling.}
Table~\ref{tab:thinking-mode-ablation} holds model parameters fixed within
each method and varies inference mode. Thinking improves \method{}'s
ID/OOD means by $2.51\%/1.72\%$, whereas both means decrease slightly for GRPO.
AffectGPT-R1's separately trained answer-only variant also outperforms
its reasoning variant on OV-MERD+ ($68.39\%$ vs.\ $66.49\%$)
\citep{lian2025affectgptr1}.
Despite different protocols, both comparisons suggest that explicit
reasoning alone need not improve prediction. GRPO's sparse outcome
rewards may leave intermediate interpretations weakly
constrained, allowing hallucinated affective cues to mislead final
predictions. \method{} addresses this supervision gap through dense
token-level targets from a rationale-informed teacher on student-generated
prefixes, guiding intermediate reasoning even when the final answer is wrong.
Table~\ref{tab:prefix-refresh-summary} further shows that online prefix
refresh exceeds both cached controls in ID and OOD means, leading on
all three ID tasks and two of four OOD tasks
(Appendix~\ref{app:additional-experiments},
Tables~\ref{tab:app-thinking-mode-details} and~\ref{tab:app-prefix-refresh}).
Together, these results support \method{}'s complementary strengths:
on-policy sampling for stronger OOD generalization and dense teacher
supervision for evidence-informed reasoning.

\begin{table*}[t]
  \centering
  \widetablesetup
  \caption{\textbf{Ablation results across ID and OOD affective benchmarks.}
  The left block reports ID results after joint training on
  MELD, MIntRec~2.0, and IEMOCAP, while the right block reports OOD
  generalization to Mandarin conversations on MC-EIU and to
  single-label and multi-label emotion recognition on MAFW.
  Bold numbers indicate the best result within each base-model block, and underlining indicates the second-best result.}
  \label{tab:ablation-domain-ood}
  \begin{adjustbox}{max width=\textwidth,center}
  \begin{tabular}{@{}llccccccc@{}}
    \toprule
    \multirow[c]{2}{*}[-1.2ex]{\textbf{Base model}}
    & \multirow[c]{2}{*}[-1.2ex]{\textbf{Method}}
    & \multicolumn{3}{c}{\textbf{In-domain (ID)}}
    & \multicolumn{4}{c}{\textbf{Out-of-domain (OOD)}} \\
    \cmidrule(lr){3-5}
    \cmidrule(lr){6-9}
    & & MELD & \shortstack{MIntRec\\2.0} & IEMOCAP
    & \shortstack{MC-EIU\\Intent-ZH}
    & \shortstack{MC-EIU\\Emotion-ZH}
    & \shortstack{MAFW\\Emotion-SL}
    & \shortstack{MAFW\\Emotion-ML}
    \\
    \midrule
    \multirow{4}{*}{Qwen2.5-Omni-3B} & Base
    & 0.5941 & 0.2936 & 0.1747
    & \textbf{0.2597} & \underline{0.2747} & \underline{0.2127} & \textbf{0.2636} \\
    & SFT
    & \underline{0.6355} & \underline{0.3161} & \underline{0.2275}
    & 0.2270 & 0.2593 & 0.1920 & \underline{0.2383} \\
    & GRPO
    & 0.6093 & 0.2921 & 0.2208
    & 0.2352 & 0.2670 & 0.2048 & 0.2298 \\
    & \textbf{\method{}}
    & \textbf{0.6427} & \textbf{0.3248} & \textbf{0.2388}
    & \underline{0.2551} & \textbf{0.2834} & \textbf{0.2159} & 0.2343 \\
    \midrule
    \multirow{4}{*}{Qwen2.5-Omni-7B} & Base
    & 0.6051 & 0.3007 & 0.1900
    & \textbf{0.2539} & \textbf{0.2839} & \textbf{0.2453} & \textbf{0.3164} \\
    & SFT
    & \underline{0.6451} & \textbf{0.3581} & \underline{0.2343}
    & 0.2214 & 0.2450 & 0.1965 & 0.2330 \\
    & GRPO
    & 0.6424 & 0.3179 & 0.2267
    & 0.2196 & \underline{0.2607} & 0.2194 & 0.2431 \\
    & \textbf{\method{}}
    & \textbf{0.6501} & \underline{0.3510} & \textbf{0.2426}
    & \underline{0.2248} & 0.2360 & \underline{0.2250} & \underline{0.2700} \\
    \bottomrule
  \end{tabular}
  \end{adjustbox}
\end{table*}

\begin{table}[t]
  \centering
  \begin{minipage}[t]{0.485\linewidth}
    \vspace{0pt}
    \centering
    \standardtablesetup
    \setlength{\tabcolsep}{3pt}
    \captionof{table}{\textbf{Effect of inference-time thinking.}
    We report mean ID and OOD macro-F1 for Qwen2.5-Omni-3B
    with and without thinking.}
    \label{tab:thinking-mode-ablation}
    \begin{tabular*}{\linewidth}{@{\extracolsep{\fill}}llcc@{}}
      \toprule
      \textbf{Method} & \textbf{Mode} & \textbf{ID Avg.} & \textbf{OOD Avg.} \\
      \midrule
      \multirow[c]{2}{*}{GRPO} & w/o Think & 0.3752 & 0.2347 \\
           & w/ Think  & 0.3741 & 0.2342 \\
      \midrule
      \multirow[c]{2}{*}{\method{}} & w/o Think & 0.3770 & 0.2300 \\
                & w/ Think  & \textbf{0.4021} & \textbf{0.2472} \\
      \bottomrule
    \end{tabular*}
  \end{minipage}\hfill
  \begin{minipage}[t]{0.485\linewidth}
    \vspace{0pt}
    \centering
    \standardtablesetup
    \setlength{\tabcolsep}{3pt}
    \captionof{table}{\textbf{Effect of on-policy prefix refresh.}
    We report mean ID and OOD macro-F1 for Qwen2.5-Omni-3B
    trained with cached student prefixes or online prefix refresh.}
    \label{tab:prefix-refresh-summary}
    \begin{tabular*}{\linewidth}{@{\extracolsep{\fill}}lcc@{}}
      \toprule
      \textbf{Prefix source} & \textbf{ID Avg.} & \textbf{OOD Avg.} \\
      \midrule
      Single cached prefix & 0.3853 & 0.2394 \\
      Frozen prefix pool & 0.3855 & 0.2331 \\
      Online refresh & \textbf{0.4099} & \textbf{0.2438} \\
      \bottomrule
    \end{tabular*}
  \end{minipage}
\end{table}

\begin{figure}[!t]
  \centering
  \begin{minipage}[t]{0.35\linewidth}
    \vspace{0pt}
    \centering
    \includegraphics[width=\linewidth]{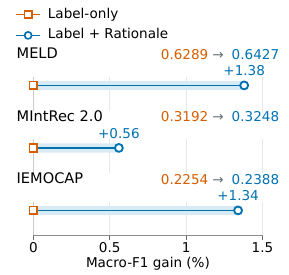}
    \captionof{figure}{\RaggedRight\mbox{\textbf{Rationale supervision.}}
    \mbox{Macro-F1} gains (\%) over \mbox{Label-only} for separately
    trained \mbox{\method{}} (3B) models.}
    \label{fig:rationale-privilege}
  \end{minipage}\hfill
  \begin{minipage}[t]{0.62\linewidth}
    \vspace{0pt}
    \centering
    \includegraphics[width=\linewidth]{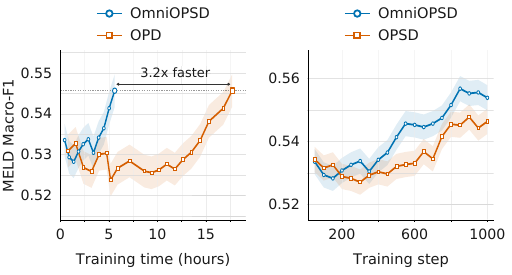}
    \captionof{figure}{\RaggedRight\textbf{Comparison with OPD and OPSD on MELD.}
    Left: \method{} versus OPD over training time, with a dashed line at
    OPD's final macro-F1 of $0.546$. Right: \method{} versus OPSD by
    optimizer step.}
    \label{fig:meld-opd-opsd}
  \end{minipage}
\end{figure}

\paragraph{Rationale-privileged self-distillation and efficiency.}
Figure~\ref{fig:rationale-privilege} compares separately trained models
under matched protocols. Adding rationales to the teacher's label context
improves macro-F1 by $1.38\%$, $0.56\%$, and $1.34\%$
on MELD, MIntRec~2.0, and IEMOCAP
(Table~\ref{tab:app-rationale-privilege}), indicating that rationales
supply useful affective evidence beyond the annotated answers.
Figure~\ref{fig:meld-opd-opsd} evaluates a separate configuration from
Table~\ref{tab:ablation-domain-ood}. In the left panel, \method{} reaches
$0.546$ MELD macro-F1 in $5.6$ hours, versus $17.7$ hours for OPD,
using approximately $68\%$ less training time in this run.
With rationales cached offline, local teacher scoring avoids frontier-model
logits and online frontier-model calls during optimization, reducing
teacher-inference overhead.
In the right panel, \method{} exceeds the OPSD baseline without
frontier-generated rationales at every evaluation from step $200$ onward,
finishing at $0.554$ versus $0.546$.
This improvement supports the combined training design in
Eq.~\eqref{eq:final-loss}: dense self-distillation targets guide intermediate
reasoning, while reinforcement learning rewards final-answer accuracy
and format compliance. Together, these comparisons support efficient
self-distillation with complementary process and outcome supervision.

\begin{figure*}[!t]
  \centering
  \includegraphics[width=\textwidth]{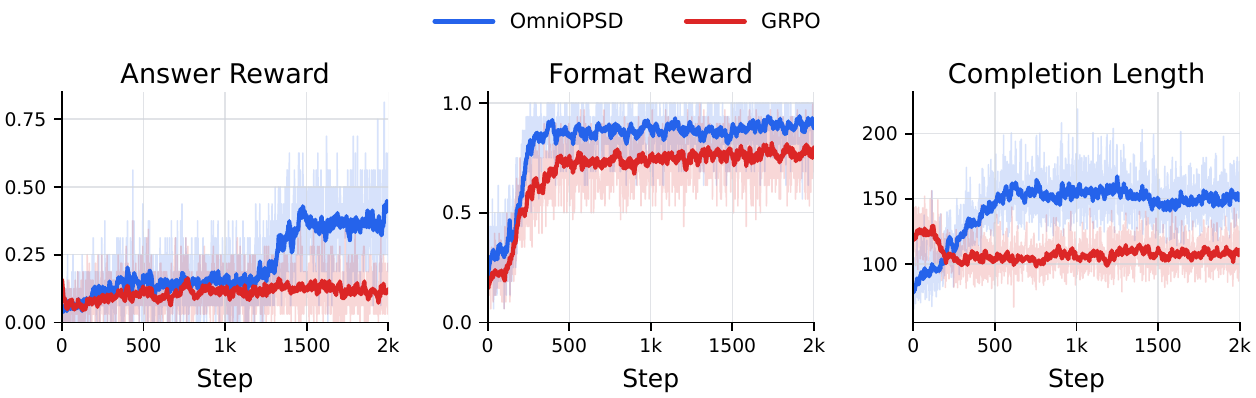}
  \caption{\textbf{Training dynamics of GRPO and \method{} on Qwen2.5-Omni-3B.}
  Three panels track the answer-accuracy reward, the format reward, and the completion length over training.}
  \label{fig:grpo-vs-omniopsd-3b-dynamics}
\end{figure*}

\begin{figure}[!t]
  \centering
  \setlength{\abovecaptionskip}{7pt}
  \includegraphics[width=\textwidth]{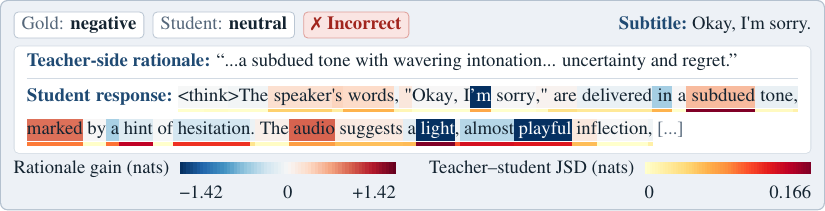}
  \caption{\textbf{Affect-related guidance on an incorrect MELD rollout.}
  Background colors show log-probability changes from adding the rationale
  to the label-only teacher context, while aligned strips show full-vocabulary
  teacher--student JSD on the same student prefixes
  (full response: Figure~\ref{fig:app-token-meld-incorrect}).
  This illustrates \method{}'s dense token-level supervision: the teacher
  favors affective cues consistent with its rationale and suppresses
  conflicting interpretations, even on an incorrect rollout.}
  \label{fig:dense-affective-guidance}
\end{figure}

\paragraph{Training dynamics.}
Figure~\ref{fig:grpo-vs-omniopsd-3b-dynamics} shows \method{}'s sharp
late-stage answer-reward gains (left) and consistently higher smoothed
format rewards than GRPO (middle). Completion length (right) increases
and stabilizes near $150$ tokens for \method{}, whereas GRPO's rises
slightly before declining toward $100$--$110$. These trends suggest that
\method{} encourages thinking through dense teacher guidance while
improving answer correctness and format adherence.

\paragraph{Dense affective guidance and qualitative cases.}
Figure~\ref{fig:dense-affective-guidance} shows an incorrect MELD rollout re-scored
with checkpoint, labels, and student prefixes fixed. Rationale conditioning
raises the probability of \emph{subdued} and decreases those of
\emph{light} and \emph{playful}, consistent with the teacher's affective
account despite the wrong final prediction.
Across 600 held-out responses, the dataset-balanced mean teacher--student
JSD is $0.01973$ for reasoning text and $0.00071$ for format tags
(Appendix~\ref{app:token-guidance}, Table~\ref{tab:app-token-regions}).
Together with the rationale ablation, these diagnostics support affect-related
guidance on student-generated reasoning, including unsuccessful rollouts.
Appendix~\ref{app:case-studies}
(Tables~\ref{tab:app-case-ir}--\ref{tab:app-case-emer-sadness}) provides
author-selected cases illustrating \method{}'s
multimodal affective reasoning strengths.

\section{Conclusion}

We presented \method{}, a rationale-privileged on-policy self-distillation
framework for omni-modal affective reasoning.
\method{} uses offline frontier-generated multimodal evidence descriptions
and annotated answers as teacher-only privileged context.
An EMA teacher provides dense token-level supervision on
student-generated prefixes, while the student uses only the original
task input during rollout and inference.
Results show that \method{} outperforms SFT and GRPO in mean
in-domain and out-of-domain macro-F1 at both 3B and 7B scales.
Among the compared models, it achieves the highest MER-UniBench mean
in all three modality settings and the highest MME-Emotion Rec-S.
Ablations support rationale conditioning and on-policy
prefix refresh.
Future work will address rationale reliability and fine-grained
open-vocabulary emotion recognition.

\subsection*{Ethics statement}

All datasets and their associated labels are obtained
through the required application procedures and used within the scope of
the granted research permissions. Training and evaluation use the provided
audio, video, and text content and their original labels, without altering
the underlying data or annotations. The experiments focus on affective
understanding and do not involve discriminatory profiling or decision-making.
Benchmark performance reflects agreement with context-dependent annotations
rather than direct measurement of individuals' internal mental states.

\subsection*{Reproducibility statement}

Appendices~\ref{app:theory}--\ref{app:algorithm} present the formal
assumptions and training algorithm.
Appendices~\ref{app:rationale-prompts}--\ref{app:experimental-settings}
document the prompts, datasets, and detailed protocols for training
and evaluation.
To support fair comparison, we apply consistent testing procedures
to the methods evaluated in our experiments within each benchmark
and compute metrics according to the corresponding evaluation protocols.
Appendix~\ref{app:additional-experiments} provides supplementary
experimental analyses that complement the main results.

\subsection*{AI use statement}

In this work, we used generative AI tools for retrieval and discovery
(e.g., finding related work), equation and notation checks,
experimental code development, multimodal cue extraction, result organization,
output normalization, benchmark evaluation, and grammar and spelling edits.
The authors take full responsibility for the manuscript, code, experiments,
and all reported results and scientific claims.

\bibliographystyle{iclr2027_conference}
\bibliography{rec}

\appendix
\renewcommand{\theHfigure}{appendix.\Alph{section}.\arabic{figure}}
\renewcommand{\theHtable}{appendix.\Alph{section}.\arabic{table}}
\begingroup
\hypersetup{pdfborder={0 0 0}}
\noindent\textbf{Appendix overview.}
The appendices provide the mathematical foundations, experimental protocols,
and supporting analyses of \method{}.
\hyperref[app:theory]{Appendix~\ref*{app:theory}} derives the properties and
gradients of the generalized JSD objective and clarifies the on-policy
semi-gradient and local token normalization.
\hyperref[app:algorithm]{Appendix~\ref*{app:algorithm}} presents the training
algorithm, optional reward surrogate, and implementation details for completion
alignment and EMA teacher updates.
\hyperref[app:rationale-prompts]{Appendix~\ref*{app:rationale-prompts}}
describes the construction of teacher-side privileged rationales and specifies
the student, teacher, and evaluation prompts and output formats.
\hyperref[app:datasets-benchmarks]{Appendix~\ref*{app:datasets-benchmarks}}
documents data sources, evaluation subsets, and scoring protocols for the
controlled ablations, MER-UniBench, and MME-Emotion.
\hyperref[app:experimental-settings]{Appendix~\ref*{app:experimental-settings}}
details the training configurations, implementation, and evaluation settings
for the main benchmarks and controlled ablations.
\hyperref[app:additional-experiments]{Appendix~\ref*{app:additional-experiments}}
presents additional experiments, including an analysis of MME-Emotion scores
and response length, as well as studies of teacher-side rationale conditioning,
inference-time thinking, and on-policy prefix refresh.
\hyperref[app:visualizations]{Appendix~\ref*{app:visualizations}} uses
controlled re-scoring of fixed student responses to characterize token-level
teacher guidance and presents qualitative cases of multimodal affective reasoning.
\par
\endgroup

\section{Theoretical Support}
\label{app:theory}

This appendix provides theoretical support for the token-normalized
distillation objective. Section~\ref{app:theory-jsd} establishes the bounds,
equality conditions, and gradients of the generalized JSD, while
Section~\ref{app:theory-on-policy} characterizes the on-policy semi-gradient
and clarifies the consequences of local valid-token normalization.

\subsection{JSD Properties and Gradients}
\label{app:theory-jsd}

Following the notation of Section~\ref{sec:method}, let $p=P_{S,t}^{(i)}(\theta)$
and $q=\widetilde P_{T,t}^{(i)}$ denote the student and stop-gradient teacher
distributions over $\mathcal{V}$ at a fixed completion prefix, using natural
logarithms throughout. For $\beta\in(0,1)$, define the mixture
\begin{equation}
 p_\beta=(1-\beta)p+\beta q,
 \label{eq:jsd-mixture}
\end{equation}
and the generalized JSD
\begin{equation}
 D_{\mathrm{JSD}}^{\beta}(p,q)
 =(1-\beta)\operatorname{KL}(p\,\|\,p_\beta)
 +\beta\operatorname{KL}(q\,\|\,p_\beta).
 \label{eq:generalized-jsd}
\end{equation}
This divergence satisfies
\begin{equation}
 0\leq D_{\mathrm{JSD}}^{\beta}(p,q)\leq h(\beta),
 \qquad
 h(\beta)=-\beta\log\beta-(1-\beta)\log(1-\beta),
 \label{eq:app-jsd-bound}
\end{equation}
with equality at the lower bound if and only if $p=q$, where non-negativity and
the equality condition follow from KL divergence.  For the upper bound, introduce
a binary variable $C\sim\operatorname{Bernoulli}(\beta)$ and sample a token $V$
from $p$ when $C=0$ and from $q$ when $C=1$, so that
$D_{\mathrm{JSD}}^{\beta}(p,q)=I(C;V)\leq H(C)=h(\beta)$.
The same bound applies to the valid-token mean in
Eq.~\eqref{eq:distill-loss}, with an upper bound of $\log 2$ at $\beta=0.5$.

Differentiating with $q$ fixed and including both occurrences of $p_\beta$ gives
\begin{equation}
 \frac{\partial D_{\mathrm{JSD}}^{\beta}(p,q)}{\partial p_v}
 =(1-\beta)\log\frac{p_v}{p_{\beta,v}}.
 \label{eq:app-jsd-prob-gradient}
\end{equation}
The non-logarithmic terms cancel because $[(1-\beta)p_v+\beta q_v]/p_{\beta,v}=1$, and
applying the chain rule to the student logits $u=u_{\theta}(c_i^S,y_{i,<t})$
with $p=\operatorname{softmax}(u/\tau)$ then yields
\begin{equation}
 \frac{\partial D_{\mathrm{JSD}}^{\beta}}{\partial u_j}
 =\frac{1-\beta}{\tau}p_j
 \left[
   \log\frac{p_j}{p_{\beta,j}}
   -\sum_{v\in\mathcal{V}}p_v\log\frac{p_v}{p_{\beta,v}}
 \right].
 \label{eq:app-jsd-logit-gradient}
\end{equation}
The mixture remains differentiable through the student while the teacher
distribution is detached, and the standard implementation applies no additional
$\tau^2$ factor.  For positive softmax distributions, the logit gradient is zero
if and only if $p=q$: a zero gradient requires $p_v/p_{\beta,v}$ to be constant
across the vocabulary, and normalization then implies $p=p_\beta=q$.
Distillation can therefore provide a token-level signal even when an optional
sequence reward is zero, although this gradient property does not guarantee
that the teacher's preferences improve task accuracy.

\subsection{On-Policy Optimization}
\label{app:theory-on-policy}

At optimizer step $k$, completions are sampled from
$\pi_{\theta_k,G}(\cdot\mid c_i^S)$ under the decoding configuration $G$ and
re-scored by both branches, with the teacher additionally conditioned on the
privileged rationale.  Holding the sampled tokens, completion
masks, and teacher targets fixed, the computed gradient for a local forward
batch $\mathcal{B}$, with token divergence $d_{i,t}(\theta)$ and
$Z_{\mathcal{B}}=\sum_{i\in\mathcal{B}}\sum_t M_{i,t}>0$, is
\begin{equation}
 \widehat g_k
 =\left.
   \frac{1}{Z_{\mathcal{B}}}
   \sum_{i\in\mathcal{B}}\sum_t
   M_{i,t}\nabla_\theta d_{i,t}(\theta)
  \right|_{\theta=\theta_k}.
 \label{eq:app-on-policy-gradient}
\end{equation}
Because the probability of sampling each prefix is held fixed,
Eq.~\eqref{eq:app-on-policy-gradient} is a semi-gradient through the re-scored
student distributions on states visited by the current rollout policy, rather
than the full derivative of an objective whose sampling distribution varies
with $\theta$.

Within $\mathcal{B}$, each retained token receives the same coefficient
$1/Z_{\mathcal{B}}$, giving a mean over valid completion tokens in which longer
retained completions contribute more terms.  This generally differs from an
average of independently length-normalized sequence losses, although both
reduce to the response's token mean when the local batch contains one
completion.  Writing
$N_{\mathcal{B}}(\theta)=\sum_{i\in\mathcal{B}}\sum_t M_{i,t}d_{i,t}(\theta)$,
the expected finite-batch loss, with the expectation over batch selection and
student rollouts at fixed $\theta_k$, is
\begin{equation}
 \mathbb{E}_{\mathcal{B}}
 \left[\widehat{\mathcal{L}}_D(\theta;\mathcal{B})\right]
 =\mathbb{E}_{\mathcal{B}}
 \left[\frac{N_{\mathcal{B}}(\theta)}{Z_{\mathcal{B}}}\right]
 \neq
 \frac{\mathbb{E}_{\mathcal{B}}[N_{\mathcal{B}}(\theta)]}
      {\mathbb{E}_{\mathcal{B}}[Z_{\mathcal{B}}]}.
 \label{eq:app-finite-batch-reduction}
\end{equation}
The ratio of expectations describes a population token-weighted objective and
generally differs from the expected local mean implemented in
Eq.~\eqref{eq:distill-loss} when batch token counts vary.

\section{OmniOPSD Algorithm and Implementation Details}
\label{app:algorithm}

This appendix translates the objective in Section~\ref{sec:method} into the
complete training procedure. Section~\ref{app:algorithm-pseudocode} presents
the EMA-teacher algorithm, Section~\ref{app:algorithm-reward} defines the
optional reward surrogate, and Section~\ref{app:algorithm-implementation}
documents completion alignment, loss reduction, and teacher-state handling.

Following Section~\ref{sec:method}, $\mathcal{B}$ denotes the local batch
processed by one standard logits forward pass, and $M_{i,t}\in\{0,1\}$ marks
valid completion positions, with prompt and padding positions assigned zero.
Completion $i$ contains $T_i=\sum_t M_{i,t}$ valid tokens, and
$Z_{\mathcal{B}}=\sum_{i\in\mathcal{B}}T_i$ counts the valid tokens in $\mathcal{B}$.

\subsection{Training Algorithm}
\label{app:algorithm-pseudocode}

Algorithm~\ref{alg:omniopsd} uses an EMA teacher with symmetric JSD
($\beta=0.5$) and distillation temperature $\tau=1$, sampling student responses
under decoding configuration $G$ and re-scoring their prefixes to obtain
full-vocabulary distributions for distillation.

The sampled completion is factorized autoregressively under the decoding configuration $G$:
\begin{equation}
 y_i\sim\pi_{\theta_k,G}(\cdot\mid c_i^S),\qquad
 \pi_{\theta_k,G}(y_i\mid c_i^S)
 =\prod_{t=1}^{L_i}
 \pi_{\theta_k,G}(y_{i,t}\mid c_i^S,y_{i,<t}).
 \label{eq:sequence-policy}
\end{equation}

\begin{algorithm}[!htbp]
\caption{\method{} with an EMA Teacher}
\label{alg:omniopsd}
\begin{algorithmic}[1]
\Require Dataset $\mathcal{D}=\{(o_i,q_i,a_i,r_i)\}$, student parameters
$\theta_0$, EMA state $\bar\theta\leftarrow\varnothing$, decay $\mu$,
decoding configuration $G$, and reward weight $\alpha\geq0$.
\For{each optimizer step $k=0,1,\ldots$}
  \State Initialize the local accumulated gradient to zero.
  \For{each local forward batch $\mathcal{B}$ in the accumulation window}
    \State Construct $c_i^S=\operatorname{Prompt}_S(o_i,q_i)$ and
    $c_i^T=\operatorname{Prompt}_T(o_i,q_i,r_i)$ for $i\in\mathcal{B}$.
    \State Sample $y_i\sim\pi_{\theta_k,G}(\cdot\mid c_i^S)$ and
    detach the sampled token IDs.
    \State Encode $(c_i^S,y_i)$ and $(c_i^T,y_i)$, then select and align their
    valid completion tokens using $M_{i,t}$.
    \State Re-score the student completion with gradients to obtain
    $P_{S,t}^{(i)}(\theta)$, evaluated at $\theta=\theta_k$.
    \State Set $\widetilde\theta_k\leftarrow\theta_k$ if
    $\bar\theta=\varnothing$, and $\widetilde\theta_k\leftarrow\bar\theta_k$
    otherwise.
    \State Under \textsc{NoGrad}, re-score the same completion with
    $\widetilde\theta_k$ and $c_i^T$ to obtain
    $\widetilde P_{T,t}^{(i)}$.
    \State Compute $\widehat{\mathcal L}_D(\theta;\mathcal B)$ by the valid-token
    mean in Eq.~\eqref{eq:distill-loss}.
    \If{$\alpha>0$}
      \State Score the completion rewards and construct detached advantages
      $A_i$ using Eq.~\eqref{eq:app-reward-advantage}.
      \State Compute $\widehat{\mathcal L}_R(\theta;\mathcal B)$ using
      Eq.~\eqref{eq:app-implemented-reward-loss}.
    \Else
      \State $\widehat{\mathcal L}_R(\theta;\mathcal B)\leftarrow0$.
    \EndIf
    \State Accumulate the gradient of
    $\widehat{\mathcal L}_D(\theta;\mathcal B)+\alpha\widehat{\mathcal L}_R(\theta;\mathcal B)$
    with the trainer's accumulation scaling.
  \EndFor
  \State Average synchronized gradients across workers and update the
  student to $\theta_{k+1}$.
  \State Set $\bar\theta_{k+1}\leftarrow\theta_{k+1}$ if
  $\bar\theta=\varnothing$ and otherwise apply Eq.~\eqref{eq:ema-teacher}.
\EndFor
\State \Return the student $\pi_\theta$.
\end{algorithmic}
\end{algorithm}

The reduction in Eq.~\eqref{eq:distill-loss} first sums the divergence over
the vocabulary and then averages over the $Z_{\mathcal B}$ valid completion
positions, reducing to a single response's token mean when the local forward
pass contains one example.  Gradient accumulation and distributed
synchronization combine these local losses with the trainer's gradient scaling,
without replacing each $Z_{\mathcal B}$ with a token count pooled over the
optimizer step, so the result generally differs from a global token mean when
local valid-token counts vary.

\subsection{Optional Reward Objective}
\label{app:algorithm-reward}

We convert the weighted sum $R_i$ of answer-accuracy and output-format
rewards in Eq.~\eqref{eq:reward} into a fixed advantage for the current update:
\begin{equation}
 A_i=\operatorname{stopgrad}\!\left(\frac{R_i-b}{s}\right).
 \label{eq:app-reward-advantage}
\end{equation}
The resolved experimental configuration selects the zero, batch-mean, or
running-mean baseline $b$ and the scaling rule, with statistics computed over
the reward-scoring batch, which may span workers.  The scale $s$ is one when
scaling is disabled or the reward variance is zero, and otherwise equals the
empirical reward standard deviation plus $10^{-4}$.

With $\ell_{i,t}(\theta)=\log P_{S,t}^{(i)}(\theta)[y_{i,t}]$ and
$\ell_{i,t}^{\mathrm{old}}=\operatorname{stopgrad}(\ell_{i,t}(\theta))$,
the implementation uses the ratio surrogate
\begin{align}
 \rho_{i,t}(\theta)
 &=\exp\!\left(\ell_{i,t}(\theta)-\ell_{i,t}^{\mathrm{old}}\right),
 \\
 \widehat{\mathcal L}_R(\theta;\mathcal B)
 &=-\frac{1}{|\mathcal B|}\sum_{i\in\mathcal B}
 \frac{\sum_t M_{i,t}\min\!\left(
 \rho_{i,t}A_i,\,
 \operatorname{clip}(\rho_{i,t},1-\epsilon_-,1+\epsilon_+)A_i
 \right)}{\max(1,T_i)}.
 \label{eq:app-implemented-reward-loss}
\end{align}
The denominator guard assigns zero contribution to empty completions, while
detaching the old log probabilities from the same forward pass makes
$\rho_{i,t}=1$ numerically.  For positive clipping radii $\epsilon_-,\epsilon_+$,
clipping is locally inactive, so the gradient of
Eq.~\eqref{eq:app-implemented-reward-loss} equals the gradient of
\begin{equation}
 -\frac{1}{|\mathcal B|}\sum_{i\in\mathcal B}
 \frac{A_i}{\max(1,T_i)}
 \sum_t M_{i,t}\ell_{i,t}(\theta).
 \label{eq:app-reward-gradient-surrogate}
\end{equation}
This equivalence holds locally for gradients with the sampled tokens, lengths,
masks, and advantages fixed, although the two loss values need not agree.  Length
normalization generally changes the objective relative to maximizing the
unnormalized expected sequence reward, and a score-function interpretation
additionally requires the re-scoring and rollout distributions to match, which
decoding transformations such as top-$p$ filtering can violate.  We therefore
use this term as a reward surrogate in the combined objective of
Eq.~\eqref{eq:final-loss}.

\subsection{Implementation Details}
\label{app:algorithm-implementation}

\paragraph{Aligned completion scoring.}
\label{app:algorithm-alignment}
With a shared tokenizer and vocabulary, the student and teacher receive the
same generated response token IDs inserted into their separately encoded
prompts, whose lengths can differ.  We shift logits and labels by one position
for next-token prediction and exclude prompt and padding positions with labels
equal to $-100$, checking the remaining student and teacher token IDs for equal
counts and identical ordering before comparing their distributions. These
comparisons use full-vocabulary distributions, independently of any top-$p$ or
top-$k$ filtering in the decoding configuration $G$.  Only the
student forward retains gradients: teacher scoring runs under
\texttt{torch.no\_grad()} with gradient checkpointing disabled for that pass.

The standard full-vocabulary logits path uses the local reduction in
Eq.~\eqref{eq:distill-loss}, and when the multimodal fallback splits an input batch
into chunks, each chunk uses the same reduction, and chunk losses are combined
in proportion to their numbers of examples.

\paragraph{EMA parameter state.}
\label{app:algorithm-ema}
The teacher state stores detached copies of trainable parameters and is
initialized lazily: teacher scoring uses the current student parameters until
the first synchronized optimizer step, after which the updated student
parameters are copied to initialize $\bar\theta$.  After gradient accumulation,
subsequent updates follow Eq.~\eqref{eq:ema-teacher} once per optimizer step.

During teacher scoring, the trainer temporarily swaps the EMA values into the
local model and restores the current student values afterward, so a single
local model instance serves as both student and teacher.  In LoRA
training, the EMA therefore covers the trainable adapters while the frozen
backbone is shared, whereas in full-parameter training it covers all parameters marked
trainable, while frozen modules such as the visual encoder and multimodal
aligner remain shared.  Alongside the EMA configuration in Algorithm~\ref{alg:omniopsd},
the implementation supports an online stop-gradient teacher that uses the
current student parameters on every scoring pass without maintaining an EMA
state, and both configurations use only the student and its original omni-modal
task prompt at inference.

\section{Privileged Rationale Construction and Prompt Templates}
\label{app:rationale-prompts}

This appendix specifies how privileged teacher context is constructed and how
it differs from student and evaluation inputs. Section~\ref{app:rationale-source}
defines the rationale components and serialization, Section~\ref{app:student-prompts}
gives the student and teacher training prompts, and
Section~\ref{app:evaluation-prompts} records the inference prompts and output
formats used for evaluation.

\subsection{Rationale Construction}
\label{app:rationale-source}
\label{app:rationale-quality}

For each MMEVerse training example $i$, let $a_i$ denote its annotated or
consensus answer and let $e_i$ denote the evidence-aware reasoning text
generated offline by GPT-4o. We serialize these components to form the
complete privileged rationale $r_i$:
\begin{equation}
  \begin{aligned}
  r_i &= \operatorname{Serialize}(e_i,a_i) \\
      &= \texttt{<think>}\,e_i\,\texttt{</think>}
         \texttt{<answer>}\,a_i\,\texttt{</answer>}.
  \end{aligned}
  \label{eq:serialized-reference-solution}
\end{equation}
During \method{} updates, the precomputed rationale conditions only the
teacher, while the student receives only the omni-modal task input. The
rationale is therefore neither included in the student rollout prompt nor
used as a supervised target sequence during these updates.

\subsection{Student and Teacher Training Prompts}
\label{app:student-prompts}
\label{app:teacher-prompt}

\paragraph{Student context.}
The system instruction below accompanies user messages containing the
omni-modal input, the subtitle when available, and a task instruction.
Preprocessing randomly selects from the first ten valid templates for the
requested task and modality, or from all valid templates if fewer than ten
are available, and shuffles the candidate labels.

\begin{promptbox}{System instruction}
You are an Affective Computing expert specializing in emotion analysis,
sentiment recognition, and intent recognition.

Core Directives:

- Comprehensive Analysis: Synthesize semantic information, pragmatic cues,
and any implied multimodal features (e.g., tone, facial expressions, action
cues) in your evaluation.

- Modality Honesty: NEVER analyze or make assumptions about missing or
unusable modalities.

- Structured Workflow: You MUST strictly adhere to the user's requested
output format. Always output your step-by-step reasoning first, followed by
your final answer.
\end{promptbox}

The representative templates below share a subtitle prefix when a transcript
is available, with braces marking example-specific fields:
\texttt{\{answers\}} and \texttt{\{modalities\}} specify the candidate labels
and modality description, and \texttt{\{target\}} denotes \texttt{sentiment}
or \texttt{speaker's intent}.

\begin{promptbox}{Student task templates}
\textbf{Shared prefix}

The subtitle of this video is: \{subtitle\}

\textbf{Emotion recognition}

Choose from: \{answers\}. Based on \{modalities\}, reason through your
interpretation in <think></think> and write your final answer in
<answer></answer>.

\textbf{Sentiment and intent recognition}

Choose from: \{answers\}. Based on \{modalities\}, explain your reasoning in
<think></think> and provide the \{target\} in <answer></answer>.

\textbf{Multi-label emotion recognition}

Choose one or more applicable emotions from: \{answers\}. Based on
\{modalities\}, reason through your interpretation in <think></think> and
write your final answer(s) in <answer></answer>. If multiple emotions apply,
separate them with a comma.
\end{promptbox}

IntentQA uses the question and its five options directly:
\begin{promptbox}{IntentQA instruction}
Question: \{question and options\} Based on the video content, explain your
reasoning in <think></think> and output only the letter (A/B/C/D/E) of the
correct option in <answer></answer>.
\end{promptbox}

\paragraph{Teacher context.}
The teacher retains the student's system instruction and omni-modal user
message, appending the suffix below to the last user message with
\texttt{\{solution\}} set to $r_i$ in
Eq.~\eqref{eq:serialized-reference-solution}.

\begin{promptbox}{Privileged teacher suffix}
Below is a reference answer for the same task.

Reference answer:\par
\{solution\}

Study the reference answer carefully. Then solve the original task again in
your own words while preserving the original output format requirements.
\end{promptbox}

The teacher uses this prompt to re-score the student-generated prefixes
without generating a separate response (Appendix~\ref{app:algorithm}).

\subsection{Evaluation Prompts and Output Format}
\label{app:evaluation-prompts}

At evaluation time, the model receives the task input without the privileged
rationale and can produce either a direct label or reasoning followed by a
tagged answer. In the single-label templates below, \texttt{\{target\}}
denotes \texttt{sentiment}, \texttt{emotion}, or \texttt{speaker's intent}, with
dataset-specific variants supplying the corresponding label spaces and task wording.

\begin{promptbox}{Single-label evaluation templates}
\textbf{Direct prediction (only\_label)}

Given the visual, auditory, and spoken content in the video, what is the most
appropriate \{task\} label? Choose from: \{label space\}. Output only the label.

\textbf{Reasoning and answer (think)}\par\nopagebreak[4]

Choose from: \{label space\}. Based on the speaker's expressions, vocal tone,
and spoken content, explain your reasoning in <think></think> and provide the
\{target\} in <answer></answer>.
\end{promptbox}

Multi-label instructions request applicable emotions separated by semicolons,
and open-vocabulary variants omit the candidate list. The evaluator applies
task-specific answer extraction and normalization, with an optional
API-assisted step for outputs that fail local parsing. Evaluation settings and
metrics are described in
Appendices~\ref{app:datasets-benchmarks} and~\ref{app:experimental-settings}.

\suppressfloats[t]

\section{Dataset and Benchmark Details}
\label{app:datasets-benchmarks}

This appendix documents the data and evaluation protocols underlying the
reported results. Section~\ref{app:training-data-composition} describes the
source datasets, prepared evaluation inventories, and controlled-ablation
splits, while Sections~\ref{app:mer-unibench-details} and
\ref{app:mme-emotion-details} then specify the benchmark subsets, metrics, and
evaluation procedures for MER-UniBench and MME-Emotion, respectively. We first
provide the provenance of the models compared in
Tables~\ref{tab:mer-unibench-main} and~\ref{tab:mme-emotion-main}.

\paragraph{Compared models.}
The external baselines in the MER-UniBench comparison comprise
OneLLM~\citep{han2024onellm}, SECap~\citep{xu2024secap},
PandaGPT~\citep{su2023pandagpt}, Qwen-Audio~\citep{chu2023qwenaudio},
SALMONN~\citep{tang2024salmonn}, Otter~\citep{li2023otter},
Video-LLaVA~\citep{lin2024videollava},
Video-ChatGPT~\citep{maaz2024videochatgpt},
VideoChat2~\citep{li2024mvbench}, LLaMA-VID~\citep{li2024llamavid},
VideoChat~\citep{li2023videochat}, Chat-UniVi~\citep{jin2024chatunivi},
mPLUG-Owl~\citep{ye2023mplugowl},
Emotion-LLaMA~\citep{cheng2024emotion},
Emotion-LLaMAv2~\citep{peng2026emotionllamav2},
R1-Omni~\citep{zhao2025r1}, AffectGPT~\citep{lian2025affectgpt},
AffectGPT-R1~\citep{lian2025affectgptr1}, and OPPO~\citep{han2026omni}.
Our model is initialized from Qwen2.5-Omni-7B~\citep{xu2025qwen25omni}.

The MME-Emotion comparison additionally includes
GPT-4o~\citep{openai2024gpt4o},
GPT-4.1\footnote{Official model page: \url{https://openai.com/index/gpt-4-1/}},
Gemini-2.0-Flash\footnote{Official model announcement:
\url{https://blog.google/technology/google-deepmind/google-gemini-ai-update-december-2024/}},
Gemini-2.5-Flash, and
Gemini-2.5-Pro~\citep{geminiteam2025gemini25},
Qwen2-Audio~\citep{chu2024qwen2audio},
Qwen2-VL-7B~\citep{wang2024qwen2},
Qwen2.5-VL-7B~\citep{bai2025qwen25vl},
Video-LLaMA~\citep{zhang2023videollama},
Video-LLaMA2~\citep{cheng2024videollama2},
Qwen2.5-Omni~\citep{xu2025qwen25omni},
Emotion-LLaMA~\citep{cheng2024emotion},
HumanOmni~\citep{zhao2025humanomni}, R1-Omni~\citep{zhao2025r1},
AffectGPT~\citep{lian2025affectgpt}, and OPPO~\citep{han2026omni}, alongside
Video-LLaVA~\citep{lin2024videollava}.

\subsection{Data Sources and Ablation Protocol}
\label{app:training-data-composition}

\paragraph{IEMOCAP.}
IEMOCAP~\citep{busso2008iemocap} contains English scripted and improvised
dyadic interactions performed by ten actors across five sessions. Its
approximately 12 hours of recordings comprise 10,039 utterances with
speech, video, motion capture, and transcripts. Annotations include nine
named emotion categories and an \emph{Other} option, together with
valence, activation, and dominance ratings.

\paragraph{MELD.}
MELD~\citep{poria2019meld} extends EmotionLines with audio and video from
the English television series \emph{Friends}. It contains 13,708
utterances in 1,433 multiparty dialogues, with seven emotion categories
and three sentiment polarities: positive, negative, and neutral.
The official training, development, and test sets contain 9,989, 1,109,
and 2,610 utterances.

\paragraph{MIntRec 2.0.}
MIntRec~2.0~\citep{zhang2024mintrec} supports multimodal intent recognition
and out-of-scope detection in English conversations from \emph{Superstore},
\emph{The Big Bang Theory}, and \emph{Friends}. It provides text, audio,
video, and speaker annotations for 15,040 utterances in 1,245 dialogues.
The dataset includes 9,304 utterances across 30 in-scope intent classes
and 5,736 out-of-scope utterances. The released in-scope partitions contain
6,165 training, 1,106 validation, and 2,033 test utterances.

\paragraph{MC-EIU.}
MC-EIU~\citep{liu2024emotion} addresses joint emotion and intent
understanding in English and Mandarin television conversations.
Its 56,012 utterances from 4,970 conversations have aligned text, audio,
and video, with seven emotion and nine intent categories.
The Mandarin portion contains 11,003 utterances, divided into 7,643
training, 1,148 validation, and 2,212 test examples while keeping
conversations intact.

\paragraph{IntentQA.}
IntentQA~\citep{li2023intentqa} studies the intentions behind human
actions through video question answering. It selects causal and temporal
questions from NExT-QA and adds action annotations to support reasoning
about why actions occur and how their intentions connect events.
The dataset contains 16,297 question--answer pairs over 4,303 videos,
with 12,119 training, 2,044 validation, and 2,134 test questions.
We use its training split as an additional source of five-way video
question--answer examples during the training process.

\paragraph{MAFW.}
MAFW~\citep{liu2022mafw} contains 10,045 audiovisual clips from films,
television programmes, and short videos, with annotations for eleven
underlying emotions, compound expressions, and emotion-related
descriptions in English and Chinese. The original single-expression
benchmark retains 9,172 clips after annotation validation, while its compound
benchmark distinguishes 43 classes, comprising eleven single emotions
and 32 combinations. The original evaluation uses five-fold cross-validation,
while our multi-label setting represents emotions as an eleven-dimensional
multi-hot vector. The evaluation inventory
contains 1,839 single-label examples and 580 examples with multi-label
annotations.

\paragraph{DFEW.}
DFEW (Dynamic Facial Expression in the Wild)~\citep{jiang2020dfew}
contains 16,372 movie clips with challenging facial expressions,
including occlusions, pose changes, and variations in illumination.
Its seven emotion categories are happiness, sadness, neutrality, anger,
surprise, disgust, and fear. The processed single-label benchmark retains
11,697 clips and uses five-fold cross-validation. Our evaluation
inventory contains the 2,341 clips in its first test fold.

\paragraph{E\textsuperscript{3}.}
$E^3$~\citep{lin20243} studies embodied emotion in multilingual
egocentric videos, with over 50 hours of visual, acoustic, and textual
content in the published collection. It provides annotations for
recognition, temporal localization, and reasoning about emotion.
The eight categories are angry, disgusted, happy, sad, sarcastic,
scared, shy, and surprised. Our recognition inventory comprises 3,288
examples: 1,662 from validation and 1,626 from test.

\paragraph{CAER.}
CAER (Context-Aware Emotion Recognition)~\citep{lee2019context}
contains 13,201 audiovisual clips from 79 television shows, supporting
emotion recognition from both facial expressions and surrounding
context. Its seven labels are anger, disgust, fear, happiness, sadness,
surprise, and neutrality. The original split allocates 70\%, 10\%, and 20\%
to training, validation, and test, respectively. Our prepared evaluation
inventory contains 3,953 clips: 1,316 from validation and 2,637 from test.

\paragraph{BoLD.}
BoLD (Body Language Dataset)~\citep{luo2020arbee} contains 9,876 movie
clips with 13,239 annotated character instances. It provides video,
audio, and estimated body poses for studying bodily expressions of
emotion. Each instance is annotated with 26 non-exclusive emotion
categories and continuous valence, arousal, and dominance scores.

\paragraph{CMU-MOSI.}
CMU-MOSI~\citep{zadeh2016mosi} contains 2,199 English opinion segments
from 93 online videos by 89 speakers. Each segment includes speech,
visual behavior, and a transcript. Sentiment intensity is obtained by
averaging five annotators' ratings on a scale from $-3$ to $+3$,
combining assessments of linguistic and nonverbal sentiment cues.
The evaluation inventory contains 686 test segments.

\paragraph{CMU-MOSEI.}
CMU-MOSEI~\citep{zadeh2018multimodal} expands this setting to 23,453
annotated segments from 3,228 English online videos, covering 1,000
speakers and 250 topics. It provides aligned transcripts and speech,
sentiment ratings in $[-3,3]$, and intensity annotations for happiness,
sadness, anger, fear, disgust, and surprise. MER-UniBench uses its
sentiment annotations, with 4,659 test segments in the evaluation
inventory.

\paragraph{CH-SIMS.}
CH-SIMS~\citep{yu2020ch} comprises 2,281 Chinese clips from films,
television series, and variety shows. Each clip has a joint multimodal
sentiment annotation and separate textual, acoustic, and visual
annotations, with scores in $[-1,1]$. These modality-specific labels
capture agreement and conflict between verbal and nonverbal cues.
The official split contains 1,368 training, 456 validation, and 457
test clips.

\paragraph{CH-SIMS v2.0.}
CH-SIMS v2.0~\citep{liu2022make} expands the collection to 4,402 labeled
clips and 10,161 unlabeled clips. It retains unimodal and multimodal
sentiment annotations while increasing the diversity of acoustic and
visual expressions. The release distinguishes supervised v2(s) and
unsupervised v2(u) subsets, with the supervised subset denoted
CH-SIMS\_v2\_s in our result tables. The evaluation inventory contains
1,034 test clips, matching the reference MER-UniBench subset.

\paragraph{MER2023.}
MER2023~\citep{lian2023mer} is a Chinese audiovisual emotion collection
based on CHEAVD and clips from films and television series. It provides
valence annotations and labels for six emotion categories: neutral,
anger, happiness, sadness, worry, and surprise. The challenge includes 3,373
training/validation examples and separate evaluation subsets for joint
emotion--valence prediction, noise robustness, and semi-supervised
learning. MER-UniBench selects the MER-MULTI subset.

\paragraph{MER2024.}
MER2024~\citep{lian2024mer} extends the collection to 5,030 labeled
training/validation clips and a pool of 115,595 additional clips from
Chinese films and television series. It retains six basic emotion
categories and supports semi-supervised, noise-robust, and open-vocabulary
evaluation. MER-UniBench uses the labeled
MER-SEMI evaluation subset. The broader evaluation inventory contains
2,339 clips, comprising 1,169 MER-SEMI and 1,170 MER-NOISE examples.
MER-NOISE applies additive speech noise to audio and blur to video to
assess robustness to degraded inputs.

\paragraph{EMER.}
EMER (Explainable Multimodal Emotion Reasoning) introduces
natural-language explanations of emotional states using audio, video,
and subtitles~\citep{lian2023emerinitial}. Its initial release contains
100 non-neutral clips selected from MER2023. Human annotators describe
facial expressions, body movements, vocal delivery, speech content,
and contextual cues, and these descriptions are consolidated with ChatGPT
and manually checked to form explanatory references. This resource
supports evaluation of both emotion predictions and their supporting
evidence.

\paragraph{MER-OV.}
MER-OV is the open-vocabulary track of MER2024~\citep{lian2024mer}.
Its annotated subset comprises 332 clips selected from the challenge's
training/validation collection of Chinese film and television material.
The annotation procedure combines acoustic and visual clues with
subtitles to produce emotion-related descriptions, from which an LLM
extracts multiple emotion labels without a predefined category set.
The task assesses nuanced and coexisting emotions.

\paragraph{OV-MERD+.}
OV-MERD+~\citep{lian2025affectgpt} extends the original 332-example
OV-MERD collection~\citep{lian2025ov} to 532 Chinese film and television
clips. Each clip is associated with an emotion description and a
variable-size set of emotion labels, without a fixed category inventory.
The annotation framework combines model-generated acoustic and visual
clues with human verification, supporting evaluation of subtle and
co-occurring emotions.

\paragraph{Controlled ablations.}
\label{app:id-ood-details}
The in-domain experiments jointly train on MELD, MIntRec~2.0, and
IEMOCAP for sentiment analysis, intent recognition, and emotion recognition,
respectively. Out-of-domain evaluation assesses generalization to Mandarin
conversations on MC-EIU through intent and emotion recognition
(\emph{Intent-ZH} and \emph{Emotion-ZH}), and to single-label and multi-label
emotion recognition on MAFW (\emph{Emotion-SL} and \emph{Emotion-ML}).
The latter uses the multi-hot representation described above.
Throughout the controlled-ablation results and associated diagnostics,
MELD, MIntRec~2.0, and IEMOCAP refer to the tasks defined here.
MER-UniBench follows its own benchmark protocol and evaluates emotion
annotations on MELD.
MELD and MIntRec~2.0 use their released training partitions, while
IEMOCAP uses Sessions 1--4 for training and Session 5 for evaluation.
For the prefix-refresh comparison in Table~\ref{tab:app-prefix-refresh}, the
primary evaluation uses the MELD test split (2,610 examples), the
MIntRec~2.0 in-scope test split (2,033 examples), and 2,170 prepared IEMOCAP
Session~5 examples. The prepared transfer inventories contain 2,166 MC-EIU
examples for each of Intent-ZH and Emotion-ZH, 1,839 MAFW Emotion-SL examples,
and 580 MAFW Emotion-ML examples. The additional 1,109
MELD development and 1,106 MIntRec validation examples in the merged prediction
files are used only for split-level diagnostics and are excluded from the
reported test-only scores.
All ablation scores are macro-F1:
\begin{equation}
  \operatorname{MacroF1}
  =\frac{1}{C}\sum_{c=1}^{C}
    \frac{2\,\mathrm{TP}_c}
    {2\,\mathrm{TP}_c+\mathrm{FP}_c+\mathrm{FN}_c},
  \label{eq:app-macro-f1}
\end{equation}
where $C$ is the evaluated label inventory and zero-denominator terms
are assigned zero. For multi-label evaluation, each emotion is treated
as a separate binary decision. This metric differs from MER-UniBench's
HIT, even when both evaluations use the same source dataset.

\subsection{MER-UniBench}
\label{app:mer-unibench-details}

MER-UniBench~\citep{lian2025affectgpt} evaluates free-form affect predictions
across nine datasets spanning sentiment, basic emotion, and fine-grained
emotion recognition, using the reference subsets listed in
Table~\ref{tab:mer-unibench-details}. We report audio--text, video--text, and
audio--video--text results in the modality groups of
Table~\ref{tab:mer-unibench-main}, with external baseline scores taken
from the corresponding publications.

\begin{table}[htbp]
  \centering
  \standardtablesetup
  \caption{\textbf{MER-UniBench evaluation subsets.} Counts are benchmark
  sizes reported by \citet{lian2025affectgpt}, before metric-specific filtering.}
  \label{tab:mer-unibench-details}
  \begin{adjustbox}{max width=\textwidth,center}
  \begin{tabular}{@{}llclc@{}}
    \toprule
    \textbf{Dataset} & \textbf{Task} & \textbf{Metric}
      & \textbf{Subset} & \textbf{Count} \\
    \midrule
    CMU-MOSI & Sentiment & WAF & Test & 686 \\
    CMU-MOSEI & Sentiment & WAF & Test & 4,659 \\
    CH-SIMS & Sentiment & WAF & Test & 457 \\
    CH-SIMS\_v2\_s & Sentiment & WAF & Test & 1,034 \\
    MER2023 & Basic emotion & HIT & MER-MULTI & 411 \\
    MER2024 & Basic emotion & HIT & MER-SEMI & 1,169 \\
    MELD & Basic emotion & HIT & Test & 2,610 \\
    IEMOCAP & Basic emotion & HIT & Session 5 & 1,241 \\
    OV-MERD+ & Fine-grained emotion & EW-F1 & All & 532 \\
    \bottomrule
  \end{tabular}
  \end{adjustbox}
\end{table}

\paragraph{Sentiment evaluation with WAF.}
Reference scores below and above zero are mapped to negative and positive
sentiment, respectively, while neutral ground-truth examples are excluded.
After extracting sentiment from the generated response, weighted-average
F1 is computed as
\begin{equation}
  \operatorname{WAF}
  =\sum_{c\in\{\mathrm{negative},\mathrm{positive}\}}
    \frac{n_c}{N}\operatorname{F1}_c,
  \label{eq:app-waf}
\end{equation}
where $n_c$ is class support among the $N$ retained examples.
The released evaluator maps an extracted neutral prediction to the
non-positive class when binarizing predictions.

\paragraph{Emotion normalization and HIT.}
The evaluator extracts emotion sets from generated responses, normalizes
word forms and synonyms, and maps related labels to core categories
using five emotion wheels. Let $G_k$ denote normalization under wheel $k$, with
$A_{ik}=G_k(Y_i)$ and $B_{ik}=G_k(\widehat Y_i)$ for the reference and
predicted sets. The released evaluator excludes examples with empty
mapped references, leaving $N_k$ examples indexed by $\mathcal I_k$.
For basic emotion recognition, each reference contains one label, and
\begin{equation}
  \operatorname{HIT}
  =\frac{1}{5}\sum_{k=1}^{5}\frac{1}{N_k}
    \sum_{i\in\mathcal I_k}
    \mathbb I\!\left[A_{ik}\cap B_{ik}\ne\varnothing\right].
  \label{eq:app-hit}
\end{equation}
HIT checks whether the prediction includes the reference emotion
without penalizing additional predicted labels.

\paragraph{Fine-grained emotion evaluation with EW-F1.}
For OV-MERD+, precision and recall are first averaged over examples
within each wheel:
\begin{equation}
  P_k=\frac{1}{N_k}\sum_{i\in\mathcal I_k}
    \frac{|A_{ik}\cap B_{ik}|}{\max(1,|B_{ik}|)},
  \qquad
  R_k=\frac{1}{N_k}\sum_{i\in\mathcal I_k}
    \frac{|A_{ik}\cap B_{ik}|}{|A_{ik}|}.
  \label{eq:app-ew-pr}
\end{equation}
An empty prediction receives zero precision and recall. The reported
emotion-wheel F1 averages the resulting wheel-level harmonic means:
\begin{equation}
  \operatorname{EW\text{-}F1}
  =\frac{1}{5}\sum_{k=1}^{5}\frac{2P_kR_k}{P_k+R_k},
  \label{eq:app-ew-f1}
\end{equation}
with zero assigned when $P_k+R_k=0$. All MER-UniBench scores are expressed
as percentages, and the overall score is the unweighted mean of the nine
dataset-level primary metrics.

\subsection{MME-Emotion}
\label{app:mme-emotion-details}

MME-Emotion~\citep{zhang2026mmeemotion} contains 6,500 video--question pairs
covering eight tasks and 27 scenarios, with tasks, source datasets, and
sample counts listed in Table~\ref{tab:mme-emotion-task-details}. It resamples
test examples from public affective datasets and supplies candidate labels
in each question.

\begin{table}[htbp]
  \centering
  \widetablesetup
  \caption{\textbf{Task composition of MME-Emotion.}
  The listed datasets are the source corpora, and counts denote
  video--question pairs in the released benchmark.}
  \label{tab:mme-emotion-task-details}
  \begin{adjustbox}{max width=\textwidth,center}
  \begin{tabular}{@{}llr@{}}
    \toprule
    \textbf{Task} & \textbf{Datasets} & \textbf{Count} \\
    \midrule
    Emotion recognition in the lab (ER-Lab) & IEMOCAP & 500 \\
    Emotion recognition in the wild (ER-Wild) & CAER, MC-EIU, E$^3$, MELD, MAFW, DFEW & 1,800 \\
    Fine-grained emotion recognition (FG-ER) & BoLD & 500 \\
    Fine-grained sentiment analysis (FG-SA) & CMU-MOSI & 500 \\
    Intent recognition (IR) & MC-EIU & 500 \\
    Multi-label emotion recognition (ML-ER) & MAFW, CMU-MOSEI & 600 \\
    Emotion recognition under noise (Noise-ER) & MER2024 & 500 \\
    Sentiment analysis (SA) & CMU-MOSI, CMU-MOSEI, MELD, CH-SIMS v2.0 & 1,600 \\
    \midrule
    \rowcolor{TableGroup}
    \textbf{Total (8 tasks)} & -- & \textbf{6,500} \\
    \bottomrule
  \end{tabular}
  \end{adjustbox}
\end{table}

\paragraph{Recognition and reasoning metrics.}
Rec-S averages the judge's binary rating of the final prediction step over
scored examples. The MME-Emotion paper describes multi-label Rec-S as the
fraction of ground-truth labels recovered, but its released judge prompt rates
the final step as correct only when the prediction matches the ground-truth
label, and \method{}'s scores follow this released prompt and metric script.
Rea-S first averages binary correctness over the reasoning steps of each
response, excluding the final prediction, and then averages over scored
examples. A response without reasoning steps receives zero. The benchmark
combines recognition and reasoning with equal weight:
\begin{equation}
  \operatorname{CoT\text{-}S}
  =\tfrac12\operatorname{Rec\text{-}S}
   +\tfrac12\operatorname{Rea\text{-}S}.
  \label{eq:app-mme-cot}
\end{equation}
The reference overall evaluation aggregates scores over examples.
When every listed example receives a valid score, a metric $S$ is
therefore aggregated as
$S_{\mathrm{overall}}=\sum_{t=1}^{8}n_tS_t/\sum_{t=1}^{8}n_t$,
where $S_t$ and $n_t$ are the score and sample count of task $t$.
This aggregation is inferred from the published task scores: it reproduces
the overall table, unlike an unweighted task mean, although the MME-Emotion
paper does not explicitly give the equation. The mean Rec-S, Rea-S, and CoT-S scores in
Tables~\ref{tab:mme-emotion-main} and~\ref{tab:app-mme-emotion-overall} are
example-weighted over all 6,500 examples. Two response-length statistics accompany the scores:
Avg Step, the mean number of extracted steps per response including the final
prediction step, and Avg Token, the mean number of response tokens.

\section{Experimental Settings and Reproducibility}
\label{app:experimental-settings}
\suppressfloats[t]

This appendix records the configurations needed to interpret and reproduce the
experiments. Section~\ref{app:training-hyperparameters} specifies the controlled
ablations and their privileged-information, thinking-mode, and prefix-source
controls,
Section~\ref{app:backbones-cold-start} describes the two-stage main-benchmark
training protocol, and Section~\ref{app:hardware-software} documents the
implementation, multimodal preprocessing, and evaluation settings.

\subsection{Controlled Ablation Settings}
\label{app:training-hyperparameters}
\label{app:baseline-implementations}

The controlled experiments jointly train on MELD, MIntRec~2.0, and IEMOCAP,
comparing SFT, GRPO, and \method{} with the corresponding Qwen2.5-Omni-3B
and Qwen2.5-Omni-7B base models
(Table~\ref{tab:ablation-domain-ood}).
All post-training methods use a learning rate of $10^{-6}$ and two epochs, with
out-of-domain evaluation assessing generalization to Mandarin conversations
on MC-EIU and to single-label and multi-label emotion recognition on MAFW.
All scores are reported as macro-F1, with task definitions and splits
described in Appendix~\ref{app:id-ood-details}.

SFT optimizes the serialized reference responses, GRPO uses answer-accuracy
and output-format rewards, and \method{} uses the reference rationale to
condition the teacher scoring the student's generated response, with prompts
given in Appendix~\ref{app:rationale-prompts}.
For \method{} ablation runs, the reward-surrogate weight in
Eq.~\eqref{eq:final-loss} is $\alpha=0.2$, set by the implementation parameter
\texttt{reward\_training\_weight}.

\paragraph{Teacher-conditioning ablation.}
Table~\ref{tab:app-rationale-privilege} evaluates two separately optimized
Qwen2.5-Omni-3B models under matched training and evaluation protocols.
Both use \method{} and differ only in teacher-side privileged information:
\emph{Label-only} supplies the ground-truth label, whereas
\emph{Label + Rationale} also supplies the evidence-aware rationale described
in Appendix~\ref{app:rationale-source}.

\paragraph{Thinking-mode comparison.}
Tables~\ref{tab:thinking-mode-ablation} and~\ref{tab:app-thinking-mode-details}
compare GRPO and \method{} with the Qwen2.5-Omni-3B backbone under the
reported \emph{w/ Think} and \emph{w/o Think} inference modes, which
enable or disable an explicit reasoning trace before the final answer.
Within each method, model parameters are held fixed across the two modes.
Thus, this comparison varies inference-time output behavior without retraining,
whereas privileged rationales condition the teacher during training.

\paragraph{Prefix-refresh control.}
The separate comparison summarized in Table~\ref{tab:prefix-refresh-summary}
and detailed in Table~\ref{tab:app-prefix-refresh} uses three
Qwen2.5-Omni-3B runs initialized from the same
cold-start weights and trained with seed 42.
Each run completes two epochs and 752 optimizer steps. The online condition
regenerates prefixes from the current student, the frozen-pool condition is
recorded as using a two-prefix initial-policy pool, and the single-cached
condition reuses one initial-policy prefix.

\subsection{Main Benchmark Training}
\label{app:backbones-cold-start}

MER-UniBench and MME-Emotion use the same Qwen2.5-Omni-7B checkpoint,
obtained through SFT cold-start training at $10^{-5}$ followed by \method{}
at $10^{-6}$ (Tables~\ref{tab:mer-unibench-main} and~\ref{tab:mme-emotion-main}).
The second stage starts from SFT step 1,000, before the first SFT epoch is
complete, and runs for one epoch comprising 2,181 optimizer steps.
Table~\ref{tab:benchmark-training-settings} gives this recipe, which differs
from the two-epoch ablation protocol. MER-UniBench evaluates audio--text,
video--text, and audio--video--text inputs, while MME-Emotion uses
audio--video--text inputs, with tasks, metrics, and aggregation rules detailed
in Appendix~\ref{app:datasets-benchmarks}.

\begin{table}[htbp]
  \centering
  \standardtablesetup
  \caption{\textbf{Training settings for the main benchmark checkpoint.}
  The settings correspond to Tables~\ref{tab:mer-unibench-main}
  and~\ref{tab:mme-emotion-main}.}
  \label{tab:benchmark-training-settings}
  \begin{tabular*}{0.72\textwidth}{@{\extracolsep{\fill}}lcc@{}}
    \toprule
    \textbf{Setting} & \textbf{SFT cold start} & \textbf{OmniOPSD} \\
    \midrule
    Learning rate & $10^{-5}$ & $10^{-6}$ \\
    Selected optimizer step & 1,000 & 2,181 \\
    Tuning & Full & Full \\
    Per-device batch size & 1 & 1 \\
    Gradient accumulation & 2 & 2 \\
    CLI \texttt{max\_length} & 8,192 & 24,000 \\
    \bottomrule
  \end{tabular*}
\end{table}

Both stages use fused AdamW with $(\beta_1,\beta_2)=(0.9,0.95)$,
$\epsilon=10^{-8}$, weight decay 0.1, gradient clipping at 1.0,
and a cosine learning-rate schedule with a 5\% warmup ratio.
Training uses DeepSpeed ZeRO-2 with the visual encoder and
multimodal aligner frozen.  The training and data seeds are both 42.
For the main benchmark run, \method{} rollouts use decoding configuration $G$
with temperature 1.0, top-$p=0.9$, top-$k=50$, and at most 512 new tokens.
For this main-benchmark run, \texttt{reward\_training\_weight} is set to
$\alpha=0.1$. The reward weights in
Eq.~\eqref{eq:reward} are $(\omega_{\mathrm{acc}},\omega_{\mathrm{fmt}})=(1.0,0.5)$.
These coefficients remain fixed during training. The implementation sets
\texttt{reward\_funcs} to \texttt{think\_acc think\_format} and
\texttt{reward\_weights} to \texttt{1 0.5}, in the same order.

\subsection{Implementation and Evaluation}
\label{app:hardware-software}
\label{app:multimodal-preprocessing}
\label{app:evaluation-reproducibility}

The ModelScope SWIFT implementation uses an EMA teacher with decay
$\mu=0.999$, full-vocabulary JSD with $\beta=0.5$, and distillation
temperature $\tau=1.0$. Appendix~\ref{app:algorithm} specifies loss reduction,
the reward surrogate, and EMA updates.

Benchmark inference uses a 512-token response cap and batch sizes of 4 for
MER-UniBench and 2 for MME-Emotion. Classification ablations use task-specific
answer parsing and macro-F1, whereas MER-UniBench and MME-Emotion follow
their benchmark scoring protocols. Any model-assisted answer extraction occurs
during evaluation post-processing, separately from student generation
(prompt formats: Appendix~\ref{app:evaluation-prompts}).

The prefix-refresh evaluation uses \texttt{think} decoding at temperature
zero, at most four video frames, and a usual inference batch size of 8.
Predictions are parsed locally without external-model repair, and unresolved
outputs are counted as errors. On its 13,564-example test-only inventory, the
online, frozen-pool, and single-cached runs contain 262 (1.93\%), 248 (1.83\%),
and 267 (1.97\%) unresolved outputs, respectively.

\section{Additional Experiments}
\label{app:additional-experiments}

This appendix first relates MME-Emotion scores to response length and then
reports task-level evidence for teacher-side privileged rationales,
inference-time thinking, and the single-seed prefix-refresh comparison. The
inference-time and prefix-refresh results are summarized in
Tables~\ref{tab:thinking-mode-ablation} and~\ref{tab:prefix-refresh-summary}.
The classification ablations report macro-F1 on a $0$--$1$ scale, with ID and OOD
averages computed as unweighted means of task-level scores across three and
four tasks, respectively, without pooling predictions or averaging across
training seeds.
Score differences that we report with $\%$ denote absolute differences on the
percentage scale. The tasks and evaluation splits are described in
Appendix~\ref{app:id-ood-details}.

\subsection{MME-Emotion Scores and Response Length}
\label{app:mme-emotion-overall}

Table~\ref{tab:app-mme-emotion-overall} complements the task-level CoT-S in
Table~\ref{tab:mme-emotion-main} with model size, input modalities, and the
response-length statistics defined in Appendix~\ref{app:mme-emotion-details}.
All $6{,}500$ \method{} responses receive valid judge scores. For $58$ of them
($0.9\%$), the multimodal judge request fails, and GPT-4o scores the steps
without video frames, using only the acoustic clues and the ground-truth label, and
excluding these responses changes each mean by at most $0.03\%$. The
unrounded means are $42.4462$ Rec-S, $60.2753$ Rea-S, and $51.3607$ CoT-S, with
$6.6589$ steps and $155.0285$ tokens per response.

\suppressfloats[t]
\begin{table}[t]
  \centering
  \widetablesetup
  \setlength{\tabcolsep}{3.8pt}
  \caption{\textbf{Overall MME-Emotion results with response length.}
  A, V, and T denote audio, video, and text inputs, and scores are percentages.
  Avg.\ steps counts the extracted steps per response, including the final
  prediction step, and Avg.\ tokens is the mean response length. Baseline
  values are taken from \citet{zhang2026mmeemotion}. $^\dagger$OPPO's scores
  are copied from its paper, whose authors re-judge all methods with
  gemini-3.1-flash-lite-preview, but its model size and response lengths are not
  reported. Bold marks the best score among the closed-source models and among
  the open-source audio--video--text models.}
  \label{tab:app-mme-emotion-overall}
  \begin{adjustbox}{max width=\textwidth,center}
  \begin{tabular}{@{}lccccccccc@{}}
    \toprule
    \multirow[c]{2}{*}[-0.6ex]{\textbf{Model}}
    & \multirow{2}{*}{\textbf{Size}}
    & \multicolumn{3}{c}{\textbf{Input}}
    & \multirow{2}{*}{\textbf{Avg. steps}}
    & \multirow{2}{*}{\textbf{Avg. tokens}}
    & \multirow{2}{*}{\textbf{Rec-S}}
    & \multirow{2}{*}{\textbf{Rea-S}}
    & \multirow{2}{*}{\textbf{CoT-S}} \\
    \cmidrule(lr){3-5}
    & & \textbf{A} & \textbf{V} & \textbf{T} & & & & & \\
    \midrule
    \tablegroup{10}{Closed-source MLLMs}
    GPT-4o & -- & & $\checkmark$ & $\checkmark$ & 4.4 & 169.4 & 27.8 & \textbf{79.8} & 53.8 \\
    GPT-4.1 & -- & & $\checkmark$ & $\checkmark$ & 5.2 & 141.2 & 28.8 & 65.2 & 47.0 \\
    Gemini-2.0-Flash & -- & & $\checkmark$ & $\checkmark$ & 4.1 & 64.7 & 36.3 & 60.0 & 48.1 \\
    Gemini-2.5-Flash & -- & & $\checkmark$ & $\checkmark$ & 4.3 & 261.8 & 34.7 & 52.7 & 43.7 \\
    Gemini-2.5-Pro & -- & & $\checkmark$ & $\checkmark$ & 5.1 & 538.6 & \textbf{39.3} & 72.7 & \textbf{56.0} \\
    \midrule
    \tablegroup{10}{Open-source MLLMs}
    Qwen2-Audio & 7B & $\checkmark$ & & $\checkmark$ & 3.0 & 40.3 & 34.1 & 50.4 & 42.3 \\
    Qwen2-VL-7B & 7B & & $\checkmark$ & $\checkmark$ & 2.9 & 68.0 & 29.2 & 38.1 & 33.7 \\
    Qwen2.5-VL-7B & 7B & & $\checkmark$ & $\checkmark$ & 4.8 & 169.7 & 28.4 & 64.8 & 46.6 \\
    Video-LLaVA & 7B & & $\checkmark$ & $\checkmark$ & 2.3 & 19.1 & 25.8 & 32.8 & 29.3 \\
    Video-LLaMA & 7B & $\checkmark$ & $\checkmark$ & $\checkmark$ & 4.5 & 122.5 & 26.1 & 48.5 & 37.3 \\
    Video-LLaMA2 & 7B & $\checkmark$ & $\checkmark$ & $\checkmark$ & 2.6 & 37.7 & 29.2 & 27.7 & 28.4 \\
    Qwen2.5-Omni & 7B & $\checkmark$ & $\checkmark$ & $\checkmark$ & 3.7 & 78.6 & 17.4 & 59.3 & 38.4 \\
    Emotion-LLaMA & 7B & $\checkmark$ & $\checkmark$ & $\checkmark$ & 1.0 & 2.3 & 25.1 & 0.4 & 12.8 \\
    HumanOmni & 7B & $\checkmark$ & $\checkmark$ & $\checkmark$ & 1.0 & 1.3 & 36.0 & 0.3 & 18.1 \\
    R1-Omni & 0.5B & $\checkmark$ & $\checkmark$ & $\checkmark$ & 5.0 & 156.2 & 26.3 & 58.6 & 42.4 \\
    AffectGPT & 7B & $\checkmark$ & $\checkmark$ & $\checkmark$ & 4.9 & 122.8 & 11.9 & 50.6 & 31.2 \\
    OPPO$^\dagger$ & -- & $\checkmark$ & $\checkmark$ & $\checkmark$ & -- & -- & 31.0 & \textbf{68.1} & 49.5 \\
    \textbf{\method{} (Ours)} & 7B & $\checkmark$ & $\checkmark$ & $\checkmark$ & 6.7 & 155.0 & \textbf{42.4} & 60.3 & \textbf{51.4} \\
    \bottomrule
  \end{tabular}
  \end{adjustbox}
\end{table}

\paragraph{Response length across models.}
Across the 17 GPT-4o-judged models with reported length statistics, Avg.\ steps
is strongly rank-correlated with Rea-S (Spearman $\rho=0.79$) and CoT-S
($\rho=0.75$) but not with Rec-S ($\rho=0.17$), and Avg.\ tokens shows the same
pattern ($\rho=0.81$, $0.79$, and $0.17$). This agrees with the positive
association between step count and performance reported by
\citet{zhang2026mmeemotion}. The association is partly structural, because a
response without reasoning steps receives zero Rea-S: Emotion-LLaMA and
HumanOmni answer directly, with $1.0$ step and at most $0.4$ Rea-S. Without
these two models, the correlation weakens but remains positive ($\rho=0.69$ for
Rea-S and $0.64$ for CoT-S). Length does not determine Rea-S, however.
\method{} produces the most steps in the table ($6.7$), yet its Rea-S of $60.3$
is below that of GPT-4o ($79.8$ with $4.4$ steps), Gemini-2.5-Pro, GPT-4.1, and
Qwen2.5-VL-7B. Relative to Qwen2.5-Omni, \method{} raises the step count from
$3.7$ to $6.7$ and nearly doubles the token count, from $78.6$ to $155.0$,
while Rec-S rises by $25.0\%$ and Rea-S by $1.0\%$. Its steps are also
short, averaging about $23$ tokens, compared with about $39$ for GPT-4o and
$106$ for Gemini-2.5-Pro. These cross-model length comparisons are approximate:
\method{} answers in the think-and-answer format of
Appendix~\ref{app:evaluation-prompts}, whereas the published baselines use the
benchmark's step-by-step instruction, and token counts depend on the tokenizer.

\paragraph{Response structure of \method{}.}
\method{} produces a nearly constant response structure. Of its $6{,}500$
responses, $97.4\%$ contain six or seven extracted steps, including the final
prediction step, and all contain between five and nine. Across tasks, the
average ranges only from $6.50$ steps on IR to $6.84$ on FG-ER and from
$149.0$ to $166.8$ tokens (Table~\ref{tab:app-mme-emotion-tasks}), whereas
CoT-S ranges from $28.3$ to $75.3$. Within the model, step count is essentially
uncorrelated with both the final-step rating and the per-response reasoning
score (Pearson $|r|<0.02$), and responses with correct and incorrect final
predictions contain $6.65$ and $6.67$ steps on average.

\begin{table}[htbp]
  \centering
  \standardtablesetup
  \setlength{\tabcolsep}{4pt}
  \caption{\textbf{Per-task response statistics of \method{} on MME-Emotion.}
  Avg.\ steps and Avg.\ tokens are means per response, and scores are
  percentages. The last two columns average Rea-S over responses whose final
  prediction step is judged correct or incorrect, with the correct-response means
  for FG-ER and ML-ER based on 36 and 34 responses, respectively.}
  \label{tab:app-mme-emotion-tasks}
  \begin{tabular*}{\linewidth}{@{\extracolsep{\fill}}lccccccc@{}}
    \toprule
    \multirow{2}{*}{\textbf{Task}}
    & \multirow{2}{*}{\textbf{Avg. steps}}
    & \multirow{2}{*}{\textbf{Avg. tokens}}
    & \multirow{2}{*}{\textbf{Rec-S}}
    & \multirow{2}{*}{\textbf{Rea-S}}
    & \multirow{2}{*}{\textbf{CoT-S}}
    & \multicolumn{2}{c}{\textbf{Rea-S by final step}} \\
    \cmidrule(l){7-8}
    & & & & & & Correct & Incorrect \\
    \midrule
    ER-Lab & 6.65 & 149.0 & 37.2 & 56.9 & 47.1 & 97.5 & 32.9 \\
    ER-Wild & 6.70 & 152.7 & 41.4 & 49.3 & 45.3 & 95.1 & 16.9 \\
    FG-ER & 6.84 & 166.8 & 7.2 & 49.4 & 28.3 & 84.2 & 46.7 \\
    FG-SA & 6.62 & 156.9 & 37.8 & 74.5 & 56.1 & 92.0 & 63.8 \\
    ML-ER & 6.64 & 157.1 & 5.7 & 63.9 & 34.8 & 93.1 & 62.1 \\
    Noise-ER & 6.64 & 151.2 & 74.4 & 76.3 & 75.3 & 96.5 & 17.7 \\
    IR & 6.50 & 162.0 & 35.4 & 61.1 & 48.2 & 84.6 & 48.2 \\
    SA & 6.64 & 153.4 & 63.8 & 66.0 & 64.9 & 94.3 & 16.3 \\
    \midrule
    Overall & 6.66 & 155.0 & 42.4 & 60.3 & 51.4 & 94.1 & 35.3 \\
    \bottomrule
  \end{tabular*}
\end{table}

\paragraph{Step-level judgments and final predictions.}
The judge's step ratings closely follow the final prediction (last two columns
of Table~\ref{tab:app-mme-emotion-tasks}). When the final
prediction step is judged correct, $94.1\%$ of a response's reasoning steps are
judged correct on average, but when it is judged incorrect, this share falls to
$35.3\%$, and every reasoning step is judged incorrect in $1{,}460$ of the
$3{,}741$ incorrect responses ($39.0\%$). In incorrect responses, the last
reasoning step, which names the predicted label in $90.0\%$ of all responses,
is judged correct least often ($22.0\%$, compared with $38.4\%$ for the first
reasoning step). The gap is smaller in tasks where incorrect predictions are
often partially correct: in ML-ER, $331$ of $600$ predicted label sets are
strict subsets of the reference set, and incorrect responses retain $62.1$
Rea-S, compared with $16.3$ in SA. Under this protocol, Rea-S therefore
reflects agreement between the reasoning and the reference label as well as
the consistency of the reasoning with the multimodal cues.

\subsection{Rationale Supervision and Inference-Time Thinking}
\label{app:thinking-mode-details}

We separate two questions: whether evidence-aware rationales improve
teacher-side supervision during training, and whether an explicit reasoning
trace improves predictions at inference. The training ablation optimizes a
separate model for each teacher-conditioning regime, whereas the output-mode
study holds model parameters fixed within each method.

\paragraph{Inference-time output-mode comparison.}
Table~\ref{tab:app-thinking-mode-details} compares GRPO and \method{} with
Qwen2.5-Omni-3B under two inference modes: \emph{w/o Think} omits an explicit
thinking phase, whereas \emph{w/ Think} generates thinking before the answer.
For each method, both modes are decoded from the same trained parameters.
Accordingly, this comparison isolates inference-time generation behavior from
the contribution of privileged rationales during training. The training recipe
is given in Appendix~\ref{app:training-hyperparameters}.

\begin{table}[H]
  \centering
  \standardtablesetup
  \setlength{\tabcolsep}{3.5pt}
  \caption{\textbf{Task-level results for inference-time thinking.}
  Macro-F1 with Qwen2.5-Omni-3B. IEMOCAP uses Session~5.
  Within each method, model parameters are held fixed across
  output modes.
  Bold indicates the best result in each column.}
  \label{tab:app-thinking-mode-details}
  \begin{tabular*}{\linewidth}{@{\extracolsep{\fill}}llcccc@{}}
    \toprule
    \multicolumn{6}{@{}l}{\textit{In-domain (ID)}} \\
    \midrule
    \textbf{Method} & \textbf{Mode}
    & \textbf{MELD}
    & \textbf{MIntRec 2.0}
    & \textbf{IEMOCAP}
    & \textbf{ID Avg.} \\
    \midrule
    GRPO & w/o Think & 0.6854 & 0.3163 & 0.1239 & 0.3752 \\
    GRPO & w/ Think & 0.6093 & 0.2921 & 0.2208 & 0.3741 \\
    \midrule
    \method{} & w/o Think & \textbf{0.6869} & 0.3157 & 0.1285 & 0.3770 \\
    \method{} & w/ Think & 0.6427 & \textbf{0.3248} & \textbf{0.2388} & \textbf{0.4021} \\
    \bottomrule
  \end{tabular*}

  \vspace{6pt}
  \begin{tabular*}{\linewidth}{@{\extracolsep{\fill}}llccccc@{}}
    \toprule
    \multicolumn{7}{@{}l}{\textit{Out-of-domain (OOD)}} \\
    \midrule
    \multirow{2}{*}{\textbf{Method}} & \multirow{2}{*}{\textbf{Mode}}
    & \multicolumn{2}{c}{\textbf{MC-EIU}}
    & \multicolumn{2}{c}{\textbf{MAFW}}
    & \multirow{2}{*}{\textbf{OOD Avg.}} \\
    \cmidrule(lr){3-4}\cmidrule(lr){5-6}
    & & Intent-ZH & Emotion-ZH & Emotion-SL & Emotion-ML & \\
    \midrule
    GRPO & w/o Think & 0.2503 & 0.2823 & 0.1602 & 0.2460 & 0.2347 \\
    GRPO & w/ Think & 0.2352 & 0.2670 & 0.2048 & 0.2298 & 0.2342 \\
    \midrule
    \method{} & w/o Think & 0.2515 & 0.2621 & 0.1597 & \textbf{0.2466} & 0.2300 \\
    \method{} & w/ Think & \textbf{0.2551} & \textbf{0.2834} & \textbf{0.2159} & 0.2343 & \textbf{0.2472} \\
    \bottomrule
  \end{tabular*}
\end{table}

For \method{}, the thinking variant raises ID and OOD mean macro-F1 from
$0.3770$ to $0.4021$ and from $0.2300$ to $0.2472$, respectively, or by
approximately $2.51\%$ and $1.72\%$. The differences are
task-dependent: macro-F1 on MELD decreases from $0.6869$ to $0.6427$, and
MAFW Emotion-ML decreases from $0.2466$ to $0.2343$, while the other five
tasks improve. GRPO's averages change only slightly, from $0.3752$ to
$0.3741$ on ID tasks and from $0.2347$ to $0.2342$ on OOD tasks.

\paragraph{Teacher-conditioning ablation.}
Table~\ref{tab:app-rationale-privilege} provides the exact scores plotted in
Figure~\ref{fig:rationale-privilege}, comparing separately optimized
Qwen2.5-Omni-3B models under matched training and evaluation protocols.
Both models use \method{}. The \emph{Label-only} variant conditions the
self-teacher on the ground-truth label alone, while \emph{Label + Rationale}
additionally supplies the evidence-aware rationale defined in
Appendix~\ref{app:rationale-source}.

\begin{table}[H]
  \centering
  \standardtablesetup
  \setlength{\tabcolsep}{4pt}
  \caption{\textbf{Effect of privileged rationale supervision.}
  Macro-F1 for two Qwen2.5-Omni-3B models separately trained with \method{}
  under matched training and evaluation protocols. The conditions differ only in the
  information available to the self-teacher. IEMOCAP uses Session~5.
  Bold indicates the better result in each column.}
  \label{tab:app-rationale-privilege}
  \begin{tabular*}{\linewidth}{@{\extracolsep{\fill}}lccc@{}}
    \toprule
    \textbf{Teacher privileged information}
    & \textbf{MELD} & \textbf{MIntRec 2.0} & \textbf{IEMOCAP} \\
    \midrule
    Label-only & 0.6289 & 0.3192 & 0.2254 \\
    Label + Rationale
    & \textbf{0.6427} & \textbf{0.3248} & \textbf{0.2388} \\
    \bottomrule
  \end{tabular*}
\end{table}

Using evidence-aware rationales to condition the self-teacher yields consistent
gains over ground-truth labels alone: macro-F1 increases by $1.38\%$,
$0.56\%$, and $1.34\%$ on MELD, MIntRec~2.0, and IEMOCAP,
respectively. Because each condition is optimized separately under matched
protocols, this ablation measures the effect of training-time teacher
conditioning rather than an inference-time prompting effect.

AffectGPT-R1 offers a complementary observation. Its separately optimized GRPO
variants attain $66.49$ on OV-MERD+ when generating a reasoning trace before
the answer and $68.39$ when predicting the answer directly
\citep{lian2025affectgptr1}. In \method{}, the rationale serves a different
role: it is available only to the self-teacher, which supplies dense token-level
targets on student-generated trajectories, and is never an imitation target for
the student. Considered alongside the fixed-parameter inference comparison in
Table~\ref{tab:thinking-mode-ablation}, these results support
rationale-informed process supervision rather than the mere inclusion of a
generated reasoning trace.

\subsection{On-Policy Prefix Refresh}
\label{app:on-policy-ablation}

We compare three sources of completion prefixes used for distillation. The
online variant samples a fresh completion from the current student at each
visit. The recorded frozen-pool configuration uses a two-prefix
initial-student pool intended for fixed rotation, whereas the single-cached
control repeatedly reuses one initial-student completion. The
three Qwen2.5-Omni-3B runs start from the same
cold-start weights, use training seed 42,
and complete two epochs, corresponding to 752 optimizer steps.

Table~\ref{tab:app-prefix-refresh} reports test-only macro-F1. MELD uses the
2,610-example test split, MIntRec~2.0 uses its 2,033-example in-scope
test split, and IEMOCAP uses the 2,170 prepared Session~5 examples. The two
MC-EIU tasks each contain 2,166 prepared examples, while MAFW Emotion-SL and
Emotion-ML contain 1,839 and 580 examples, respectively. Evaluation uses
deterministic local answer extraction, counts unresolved outputs as errors,
and invokes no external model for output repair.

\begin{table}[htbp]
  \centering
  \standardtablesetup
  \setlength{\tabcolsep}{3.5pt}
  \caption{\textbf{Task-level results for on-policy prefix refresh.}
  Test-only macro-F1 with Qwen2.5-Omni-3B under training seed 42.
  We compare reuse of a single cached prefix, a frozen pool of prefixes
  generated by the initial student, and online refresh from the current
  student. IEMOCAP uses Session~5.
  Bold indicates the best result in each column.}
  \label{tab:app-prefix-refresh}
  \begin{tabular*}{\linewidth}{@{\extracolsep{\fill}}lcccc@{}}
    \toprule
    \multicolumn{5}{@{}l}{\textit{In-domain (ID)}} \\
    \midrule
    \textbf{Prefix source}
    & \textbf{MELD}
    & \textbf{MIntRec 2.0}
    & \textbf{IEMOCAP}
    & \textbf{ID Avg.} \\
    \midrule
    Single cached prefix
    & 0.6062 & 0.2978 & 0.2520 & 0.3853 \\
    Frozen prefix pool
    & 0.6021 & 0.3058 & 0.2485 & 0.3855 \\
    Online refresh
    & \textbf{0.6279} & \textbf{0.3383} & \textbf{0.2636}
    & \textbf{0.4099} \\
    \bottomrule
  \end{tabular*}

  \vspace{6pt}
  \begin{tabular*}{\linewidth}{@{\extracolsep{\fill}}lccccc@{}}
    \toprule
    \multicolumn{6}{@{}l}{\textit{Out-of-domain (OOD)}} \\
    \midrule
    \multirow{2}{*}{\textbf{Prefix source}}
    & \multicolumn{2}{c}{\textbf{MC-EIU}}
    & \multicolumn{2}{c}{\textbf{MAFW}}
    & \multirow{2}{*}{\textbf{OOD Avg.}} \\
    \cmidrule(lr){2-3}\cmidrule(lr){4-5}
    & Intent-ZH & Emotion-ZH & Emotion-SL & Emotion-ML & \\
    \midrule
    Single cached prefix
    & \textbf{0.2468} & 0.2768 & 0.2031 & \textbf{0.2308} & 0.2394 \\
    Frozen prefix pool
    & 0.2378 & 0.2704 & 0.2093 & 0.2148 & 0.2331 \\
    Online refresh
    & 0.2375 & \textbf{0.2961} & \textbf{0.2110} & 0.2305 & \textbf{0.2438} \\
    \bottomrule
  \end{tabular*}
\end{table}

Online refresh obtains the highest macro-F1 on all three in-domain tests. Its
ID mean is $0.4099$, compared with $0.3855$ for the frozen pool and $0.3853$
for a single cached prefix. The corresponding OOD means are $0.2438$,
$0.2331$, and $0.2394$. Across all seven tasks, their unweighted overall
means are $0.3150$, $0.2984$, and $0.3019$, respectively. Online refresh
achieves the highest macro-F1 among the three variants on five of the seven
tasks. It also yields higher macro-F1 than the frozen-pool control on six
of the seven tasks, with MC-EIU Intent-ZH as the sole exception. The improvement is not uniform: the single-cached control
is higher on MC-EIU Intent-ZH and slightly higher on MAFW Emotion-ML
($0.2308$ versus $0.2305$). Thus, the single-seed comparison provides
descriptive evidence for an aggregate benefit from current-policy prefixes,
with larger gains in-domain than out-of-domain.
These results should be interpreted within the three prefix-source variants
under their recorded initialization and evaluation protocol.

\begingroup
\hypersetup{pdfborder={0 0 0}}
\section{Qualitative Visualizations}
\label{app:visualizations}

This appendix complements the aggregate results with two views of model
behavior. Section~\ref{app:token-guidance} quantifies and visualizes how
rationale conditioning changes token-level teacher guidance on fixed
student-generated trajectories, while Section~\ref{app:case-studies} examines
representative multimodal reasoning cases across affective tasks.

\subsection{Token-Level Teacher--Student Guidance}
\label{app:token-guidance}

We examine rationale-conditioned supervision using response-level statistics
over 600 held-out examples and six cases: one correct and one incorrect
prediction each for MELD, MIntRec~2.0, and IEMOCAP, following the task
definitions in Appendix~\ref{app:id-ood-details}.

\paragraph{Controlled re-scoring of student trajectories.}
Using the Qwen2.5-Omni-3B cold-start model with the \method{} LoRA adapter at
checkpoint 2000, the student generates one response per video from the original
omni-modal task prompt, without the ground-truth label or reference rationale.
The decoding configuration $G$ uses temperature 1, top-$p=0.9$, top-$k=50$, and at most 512 new tokens.
We re-score these responses with the teacher conditioned on either the answer
label alone or the complete privileged rationale, while holding the frozen
checkpoint $\theta_c$, media inputs, response prefixes, and correct labels fixed. This post-hoc diagnostic isolates
conditioning effects without reconstructing historical EMA targets or
performing optimizer updates.

For a response $y\sim\pi_{\theta_c,G}(\cdot\mid c^S)$, omitting the example
index $i$, define the next-token distributions at its sampled prefixes as
\begin{equation}
\begin{aligned}
p_t &= \operatorname{softmax}\!\left(u_{\theta_c}(c^S,y_{<t})\right),\\
q_t^L &= \operatorname{softmax}\!\left(u_{\theta_c}(c_L^T,y_{<t})\right),\qquad
q_t^R = \operatorname{softmax}\!\left(u_{\theta_c}(c_R^T,y_{<t})\right),
\end{aligned}
\label{eq:app-guidance-distributions}
\end{equation}
where $c^S$ and $c_R^T=c^T$ are the student and teacher contexts in
Eqs.~\eqref{eq:student-context}--\eqref{eq:teacher-context}, and $c_L^T$
replaces the rationale $r$ in $c^T$ with the answer-only reference
\texttt{<answer>}$a$\texttt{</answer>}. Thus $p_t$ and $q_t^R$ instantiate
Eqs.~\eqref{eq:student-distribution}--\eqref{eq:teacher-distribution} with
$\tau=1$ and $\theta=\widetilde\theta_k=\theta_c$.
Scoring evaluates the student prefixes using full-vocabulary logits with FP32
log-softmax at temperature 1, without the top-$p$/top-$k$ truncation used for generation.

\paragraph{Metrics and visual encoding.}
Token backgrounds show the change in each sampled token's log-probability
when the rationale is added to the label-only teacher context,
\begin{equation}
g_t = \log q_t^R(y_t)-\log q_t^L(y_t).
\label{eq:app-rationale-gain}
\end{equation}
Red indicates that adding the rationale increases the probability of the sampled
token, blue indicates a decrease, and near-white indicates little change.
The ratio $\exp(g_t)=q_t^R(y_t)/q_t^L(y_t)$ gives the multiplicative change in
probability. Because teacher weights, labels, and student prefixes are held fixed,
$g_t$ measures the incremental effect of rationale conditioning.
It is distinct from the teacher--student sampled-token log-probability difference,
\begin{equation}
\Delta_t=\log q_t^R(y_t)-\log p_t(y_t).
\label{eq:app-guidance-gap}
\end{equation}
The thin strip aligned below each token shows the symmetric, full-vocabulary JSD
of Eq.~\eqref{eq:generalized-jsd} at $\beta=0.5$,
\begin{equation}
d_t=D_{\mathrm{JSD}}^{0.5}\!\left(p_t,q_t^R\right).
\label{eq:app-guidance-jsd}
\end{equation}
The strip measures full-vocabulary disagreement, whereas the background
measures the displayed token's probability change, and neither encodes token
correctness or an optimizer gradient. All logarithms are natural, so the
metrics are in nats and $0\leq d_t\leq\log 2$ by Eq.~\eqref{eq:app-jsd-bound}. Color limits are fixed across
the six figures from the 600-response cohort: approximately $\pm\,1.421$ for
$g_t$ and $0$--$0.166$ for $d_t$, with saturation applied only to displayed
colors and numerical values left unclipped.

\paragraph{Guidance beyond formatting.}
We distinguish reasoning text, answer text, and structural tags using the original
token spans. For each response $i$ and region $\kappa$ with token index set
$\mathcal T_{i,\kappa}$, we first compute the mean of a token metric
$\phi_{i,t}$ within that region,
\begin{equation}
\overline\phi_{i,\kappa}
=\frac{1}{|\mathcal T_{i,\kappa}|}\sum_{t\in\mathcal T_{i,\kappa}} \phi_{i,t},
\qquad \phi\in\{d,|\Delta|,|g|\}.
\label{eq:app-guidance-region-mean}
\end{equation}
For the 200 held-out videos per dataset,
Table~\ref{tab:app-token-regions} averages the response-level means within
each dataset and weights the three datasets equally, preventing longer
reasoning traces from dominating the comparison.

\begin{table}[htbp]
  \centering
  \standardtablesetup
  \caption{\textbf{Token supervision by response region.} Dataset-balanced means
  are computed over 600 held-out responses, with log-probability measures in nats. Token share
  is the mean within-response fraction.}
  \label{tab:app-token-regions}
  \begin{tabular*}{0.78\textwidth}{@{\extracolsep{\fill}}lcccc@{}}
    \toprule
    \textbf{Region} & \textbf{Mean JSD} & \textbf{Mean $|\Delta_t|$}
      & \textbf{Mean $|g_t|$} & \textbf{Token share} \\
    \midrule
    Reasoning text & 0.01973 & 0.2282 & 0.2021 & 91.17\% \\
    Answer text    & 0.00479 & 0.0357 & 0.0275 & 0.97\% \\
    Format tags    & 0.00071 & 0.0065 & 0.0045 & 5.78\% \\
    \bottomrule
  \end{tabular*}
\end{table}

Uncertainty is estimated by resampling whole source clusters within each dataset, with 2000 bootstrap replicates.
Reasoning text has a higher mean JSD than format tags, with a paired difference
of $0.01901$ and a 95\% percentile interval of $[0.01841,0.01964]$.
The IEMOCAP interval contribution concerns the
sampled Session 5 recordings, not variation across sessions.
Tokens spanning region boundaries and termination tokens are retained in the
share denominator but omitted from the three reported regions, so shares do not
sum to 100\%.
\paragraph{Context-dependent guidance on affective cues.}
Figures~\ref{fig:app-token-meld-correct}--\ref{fig:app-token-mintrec-incorrect}
show complete student responses with chronological video frames, transcript,
teacher-conditioning excerpts, and numbered token annotations.
We select the six illustrative cases after inspecting frozen outputs and
source videos for acoustic, visual, and affective reasoning cues; these cases
are not a random sample, whereas the statistics above use the full
cohort. Teacher references are model-generated conditioning text rather than
independent perceptual annotations, and terminal end-of-message tokens are
omitted from the figures but retained in the underlying records.

In Figure~\ref{fig:app-token-meld-correct}, rationale conditioning favors
\emph{playful} in a positive interpretation ($g_t=0.897$, a probability ratio
of $2.45$ relative to label-only conditioning). In
Figure~\ref{fig:app-token-meld-incorrect}, the same word is suppressed
($g_t=-1.478$, ratio $0.23$), while \emph{subdued} is favored within the same
incorrect response (excerpted in Figure~\ref{fig:dense-affective-guidance}).
In the IEMOCAP pair,
\emph{lively} and \emph{upbeat} receive higher probabilities in the correct
Excited response, whereas the Anger example predicted as Frustration favors
\emph{strained} and suppresses \emph{slow} and \emph{deliberate}.
The MIntRec 2.0 cases connect posture-related wording to a Complain interpretation
and suppress \emph{cheerful} and \emph{smile} in a Greet interpretation of a
Criticize example. Across the 14 numbered anchors, $g_t$ and $\Delta_t$ have
matching signs, although these quantities need not agree in general.

\paragraph{Interpretation.}
Rationale conditioning can increase the probabilities of reference-consistent
cues and decrease those of incompatible wording within an incorrect student
response. These changes help interpret the full-vocabulary teacher targets
used by the token-normalized JSD in Eq.~\eqref{eq:distill-loss}, with a detached
teacher target, while $g_t$ is not an additional reward. The diagnostic does not
measure training-induced correction or the causal effects of on-policy
sampling and EMA, which require the corresponding training comparisons.

\subsection{Reasoning Case Studies}
\label{app:case-studies}

We present five illustrative case studies covering intent recognition,
emotion recognition, multi-label emotion recognition, fine-grained sentiment,
and explainable emotion reasoning.
Tables~\ref{tab:app-case-ir}--\ref{tab:app-case-emer-sadness}
pair frames and speech with model outputs and recorded assessments, with
\method{} recovering the reference affect in all five cases.

\paragraph{Supportive intent beyond literal wording.}
In Table~\ref{tab:app-case-ir}, the speaker explains that the issue concerns his
sister's future rather than money. \method{} connects this statement to the
serious interpersonal context and supportive body language, correctly identifying
\emph{consoling}. Qwen3.8-Flash also recognizes the reassurance, while
Gemini-3.8-Flash treats the utterance as a neutral explanation and GPT-5.6-Luna
predicts \emph{suggesting}. One intermediate step in the \method{} response
overemphasizes financial matters, although its final prediction matches the
reference intent.

\paragraph{Sadness grounded in speech and context.}
Table~\ref{tab:app-case-erlab-sadness} concerns a conversation about someone's
cancer-related illness. \method{} combines subdued speech, pauses, a lowered
gaze, and slumped posture into a consistent interpretation of \emph{sadness}.
Gemini-3.8-Flash also identifies sadness from the quiet delivery and illness
context, while MiMo-v2.5 and GPT-5.6-Luna predict \emph{happiness}. AffectGPT
returns the correct emotion but introduces unsupported visual details in a
rejected reasoning step.

\paragraph{Recovering co-occurring emotions.}
The question \lq\lq Supposing I don't want flowers?'' in
Table~\ref{tab:app-case-mler-question} expresses more than a single negative
emotion. \method{} associates forceful delivery and rejection with
\emph{anger}, while its facial interpretation supports the accompanying
\emph{disgust}. Emotion-LLaMA returns anger alone, while GPT-5.6-Luna and
Gemini-3.8-Flash include anger but substitute other categories for disgust,
and MiMo-v2.5 shifts toward sadness and anxiety.

\paragraph{Negative sentiment behind a conversational expression.}
In Table~\ref{tab:app-case-fgsa-criticism}, the remark \lq\lq It was really kind of
dumbed down for us'' carries criticism despite a relatively conversational
facial expression. \method{} links the evaluative wording and subdued delivery
to the reference \emph{negative} sentiment. GPT-5.6-Luna instead emphasizes
the smile and predicts positive sentiment, while Qwen3.8-Flash predicts
weak positive. VideoLLaMA and Gemini-3.8-Flash recover negative polarity but
assign strong negative and weak negative, respectively.

\paragraph{Explaining sadness despite a smile.}
The final case, from the EMER Dataset, evaluates an explanation against
annotated emotional clues, following the clue-based perspective described
by \citet{cheng2024emotion}. In Table~\ref{tab:app-case-emer-sadness}, smiling
coexists with longing for an absent person and subdued speech.
PandaGPT and Video-ChatGPT misread the smile as positive affect.
Both Emotion-LLaMA and \method{} identify sadness, with \method{} linking
more vocal and facial evidence and receiving a clue score of 9/10, compared
with 8/10 for Emotion-LLaMA. The assessment also notes that \method{} omits
tearful eyes and an explicit reference to the absent person.

\clearpage

\begingroup
\makeatletter
\setlength{\@fptop}{0pt}
\setlength{\@fpbot}{0pt plus 1fil}
\makeatother

\begin{figure}[p]
  \centering
  \includegraphics[width=\textwidth]{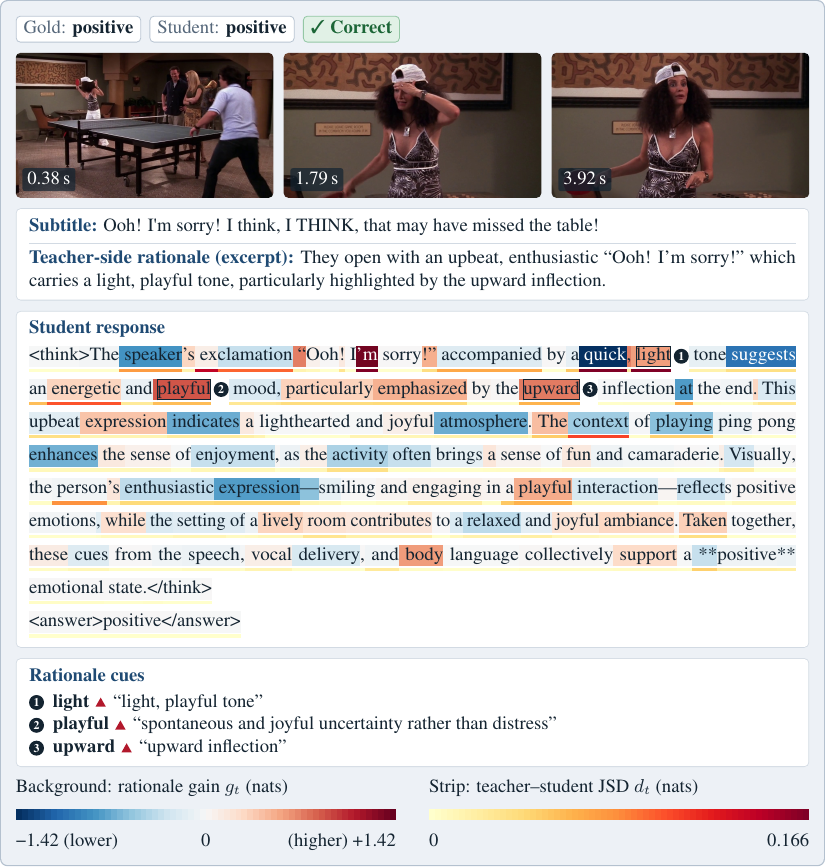}
  \caption{\textbf{MELD: correct prediction} (\texttt{test\_dia162\_utt0}).
  During a playful table-tennis apology, rationale conditioning favors
  \emph{light}, \emph{playful}, and \emph{upward} relative to label-only conditioning.}
  \label{fig:app-token-meld-correct}
\end{figure}

\begin{figure}[p]
  \centering
  \includegraphics[width=\textwidth]{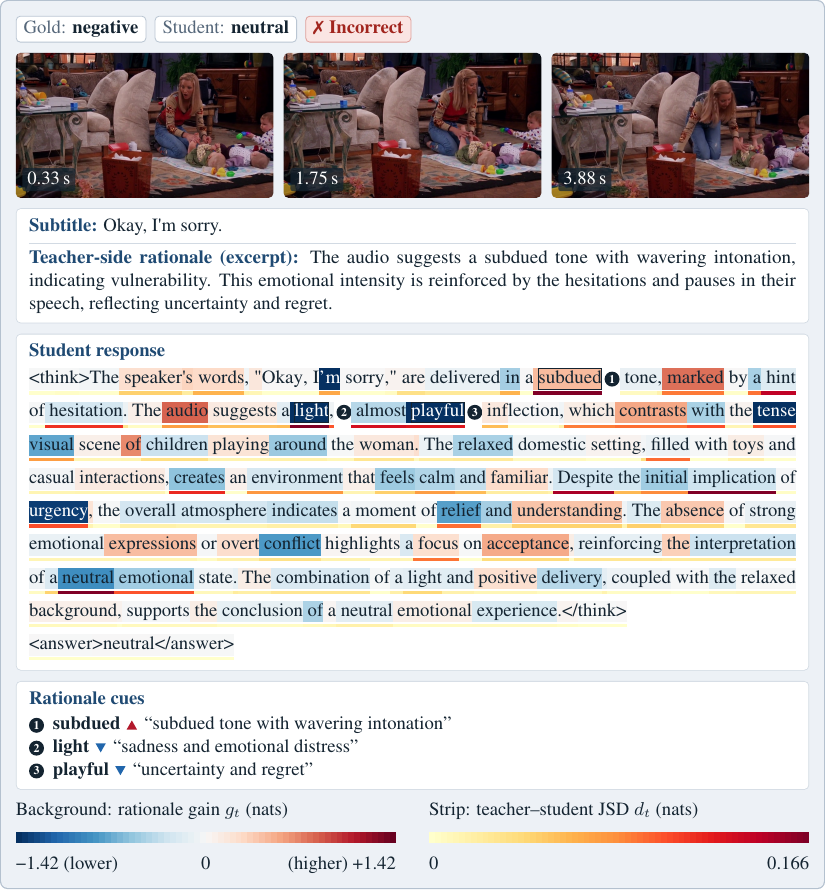}
  \caption{\textbf{MELD: incorrect prediction} (\texttt{test\_dia100\_utt5}).
  For a negative example predicted as neutral, rationale conditioning favors
  \emph{subdued} while suppressing the reference-inconsistent
  \emph{light} and \emph{playful} interpretation.}
  \label{fig:app-token-meld-incorrect}
\end{figure}

\begin{figure}[p]
  \centering
  \includegraphics[width=\textwidth]{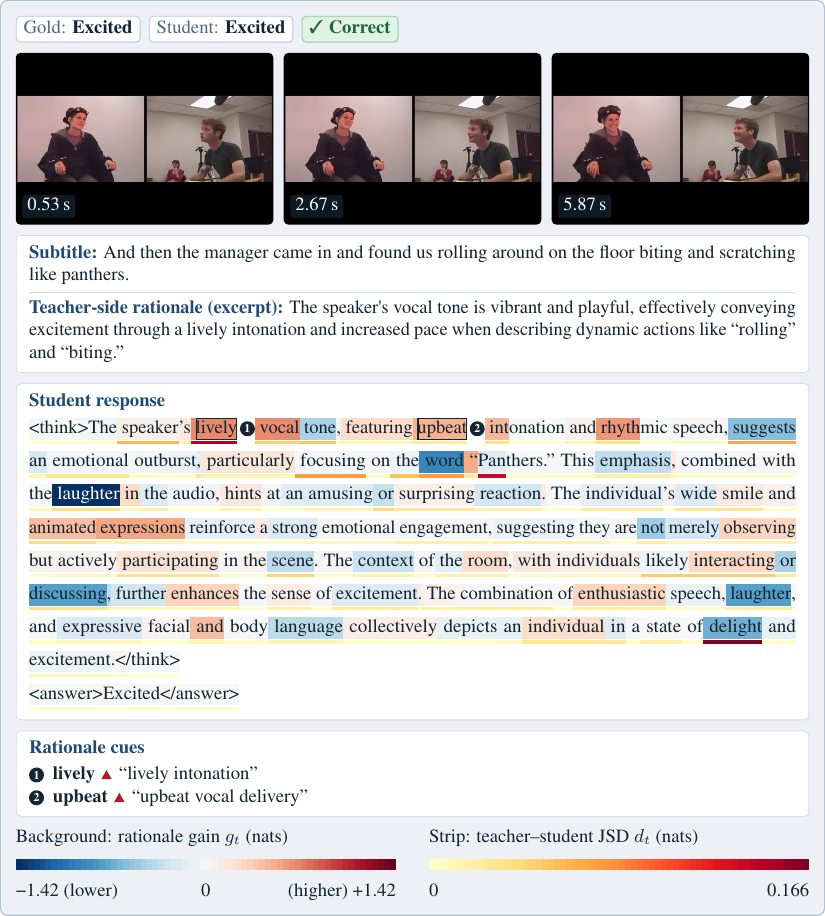}
  \caption{\textbf{IEMOCAP: correct prediction} (\texttt{Ses05F\_script03\_1\_F030}).
  Rationale conditioning increases the probabilities of \emph{lively} and
  \emph{upbeat} in the student's explanation of an Excited prediction.}
  \label{fig:app-token-iemocap-correct}
\end{figure}

\begin{figure}[p]
  \centering
  \includegraphics[width=\textwidth]{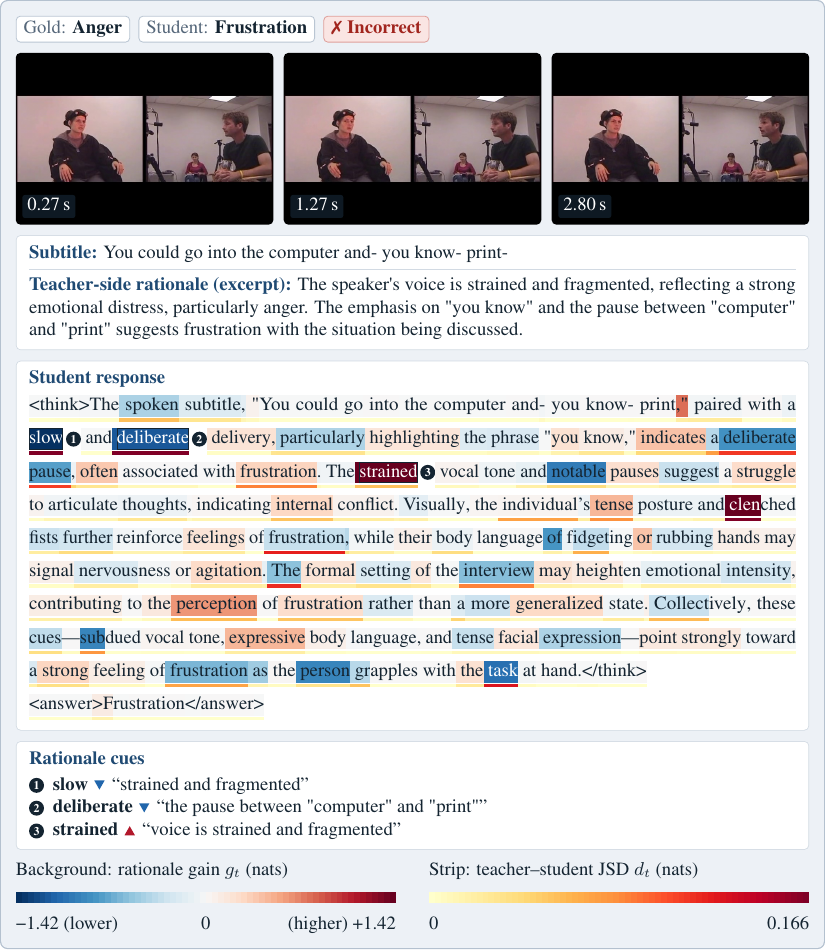}
  \caption{\textbf{IEMOCAP: incorrect prediction} (\texttt{Ses05F\_impro05\_F035}).
  In a response predicting Frustration instead of Anger, rationale conditioning
  suppresses \emph{slow} and \emph{deliberate} while favoring \emph{strained}.}
  \label{fig:app-token-iemocap-incorrect}
\end{figure}

\begin{figure}[p]
  \centering
  \includegraphics[width=\textwidth]{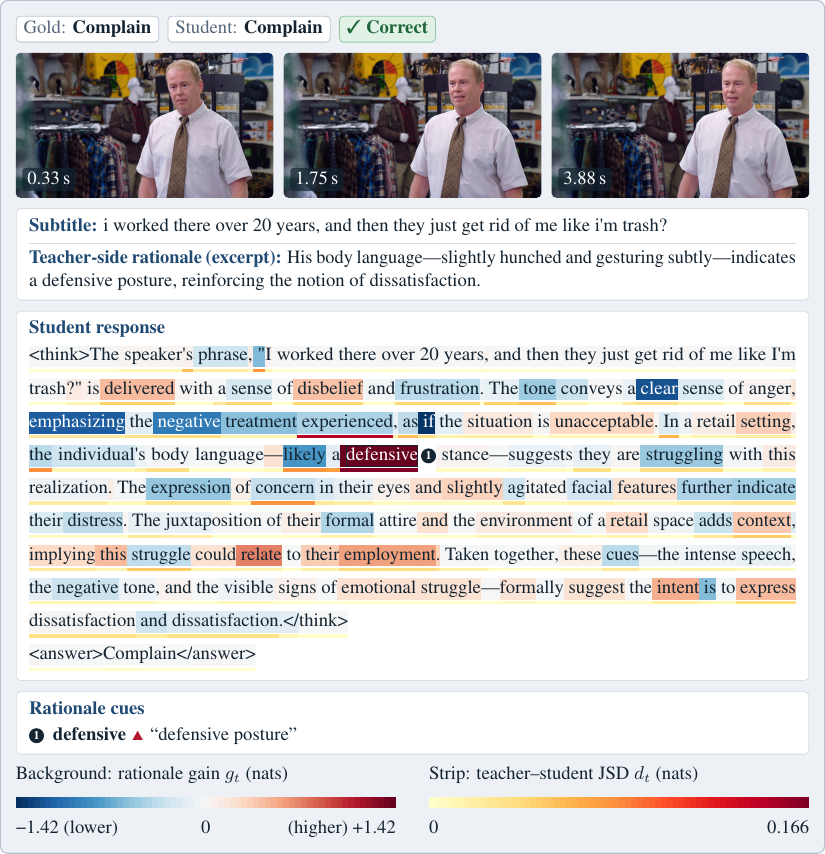}
  \caption{\textbf{MIntRec 2.0: correct prediction} (\texttt{dia505\_utt7}).
  In a complaint about dismissal, rationale conditioning favors \emph{defensive},
  linking posture-related wording to the student's Complain interpretation.}
  \label{fig:app-token-mintrec-correct}
\end{figure}

\begin{figure}[p]
  \centering
  \includegraphics[width=\textwidth]{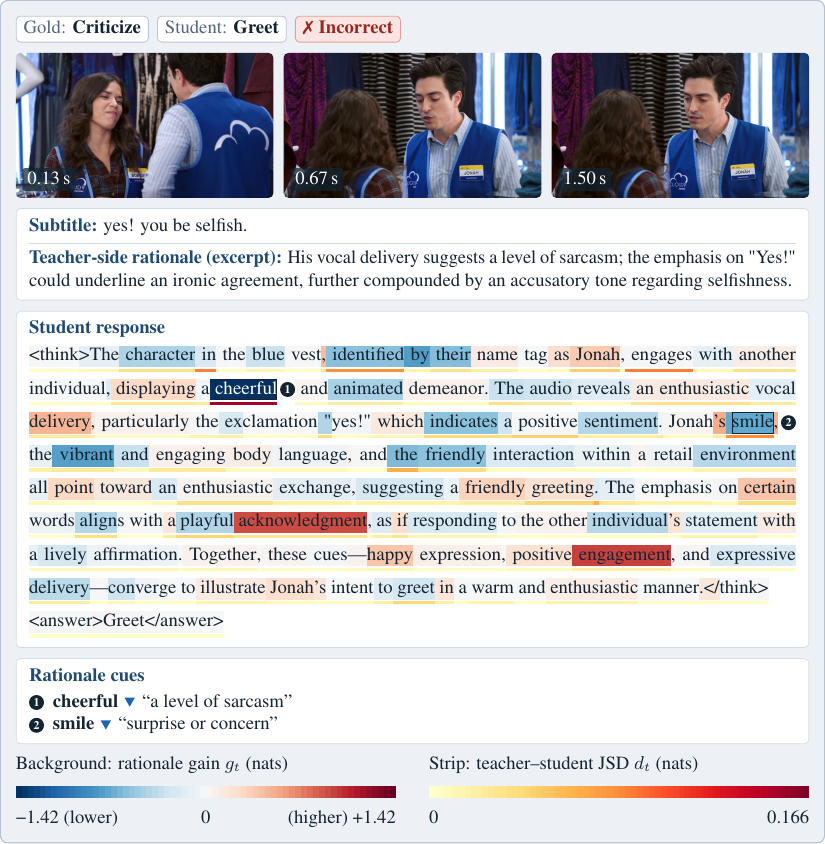}
  \caption{\textbf{MIntRec 2.0: incorrect prediction} (\texttt{dia673\_utt3}).
  For a selfishness remark labeled Criticize but predicted as Greet,
  rationale conditioning suppresses \emph{cheerful} and \emph{smile}
  relative to label-only conditioning.}
  \label{fig:app-token-mintrec-incorrect}
\end{figure}

\clearpage
\endgroup

\definecolor{CaseInk}{HTML}{142431}
\definecolor{CaseMuted}{HTML}{55677A}
\definecolor{CaseAccent}{HTML}{1F4A73}
\definecolor{CasePanelBG}{HTML}{EDF1F6}
\definecolor{CasePanelFrame}{HTML}{B4C1CF}
\definecolor{CaseCardFrame}{HTML}{D2DAE3}
\definecolor{CaseOkFill}{HTML}{E1F1E6}
\definecolor{CaseOkFrame}{HTML}{94C7A5}
\definecolor{CaseOkInk}{HTML}{1C6637}
\definecolor{CaseMidFill}{HTML}{FCEFD4}
\definecolor{CaseMidFrame}{HTML}{E2BC73}
\definecolor{CaseMidInk}{HTML}{85560A}
\definecolor{CaseBadFill}{HTML}{FAE5E3}
\definecolor{CaseBadFrame}{HTML}{DFA09B}
\definecolor{CaseBadInk}{HTML}{A1251E}
\definecolor{CaseOursFill}{HTML}{EAF1FA}
\definecolor{CaseOursFrame}{HTML}{2E6AA3}

\newcommand\CaseFont{\fontsize{9}{10}\selectfont\color{CaseInk}}
\newcommand\CaseLabel[1]{{\bfseries\color{CaseAccent}#1}}
\newcommand\CaseHead[1]{{\bfseries\color{CaseMuted}#1}}
\newcommand\CaseJustify{\leftskip=0pt\rightskip=0pt\parfillskip=0pt plus 1fil\relax
  \tolerance=1000\emergencystretch=8pt\finalhyphendemerits=0\relax}
\newcommand\CaseDing[1]{{\usefont{U}{pzd}{m}{n}\char#1}}

\newlength\CaseGap      \setlength\CaseGap{5pt}      %
\newlength\CaseFrameGap \setlength\CaseFrameGap{5pt} %
\newlength\CaseNameW    \setlength\CaseNameW{62pt}   %
\newlength\CaseColGap   \setlength\CaseColGap{6pt}
\newlength\CaseRateW    \setlength\CaseRateW{20pt}   %
\newlength\CaseStepGap  \setlength\CaseStepGap{0pt}
\newsavebox\CaseBox

\newtcolorbox{CasePanel}{enhanced, width=\textwidth, colback=CasePanelBG,
  colframe=CasePanelFrame, boxrule=0.6pt, arc=3pt, boxsep=0pt,
  left=6pt, right=6pt, top=6pt, bottom=6pt, before skip=0pt, after skip=0pt,
  fontupper=\CaseFont}
\newtcolorbox{CaseCard}[1][]{enhanced, colback=white, colframe=CaseCardFrame,
  boxrule=0.4pt, arc=2pt, boxsep=0pt, left=5pt, right=5pt, top=4pt, bottom=4pt,
  before skip=0pt, after skip=0pt, fontupper=\CaseFont, #1}
\newcommand\CaseNext{\par\vspace{\CaseGap}\nointerlineskip}
\newcommand\CaseRule{\par\vspace{3pt}{\color{CaseCardFrame}\hrule height .4pt}\vspace{3pt}}

\newcommand\CaseChipBox[4]{%
  \tikz[baseline=(c.base)]\node[fill=#1, draw=#2, line width=.4pt,
    rounded corners=2pt, inner xsep=2.8pt, inner ysep=.7pt,
    text height=6.2pt, text depth=2.1pt, font=\CaseFont\color{#3}] (c) {#4};}
\newcommand\CaseHalf{\tikz[baseline=-.2pt]{\draw[CaseMidInk, line width=.55pt]
  (2.6pt,2.6pt) circle (2.3pt); \fill[CaseMidInk] (2.6pt,4.9pt) arc (90:270:2.3pt) -- cycle;}}
\newcommand\CaseCorrect{\CaseChipBox{CaseOkFill}{CaseOkFrame}{CaseOkInk}%
  {\bfseries\CaseDing{51}\hspace{2pt}Correct}}
\newcommand\CasePartial{\CaseChipBox{CaseMidFill}{CaseMidFrame}{CaseMidInk}%
  {\bfseries\CaseHalf\hspace{2pt}Partial}}
\newcommand\CaseIncorrect{\CaseChipBox{CaseBadFill}{CaseBadFrame}{CaseBadInk}%
  {\bfseries\CaseDing{55}\hspace{2pt}Incorrect}}
\newcommand\CaseRefChip[1]{\CaseChipBox{CaseAccent}{CaseAccent}{white}%
  {Reference:\hspace{3pt}\bfseries #1}}
\newcommand\CaseRate[1]{\if\relax\detokenize{#1}\relax\else\CaseRateParse#1\relax\fi}
\def\CaseRateParse#1/#2\relax{%
  \ifnum#1=#2\relax\def\CaseRateStyle{CaseOk}%
  \else\ifnum#1=0\relax\def\CaseRateStyle{CaseBad}%
  \else\def\CaseRateStyle{CaseMid}\fi\fi
  \tikz[baseline=(r.base)]\node[fill=\CaseRateStyle Fill, draw=\CaseRateStyle Frame,
    line width=.4pt, rounded corners=1.8pt, inner xsep=0pt, inner ysep=.8pt,
    minimum width=\CaseRateW, text height=6.1pt, text depth=1.3pt,
    font=\CaseFont\color{\CaseRateStyle Ink}] (r) {#1/#2};}

\newlength\CaseFrameW
\newlength\CaseFrameH
\newlength\CaseFrameFullH
\newlength\CaseFrameShift
\newcommand\CaseFrameGeometry[4]{%
  \setlength\CaseFrameW{\dimexpr(\linewidth-2\CaseFrameGap)/3\relax}%
  \setlength\CaseFrameFullH{\dimexpr\CaseFrameW*#2/#1\relax}%
  \setlength\CaseFrameH{\dimexpr\CaseFrameW*\numexpr#2-#3-#4\relax/#1\relax}%
  \setlength\CaseFrameShift{\dimexpr\CaseFrameW*#4/#1\relax}}
\newcommand\CaseFrame[2]{%
  \begin{tikzpicture}[baseline=(current bounding box.north)]
    \begin{scope}
      \clip[rounded corners=2.2pt] (0,0) rectangle (\CaseFrameW,\CaseFrameH);
      \node[anchor=south west, inner sep=0pt, outer sep=0pt] at (0,-\CaseFrameShift)
        {\includegraphics[width=\CaseFrameW,height=\CaseFrameFullH]{#1}};
    \end{scope}
    \node[anchor=south west, fill=CaseInk, fill opacity=.72, text opacity=1,
      rounded corners=1.2pt, inner xsep=2.4pt, inner ysep=1.4pt,
      text height=6.3pt, text depth=2pt, font=\CaseFont\color{white}]
      at (3pt,3pt) {#2\,s};
  \end{tikzpicture}}
\newcommand\CaseFrames[6]{\noindent\CaseFrame{#1}{#2}\hfill\CaseFrame{#3}{#4}\hfill
  \CaseFrame{#5}{#6}\par}

\newcommand\CaseContextLine[2]{%
  \sbox\CaseBox{\CaseRefChip{#2}}%
  \noindent\begin{minipage}[t]{\dimexpr\linewidth-\wd\CaseBox-8pt\relax}\CaseJustify
    #1\end{minipage}\hfill\usebox\CaseBox\par}

\newcommand\CaseModelHeader[1]{%
  \noindent\hbox to \linewidth{\hbox to \CaseNameW{\CaseHead{Model}\hss}%
    \hskip\CaseColGap\CaseHead{#1}\hfil\CaseHead{Rating}}\par
  \vspace{2.5pt}{\color{CaseCardFrame}\hrule height .4pt}\vspace{3.5pt}}
\newcommand\CaseModel[3]{%
  \par\noindent
  \begin{minipage}[t]{\CaseNameW}\raggedright\leavevmode\bfseries #1\par\vspace{3pt}\mdseries #2\end{minipage}%
  \hspace{\CaseColGap}%
  \begin{minipage}[t]{\dimexpr\linewidth-\CaseNameW-\CaseColGap\relax}%
    \lineskip=0pt\lineskiplimit=0pt\relax #3\end{minipage}\par}
\newcommand\CaseOursName[1]{{\color{CaseAccent}#1}\par\mdseries\color{CaseMuted}(ours)}
\newcommand\CaseStep[3]{%
  \par\noindent
  \begin{minipage}[t]{\dimexpr\linewidth-\CaseRateW-5pt\relax}\CaseJustify
    \strut\textbf{#1}\ #2\nobreak\hskip0pt\strut\end{minipage}\hfill\CaseRate{#3}\par
  \vspace{\CaseStepGap}}%
\newcommand\CaseNote[1]{\par\noindent\strut{\itshape\color{CaseMuted}#1}\strut\par\vspace{\CaseStepGap}}
\newenvironment{CaseOurs}{\begin{CaseCard}[colback=CaseOursFill, colframe=CaseOursFrame,
  boxrule=.5pt, leftrule=2.4pt, left=3pt, right=4.9pt, top=3.9pt, bottom=3.9pt]}{\end{CaseCard}}

\newcommand\CaseScore[1]{%
  {\bfseries\color{CaseAccent}#1/10}\par\vspace{2pt}%
  \tikz{\fill[CaseCardFrame, rounded corners=1pt] (0,0) rectangle (\CaseNameW-4pt,3pt);
    \fill[CaseAccent, rounded corners=1pt] (0,0) rectangle ({(\CaseNameW-4pt)*#1/10},3pt);}}
\newcommand\CaseEmerModel[4]{%
  \par\noindent
  \begin{minipage}[t]{\CaseNameW}\raggedright\leavevmode\bfseries #1\par\vspace{3pt}\mdseries\CaseScore{#2}\end{minipage}%
  \hspace{\CaseColGap}%
  \begin{minipage}[t]{\dimexpr\linewidth-\CaseNameW-\CaseColGap\relax}\CaseJustify
    #3\par\vspace{2pt}{\color{CaseMuted}\textit{Criteria:} #4\par}\end{minipage}\par}
\newcommand\CaseEmerHeader{%
  \noindent\hbox to \linewidth{\hbox to \CaseNameW{\CaseHead{Model / score}\hss}%
    \hskip\CaseColGap\CaseHead{Response excerpt and clue-level assessment}\hfil}\par
  \vspace{2.5pt}{\color{CaseCardFrame}\hrule height .4pt}\vspace{3.5pt}}

\begingroup
\makeatletter
\long\def\@makecaption#1#2{%
  \vskip\abovecaptionskip
  {\normalfont\fontsize{10}{11.5}\selectfont\raggedright #1: #2\par}%
  \vskip6pt}
\setlength{\@fptop}{0pt}
\setlength{\@fpbot}{0pt plus 1fil}
\makeatother

\begin{table}[p]
\centering
\caption{\textbf{IR: consolation in a serious conversation.}\newline MME-Emotion example \texttt{IR\_\_dia\_560\_utt\_11\_ch} has the reference label \emph{consoling}.}
\label{tab:app-case-ir}
\begin{CasePanel}
\CaseFrameGeometry{1024}{576}{0}{0}
\CaseFrames{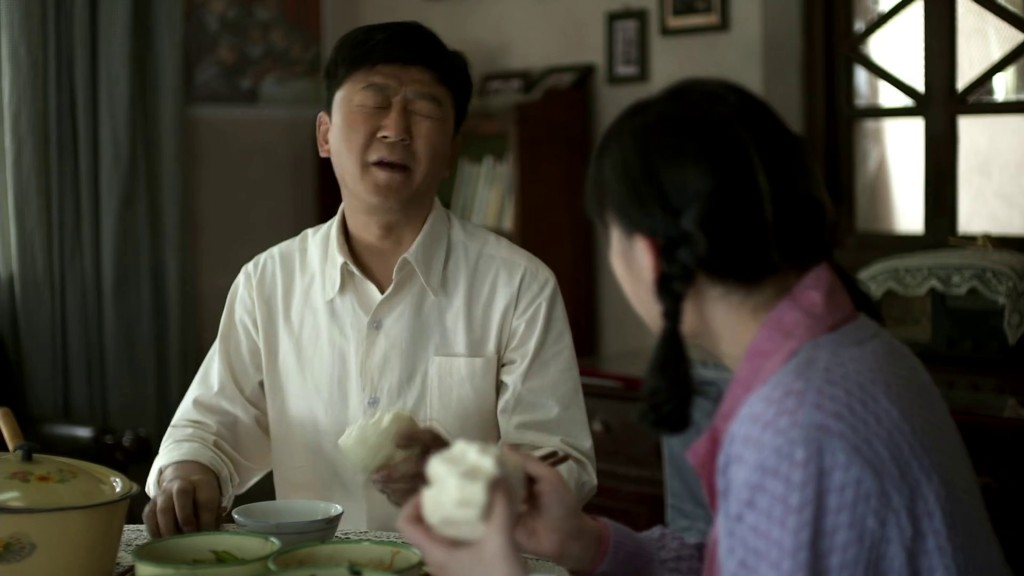}{0.32}{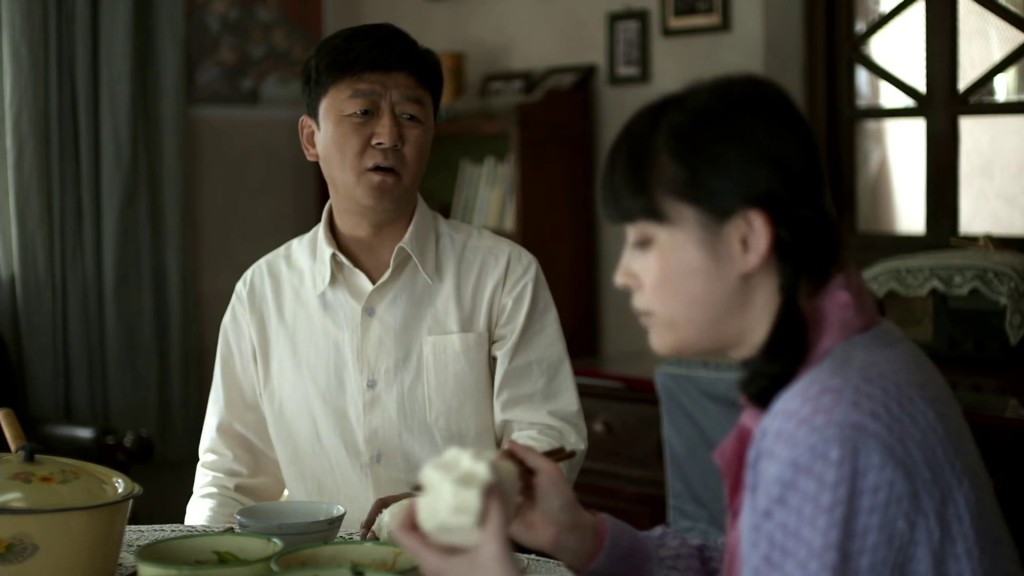}{2.40}{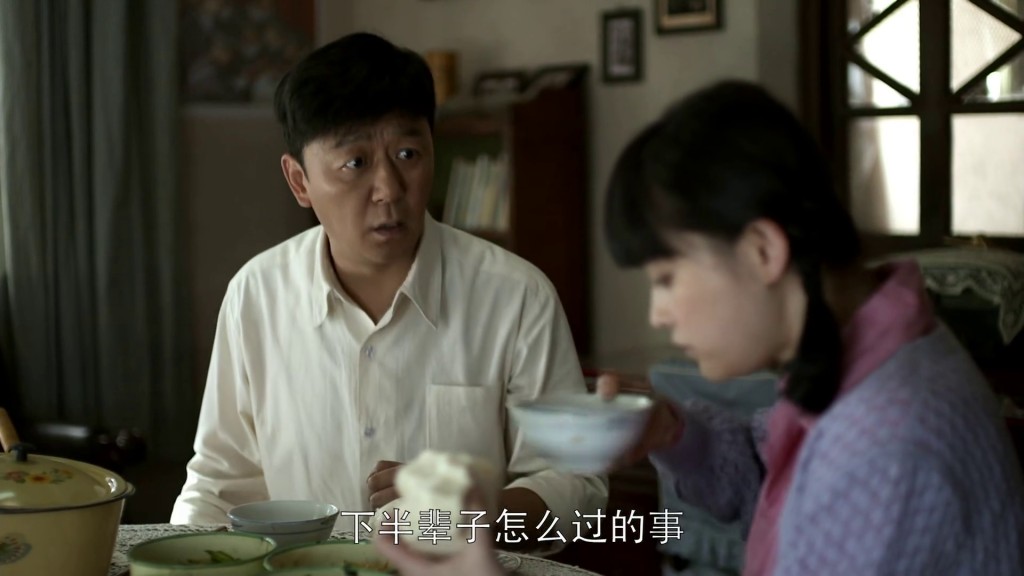}{5.20}
\CaseNext
\begin{CaseCard}
\CaseContextLine{\CaseLabel{Speech (English translation):} ``Oh, this isn't about money. It's about how my sister is going to live the rest of her life.''}{consoling}
\end{CaseCard}
\CaseNext
\begin{CaseCard}
\CaseModelHeader{Reasoning and final prediction}
\CaseModel{VideoLLaVA}{\CaseCorrect}{%
\CaseStep{Step 1:}{Observe the man eating cake and the woman watching him.}{1/1}
\CaseStep{Step 2:}{Interpret her attention as an attempt to provide comfort or support.}{1/1}
\CaseStep{Step 3 (prediction):}{The predicted intent is consoling.}{1/1}}
\CaseRule
\CaseModel{Qwen3.8-\newline Flash}{\CaseCorrect}{%
\CaseStep{Step 1:}{Observe the man and woman dining together, with the woman appearing downcast and distressed.}{1/1}
\CaseStep{Step 2:}{Note the man's sigh, serious expression, concerned gaze, and earnest tone.}{1/1}
\CaseStep{Step 3:}{Interpret his statement about prioritizing his sister's lifelong well-being over money as reassurance and emotional support.}{1/1}
\CaseStep{Step 4 (prediction):}{The predicted intent is consoling.}{1/1}}
\CaseRule
\CaseModel{Gemini-3.8-\newline Flash}{\CaseIncorrect}{%
\CaseStep{Step 1:}{Interpret the speaker's statement about the issue concerning his sister's future.}{1/1}
\CaseStep{Step 2:}{Identify the utterance as a plain explanation of his perspective.}{0/1}
\CaseStep{Step 3:}{Exclude questioning, consoling, suggesting, encouraging, agreeing, wishing, and acknowledging intents.}{0/1}
\CaseStep{Step 4 (prediction):}{The predicted intent is neutral.}{0/1}}
\CaseRule
\CaseModel{GPT-5.6-Luna}{\CaseIncorrect}{%
\CaseStep{Step 1:}{Interpret the dialogue as concern for the younger sister's future.}{1/1}
\CaseStep{Step 2:}{Note the man's earnest delivery and emphatic gestures.}{1/1}
\CaseStep{Step 3:}{Identify his attempt to persuade or advise the woman.}{1/1}
\CaseStep{Step 4 (prediction):}{The predicted intent is suggesting.}{0/1}}
\end{CaseCard}
\CaseNext
\begin{CaseOurs}
\CaseModel{\CaseOursName{\method{}}}{\CaseCorrect}{%
\CaseStep{Step 1:}{The speaker's words, "This is not about money," convey a sense of urgency and concern, particularly regarding the future of their sister.}{1/1}
\CaseStep{Step 2:}{Their vocal tone is described as neutral, with a slight emphasis on "money", indicating a serious discussion about financial matters.}{0/1}
\CaseStep{Step 3:}{This aligns with the visual context of a domestic setting, where the speaker's body language - such as leaning forward and maintaining eye contact - suggests a desire to console and support the other person.}{1/1}
\CaseStep{Step 4:}{The overall atmosphere, characterized by a serious conversation, further reinforces the intent to console, as the speaker appears to be addressing a significant emotional issue.}{1/1}
\CaseStep{Step 5:}{The combination of verbal, vocal, and visual cues leads to a clear interpretation of the speaker's intent as one of consolation.}{1/1}
\CaseStep{Step 6 (prediction):}{The predicted intent is consoling.}{1/1}}
\end{CaseOurs}
\end{CasePanel}
\end{table}
\clearpage

\begin{table}[p]
\centering
\caption{\textbf{ER-Lab: sadness when discussing an illness.}\newline MME-Emotion example \texttt{ER-Lab\_\_Ses05M\_impro06\_M002} has the reference label \emph{sadness}.}
\label{tab:app-case-erlab-sadness}
\begin{CasePanel}
\CaseFrameGeometry{720}{480}{123}{122}
\CaseFrames{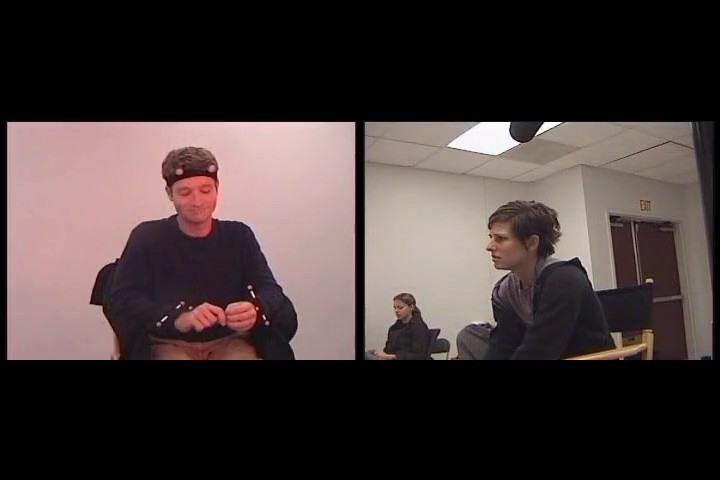}{0.37}{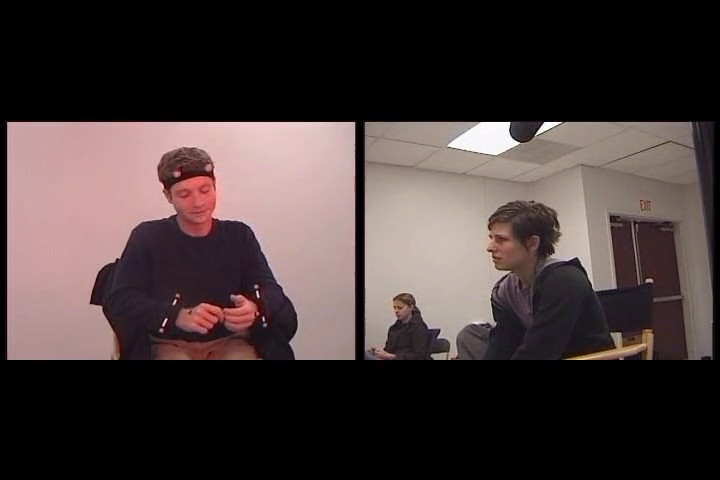}{2.70}{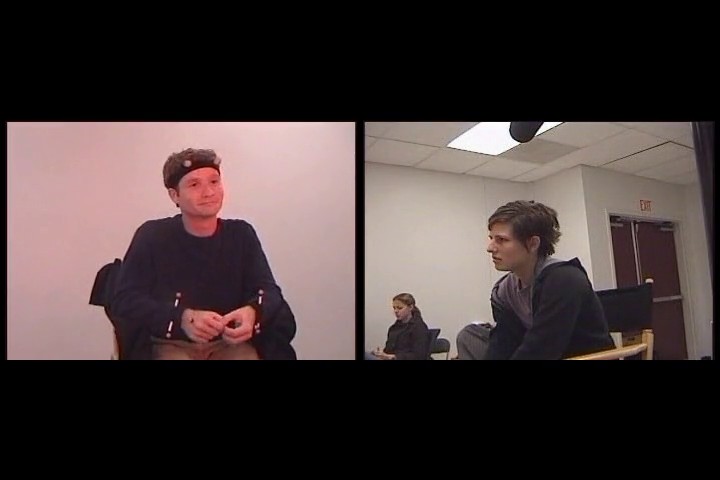}{5.81}
\CaseNext
\begin{CaseCard}
\CaseContextLine{\CaseLabel{Speech:} ``Yeah, he's been sick for a while. He had cancer.''}{sadness}
\end{CaseCard}
\CaseNext
\begin{CaseCard}
\CaseModelHeader{Reasoning and final prediction}
\CaseModel{AffectGPT}{\CaseCorrect}{%
\CaseStep{Step 1:}{Observe crying, wet makeup, and face-clasping, indicating strong distress.}{0/1}
\CaseStep{Step 2:}{Note the shaking voice and monotonous delivery, suggesting fear, anxiety, or emotional detachment.}{1/1}
\CaseStep{Step 3:}{Combine the visual and auditory cues to identify the predominant negative emotion.}{1/1}
\CaseStep{Step 4 (prediction):}{The predicted emotion is Sadness.}{1/1}}
\CaseRule
\CaseModel{MiMo-v2.5}{\CaseIncorrect}{%
\CaseStep{Step 1:}{Analyze the visual cues in the video, focusing on facial expressions and body language of the participants.}{1/1}
\CaseStep{Step 2:}{Examine the audio transcript and tone for emotional content.}{1/1}
\CaseStep{Step 3:}{Identify any conflict between the semantic content of the speech and the visual expressions.}{1/1}
\CaseStep{Step 4:}{Prioritize the displayed facial expression as the primary indicator of emotion.}{0/1}
\CaseStep{Step 5 (prediction):}{Based on the sustained smile and relaxed demeanor, the predicted emotion is Happiness.}{0/1}}
\CaseRule
\CaseModel{Gemini-3.8-\newline Flash}{\CaseCorrect}{%
\CaseStep{Step 1:}{Observe visual cues such as downward gaze, nervous fidgeting, and subdued facial expressions indicating grief.}{1/1}
\CaseStep{Step 2:}{Analyze auditory cues, noting the quiet, low, and resigned tone of voice when discussing illness.}{1/1}
\CaseStep{Step 3:}{Synthesize visual and auditory information with the context of discussing severe illness and death.}{1/1}
\CaseStep{Step 4 (prediction):}{The predicted emotion is Sadness.}{1/1}}
\CaseRule
\CaseModel{GPT-5.6-Luna}{\CaseIncorrect}{%
\CaseStep{Step 1:}{Observe the initial expressions of both subjects, noting that the left subject appears downcast and the right subject looks serious.}{1/1}
\CaseStep{Step 2:}{Identify the change in the left subject's expression, specifically the smile and obvious grin at the five-second mark.}{0/1}
\CaseStep{Step 3:}{Compare the emotional cues from both subjects, with the left subject showing happiness and the right subject remaining neutral or concerned.}{0/1}
\CaseStep{Step 4:}{Synthesize the observations to determine the dominant emotion displayed in the video.}{0/1}
\CaseStep{Step 5 (prediction):}{The predicted emotion is happiness.}{0/1}}
\end{CaseCard}
\CaseNext
\begin{CaseOurs}
\CaseModel{\CaseOursName{\method{}}}{\CaseCorrect}{%
\CaseStep{Step 1:}{The speaker's subdued tone and slow speech rate, particularly the emphasis on "sick" and "cancer," indicate a deep sense of sadness and concern.}{1/1}
\CaseStep{Step 2:}{The slight pause before mentioning "cancer" suggests a moment of reflection, reinforcing the emotional weight of the statement.}{1/1}
\CaseStep{Step 3:}{The individual is seated with a downward gaze and a slumped posture, which are strong indicators of sadness and emotional distress.}{1/1}
\CaseStep{Step 4:}{Their body language, characterized by a lack of engagement and a downward gaze, further emphasizes feelings of melancholy.}{1/1}
\CaseStep{Step 5:}{The overall scene, set in a simple environment, contributes to the somber atmosphere.}{1/1}
\CaseStep{Step 6:}{Collectively, these cues from speech, audio characteristics, and visual expressions converge to suggest a profound sense of sadness.}{1/1}
\CaseStep{Step 7 (prediction):}{The predicted emotion is Sadness.}{1/1}}
\end{CaseOurs}
\end{CasePanel}
\end{table}
\clearpage

\begin{table}[p]
\centering
\caption{\textbf{ML-ER: anger and disgust in a rejecting question.}\newline MME-Emotion example \texttt{ML-ER\_\_00025} has the reference labels \emph{anger, disgust}.}
\label{tab:app-case-mler-question}
\begin{CasePanel}
\CaseFrameGeometry{960}{720}{0}{0}
\CaseFrames{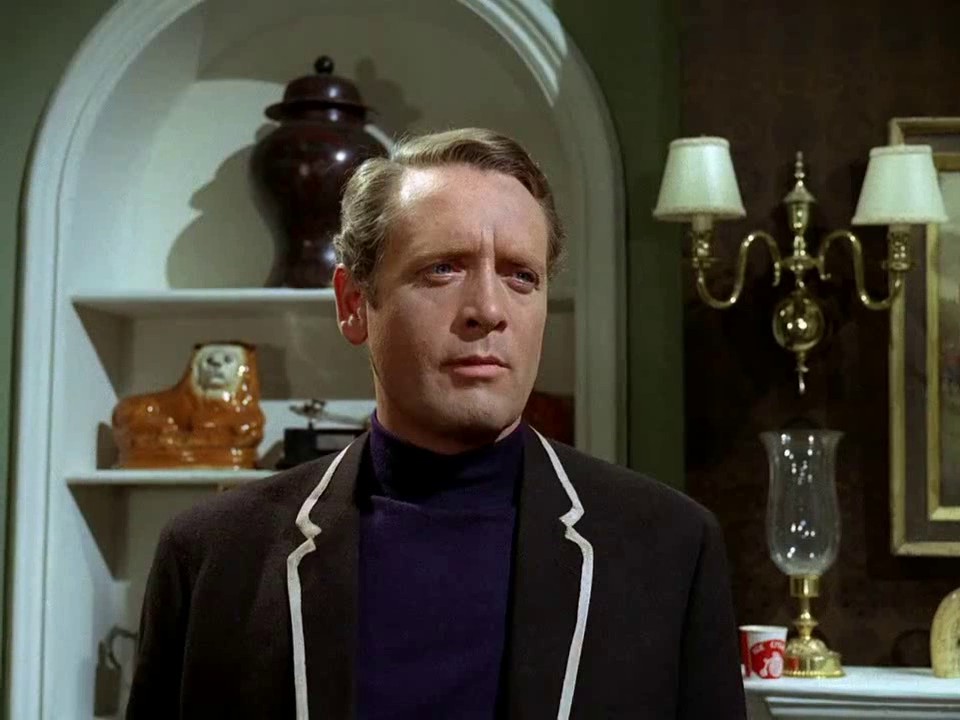}{0.08}{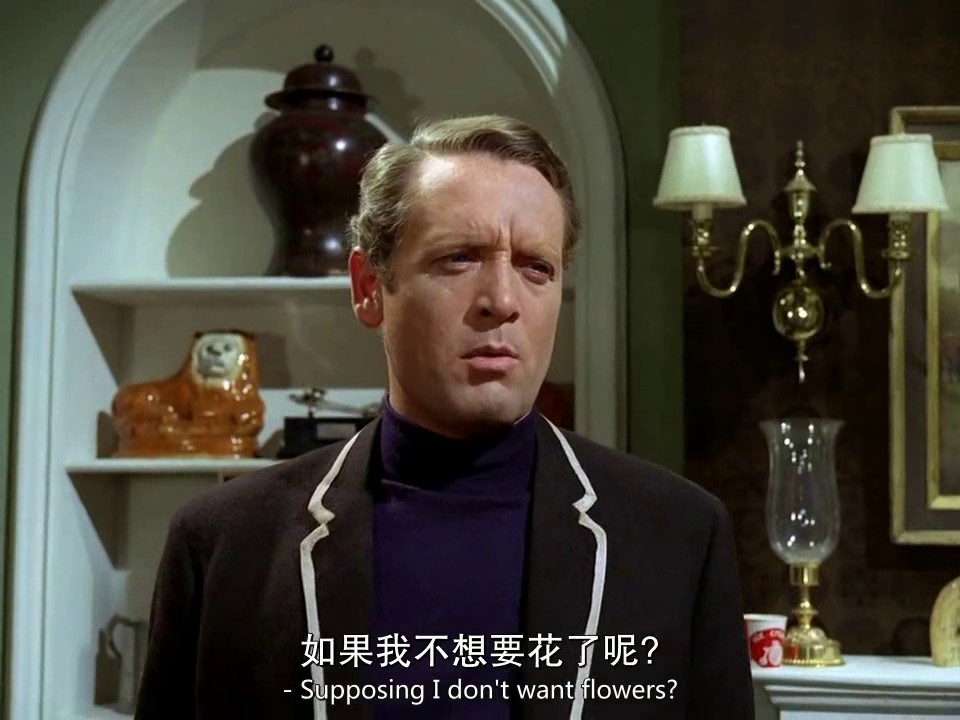}{0.83}{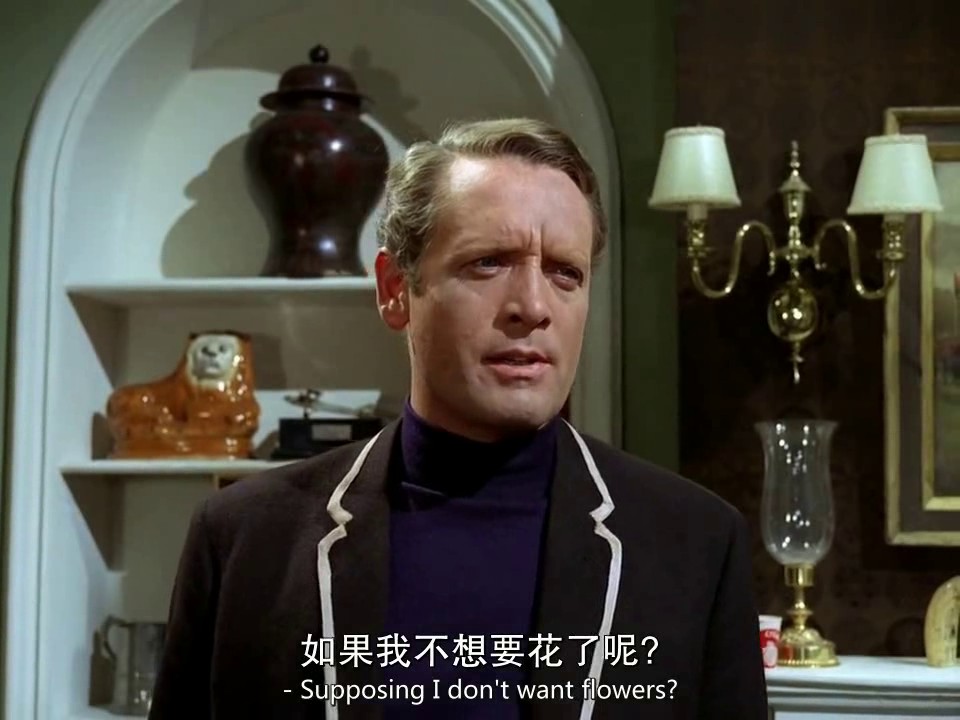}{1.83}
\CaseNext
\begin{CaseCard}
\CaseContextLine{\CaseLabel{Speech:} ``Supposing I don't want flowers?''}{anger, disgust}
\end{CaseCard}
\CaseNext
\begin{CaseCard}
\CaseModelHeader{Reasoning and final prediction}
\CaseModel{Emotion-\newline LLaMA}{\CasePartial}{%
\CaseNote{No reasoning returned.}
\CaseStep{Step 1 (prediction):}{The predicted emotion is anger.}{1/2}}
\CaseRule
\CaseModel{GPT-5.6-Luna}{\CasePartial}{%
\CaseStep{Step 1:}{Observe the man's facial expressions, such as furrowed brow, narrowed eyes, and tense mouth.}{1/1}
\CaseStep{Step 2:}{Note the stern questioning and slightly dismissive look.}{1/1}
\CaseStep{Step 3:}{Interpret these cues as indicators of specific emotions.}{1/1}
\CaseStep{Step 4 (prediction):}{The predicted emotions are anger, contempt, and disappointment.}{1/2}}
\CaseRule
\CaseModel{MiMo-v2.5}{\CaseIncorrect}{%
\CaseStep{Step 1:}{Observe facial features such as furrowed brows (AU1+AU4), eye focus, and mouth shape.}{1/1}
\CaseStep{Step 2:}{Relate these features to emotions, mapping them to distress/sadness and worry/anxiety.}{0/1}
\CaseStep{Step 3:}{Integrate context from the spoken line "Supposing I don't want flowers?" to reinforce disappointment or dissatisfaction.}{0/1}
\CaseStep{Step 4:}{Match observed cues to candidate emotion labels.}{0/1}
\CaseStep{Step 5 (prediction):}{The predicted emotions are sadness, anxiety, and disappointment.}{0/2}}
\CaseRule
\CaseModel{Gemini-3.8-\newline Flash}{\CasePartial}{%
\CaseStep{Step 1:}{Identify candidate emotion labels relevant to the video.}{1/1}
\CaseStep{Step 2:}{Evaluate visual cues such as facial tension, tone, and demeanor.}{1/1}
\CaseStep{Step 3:}{Infer the emotions conveyed based on observed cues.}{1/1}
\CaseStep{Step 4 (prediction):}{The predicted emotions are anger and contempt.}{1/2}}
\end{CaseCard}
\CaseNext
\begin{CaseOurs}
\CaseModel{\CaseOursName{\method{}}}{\CaseCorrect}{%
\CaseStep{Step 1:}{The speaker's vocal tone is characterized by a raised pitch and a sharp, forceful delivery, indicating a heightened emotional state.}{1/1}
\CaseStep{Step 2:}{The emphasis on "supposing" suggests a strong disagreement or frustration, reinforcing the sense of anger.}{1/1}
\CaseStep{Step 3:}{Visually, the person's furrowed brow and tight-lipped expression, combined with a slight upward curve of the lips, suggest a complex emotional state that could be interpreted as a mix of anger and disgust.}{1/1}
\CaseStep{Step 4:}{The context of the question "supposing I don't want flowers?" implies a rejection of a suggestion, further supporting the emotional intensity.}{1/1}
\CaseStep{Step 5:}{The overall body language, including a slight forward tilt of the head, aligns with the emotional cues of anger and disgust.}{1/1}
\CaseStep{Step 6:}{These consistent signals across speech, facial expressions, and body language converge to indicate a strong emotional response of anger and disgust.}{1/1}
\CaseStep{Step 7 (prediction):}{The predicted emotions are anger and disgust.}{1/1}}
\end{CaseOurs}
\end{CasePanel}
\end{table}
\clearpage

\begin{table}[p]
\centering
\caption{\textbf{FG-SA: criticism behind a conversational expression.}\newline MME-Emotion example \texttt{FG-SA\_\_nzpVDcQ0ywM\_6} has the reference label \emph{negative}.}
\label{tab:app-case-fgsa-criticism}
\begin{CasePanel}
\CaseFrameGeometry{480}{360}{0}{0}
\CaseFrames{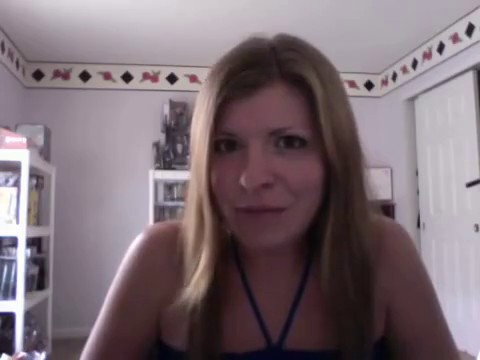}{0.27}{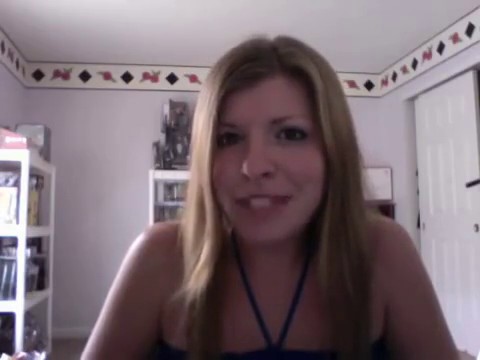}{1.87}{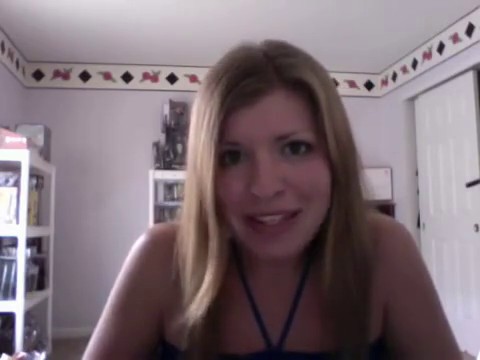}{4.03}
\CaseNext
\begin{CaseCard}
\CaseContextLine{\CaseLabel{Speech:} ``It was really kind of dumbed down for us''}{negative}
\end{CaseCard}
\CaseNext
\begin{CaseCard}
\CaseModelHeader{Reasoning and final prediction}
\CaseModel{VideoLLaMA}{\CaseIncorrect}{%
\CaseStep{Step 1:}{Observe the young woman looking at the camera with a scowl.}{0/1}
\CaseStep{Step 2:}{Note her serious expression while speaking, reinforcing her unhappiness.}{1/1}
\CaseStep{Step 3 (prediction):}{The predicted sentiment is strong negative.}{0/1}}
\CaseRule
\CaseModel{GPT-5.6-Luna}{\CaseIncorrect}{%
\CaseStep{Step 1:}{Observe the woman's facial expression, noting her slight smile and friendly, engaged demeanor.}{0/1}
\CaseStep{Step 2:}{Assess the absence of negative cues and the presence of moderate positive intensity in her expression.}{0/1}
\CaseStep{Step 3:}{Synthesize these observations to determine the overall sentiment.}{0/1}
\CaseStep{Step 4 (prediction):}{The predicted sentiment is positive.}{0/1}}
\CaseRule
\CaseModel{Gemini-3.8-\newline Flash}{\CaseIncorrect}{%
\CaseStep{Step 1:}{Analyze the speaker's verbal content for sentiment cues, such as criticism or disappointment.}{1/1}
\CaseStep{Step 2:}{Observe the speaker's facial expressions and body language for emotional indicators.}{1/1}
\CaseStep{Step 3:}{Integrate verbal and non-verbal cues to assess the overall sentiment intensity.}{1/1}
\CaseStep{Step 4 (prediction):}{The predicted sentiment is weak negative.}{0/1}}
\CaseRule
\CaseModel{Qwen3.8-\newline Flash}{\CaseIncorrect}{%
\CaseStep{Step 1:}{Examine each video frame for facial expressions, focusing on mouth shape, teeth visibility, and cheek movement to distinguish between smiling and talking.}{1/1}
\CaseStep{Step 2:}{Assess the presence and intensity of positive cues (e.g., mild smiles, raised cheeks) versus neutral or negative cues (e.g., pressed lips, furrowed brows).}{0/1}
\CaseStep{Step 3:}{Evaluate the overall demeanor across frames, considering whether the affect is consistently positive, neutral, or mixed.}{0/1}
\CaseStep{Step 4:}{Synthesize the observations to determine the most prominent fine-grained sentiment.}{0/1}
\CaseStep{Step 5 (prediction):}{The predicted sentiment is weak positive.}{0/1}}
\end{CaseCard}
\CaseNext
\begin{CaseOurs}
\CaseModel{\CaseOursName{\method{}}}{\CaseCorrect}{%
\CaseStep{Step 1:}{The speaker's subdued tone and slow speech rate, particularly the emphasis on "dumb down," suggest a sense of frustration or disappointment.}{1/1}
\CaseStep{Step 2:}{This is further supported by the visual cues: a slight frown and downturned mouth indicate a negative emotional state, while the overall relaxed setting contrasts with the intensity of the feelings expressed.}{1/1}
\CaseStep{Step 3:}{The facial expressions, including a frown and lip corner depressor, reinforce the negative sentiment, suggesting a struggle to maintain composure.}{1/1}
\CaseStep{Step 4:}{The combination of these vocal, facial, and contextual elements creates a coherent narrative of a person grappling with negative emotions, leading to the conclusion that the character's emotional state is indeed negative.}{1/1}
\CaseStep{Step 5 (prediction):}{The predicted sentiment intensity is negative.}{1/1}}
\end{CaseOurs}
\end{CasePanel}
\end{table}
\clearpage
\endgroup

\begingroup
\makeatletter
\long\def\@makecaption#1#2{%
  \vskip\abovecaptionskip
  {\normalfont\fontsize{10}{11.5}\selectfont\raggedright #1: #2\par}%
  \vskip6pt}
\setlength{\@fptop}{0pt}
\setlength{\@fpbot}{0pt plus 1fil}
\makeatother
\begin{table}[p]
\centering
\caption{\textbf{EMER: sadness despite a smile.}\newline
\texttt{sample\_00002518} (EMER Dataset, 9.80\,s) has the reference emotion \emph{sadness}.}
\label{tab:app-case-emer-sadness}
\begin{CasePanel}
\CaseFrameGeometry{1280}{720}{92}{92}
\CaseFrames{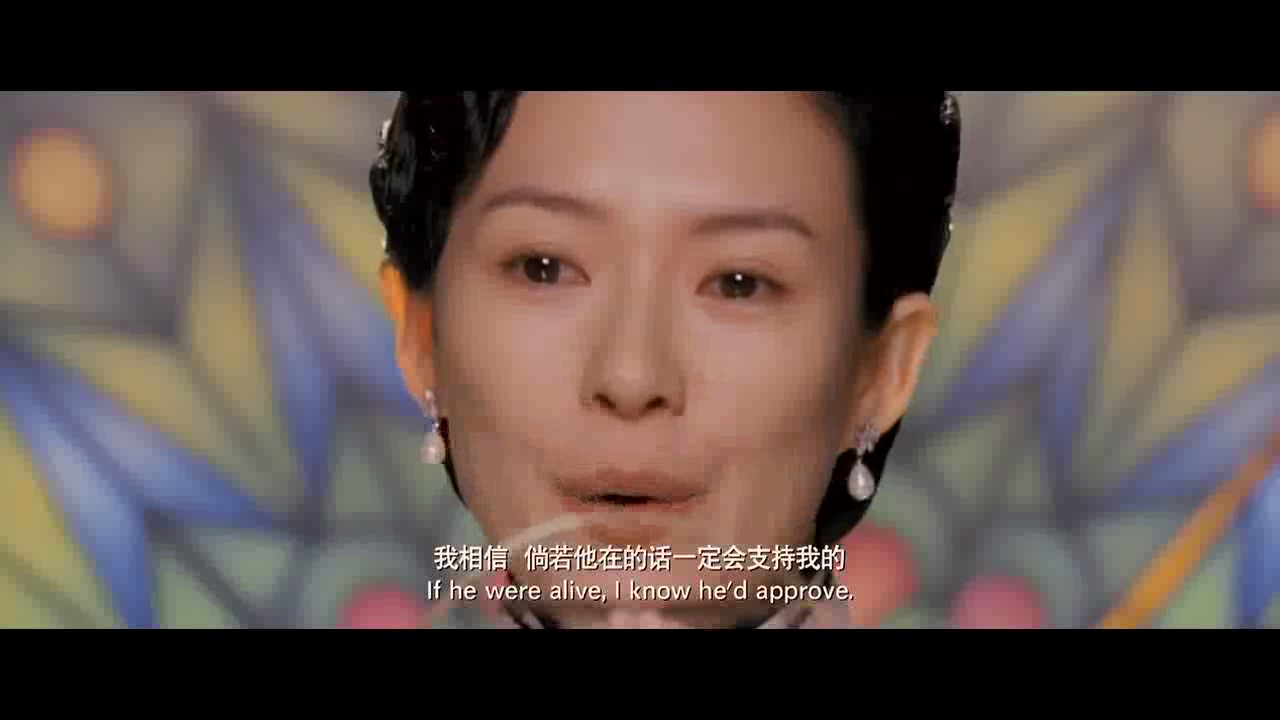}{0.83}{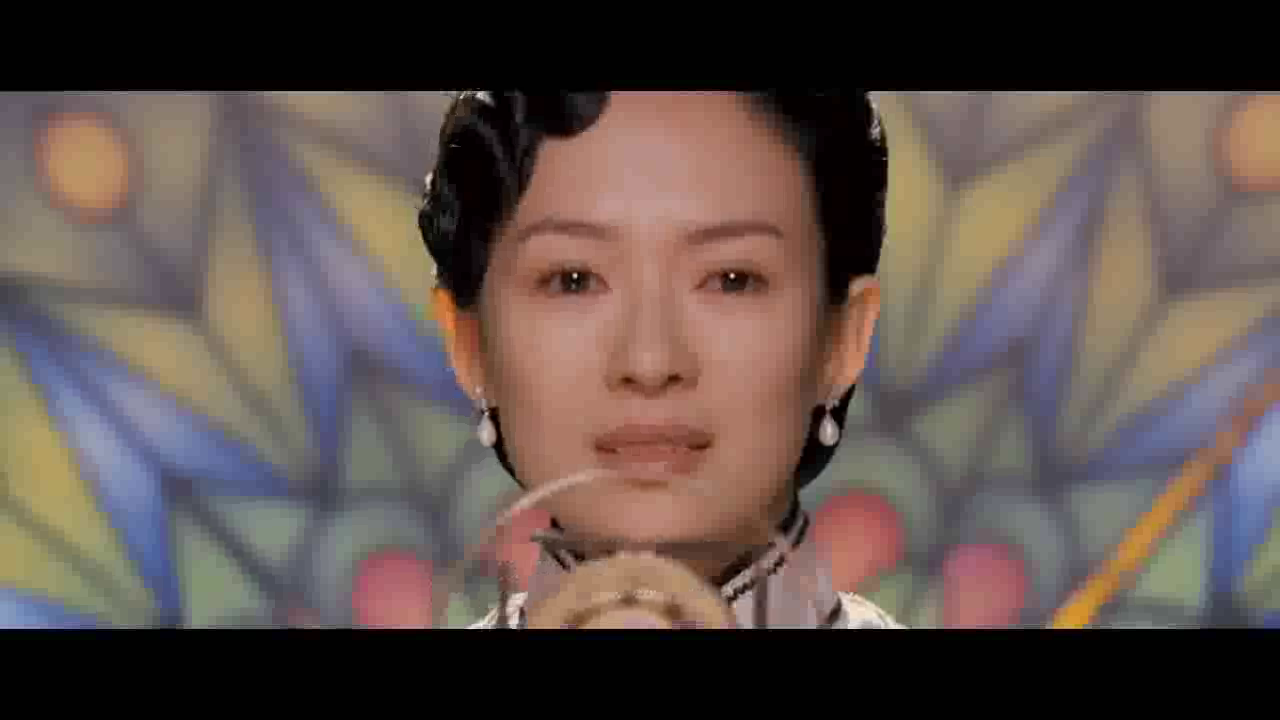}{4.80}{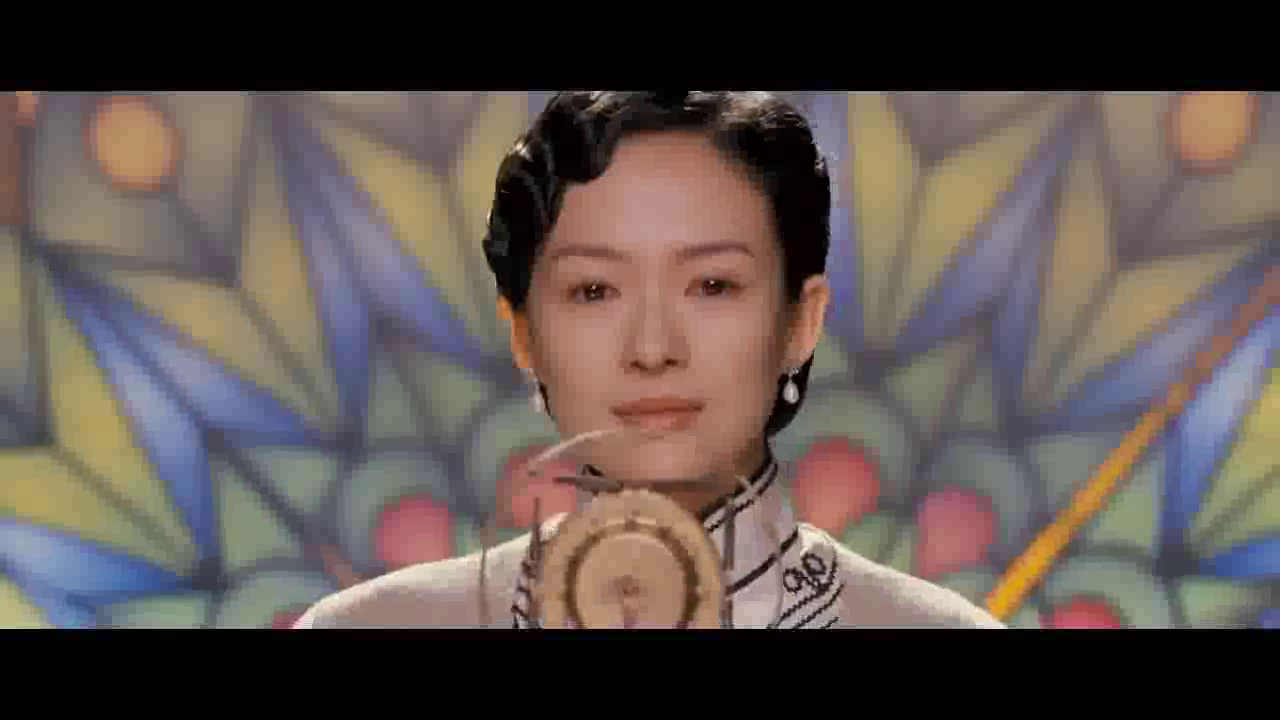}{8.80}
\CaseNext
\begin{CaseCard}
\CaseContextLine{\CaseLabel{Subtitle (dataset translation):} ``If he is here, he will definitely support me.''}{sadness}
\CaseRule
{\CaseJustify\noindent\CaseLabel{Reference (summary):} The speaker misses someone who is absent. Despite smiling, the speaker has tearful, red eyes and speaks slowly, with long pauses and a soft voice, indicating sadness.\par}
\end{CaseCard}
\CaseNext
\begin{CaseCard}
\CaseEmerHeader
\CaseEmerModel{PandaGPT}{2.0}{Based on the image, this person's emotional state appears to be positive and confident. The woman is smiling and looking up, suggesting that she feels happy and optimistic. [\ldots{}]}{Positivity and confidence contradict the reference sadness. Only the smile overlaps, while tears, red eyes, and long pauses are missed.}
\CaseRule
\CaseEmerModel{Valley}{3.0}{[\ldots{}] She seems sad and disappointed in some parts of the video, especially when holding her phone and looking at it. [\ldots{}] Overall, the woman's emotional state throughout the video seems to be a mixture of sadness and happiness.}{Sadness provides partial overlap. Added disappointment, happiness, and contentment weaken the match to the clearly sad reference state.}
\CaseRule
\CaseEmerModel{Video-ChatGPT}{2.0}{We can infer that this person feels happy and excited about the support given by the man in the video. He is smiling and looking at the camera, indicating a positive emotional state.}{Happiness and excitement oppose sadness from missing someone. The smile is shared, but its emotional meaning is misinterpreted.}
\CaseRule
\CaseEmerModel{Video-LLaMA}{1.0}{[\ldots{}] Throughout the video, she keeps smiling and giving a speech. [\ldots{}] Overall, the video portrays a confident and elegant woman.}{Confidence, elegance, and smiling fail to capture sadness, tears, longing, or subdued affect. Emotional overlap with the reference is minimal.}
\CaseRule
\CaseEmerModel{Emotion-\newline LLaMA}{8.0}{[\ldots{}] her tone and intonation convey a sense of longing, as she says, if he were here, he would definitely support me [\ldots{}] therefore, it is inferred that his emotional state is sad}{Sadness and longing are captured. Tearful or red eyes, slow speech with long pauses, and a soft voice are not described.}
\end{CaseCard}
\CaseNext
\begin{CaseOurs}
\CaseEmerModel{\CaseOursName{\method{}}}{9.0}{The speaker's tone is notably subdued, with a slow speech rate and a downward inflection on "support," indicating a sense of sadness. [\ldots{}] The person's expression is marked by a downward gaze and a slight frown, which are common indicators of sadness. [\ldots{}]}{Sadness, subdued tone, slow speech, and facial cues yield high overlap. Tears, red eyes, and the absent person are not explicitly mentioned.}
\end{CaseOurs}
\CaseNext
{\CaseJustify\noindent\color{CaseMuted}\textbf{Scoring (1--10):} Summarize the reference and predicted emotional states, compare their overlap, and assign a higher score for greater overlap. Responses are excerpted, with [\ldots{}] marking omissions. Chinese responses and saved assessment comments are translated, and criteria are abridged. Scores apply to complete responses.\par}
\end{CasePanel}
\end{table}
\clearpage
\endgroup

\endgroup

\end{document}